\documentclass[10pt,twocolumn,letterpaper]{article}

 \usepackage{cvpr}              % To produce the CAMERA-READY version
\usepackage[dvipsnames]{xcolor}

\usepackage[utf8]{inputenc} % allow utf-8 input
\usepackage[T1]{fontenc}    % use 8-bit T1 fonts
\usepackage{url}            % simple URL typesetting
\usepackage{booktabs}       % professional-quality tables
\usepackage{amsfonts}       % blackboard math symbols
\usepackage{nicefrac}       % compact symbols for 1/2, etc.
\usepackage{microtype}      % microtypography
\usepackage{xcolor}         % colors

\usepackage{amsmath}
\usepackage{amssymb}
\usepackage{wrapfig}
\usepackage{subcaption}
\usepackage{color}
\usepackage{balance}
\usepackage{algorithm}
\usepackage{algorithmic}
\usepackage{graphicx}

\usepackage{pgfplots}
\usepackage{tikz}
\usepackage{tikz-3dplot}

\usepackage{ragged2e}
\usepackage{fp}

\usepackage{ifthen}

\usepackage{tikz}
\usetikzlibrary{shapes.geometric, arrows, calc,automata,positioning,decorations,decorations.text}

\usetikzlibrary{mindmap,snakes}

\usepackage{multirow}

\usepackage{algorithm}
\usepackage{algorithmic}
\usepackage{graphicx}
\usepackage{array}
\usepackage[export]{adjustbox}

\usepackage{nicefrac}

\usepackage{xcolor}

\usepackage{colortbl}
\usepackage{makecell}
\usepackage{pifont}

\usepackage{booktabs}
\newcommand{\cmark}{\ding{51}}%
\newcommand{\xmark}{\ding{55}}

\graphicspath{{images/}}

\usepackage[font=small]{caption}

\usepackage{bm}

\usepackage[outline]{contour}

\definecolor{mycitecolor}{rgb}{0.005, 0.3, 0.7}
\usepackage{url}

\colorlet{accclr}{white!20!black}
\colorlet{flopclr}{green!40!blue}
\colorlet{flopclr}{black!10!flopclr}

\colorlet{baseclr}{gray}
\colorlet{rwclr}{white!90!baseclr}
\colorlet{cellclr}{white!90!baseclr}

 \colorlet{dotaclr}{black!40!white}
\colorlet{dotbclr}{red}

\newcommand{\bdota}{\tikz{\node[circle, draw=dotaclr, fill=white!00!dotaclr,scale=0.4]{};}}
\newcommand{\bdotb}{\tikz{\node[circle, draw=dotbclr, fill=white!60!dotbclr,scale=0.4]{};}}

\newcommand{\cmarkc}[1]{\textcolor{#1}{\cmark}}
\newcommand{\xmarkc}[1]{\textcolor{#1}{\xmark}}

\colorlet{xmarkclr}{cyan}
\colorlet{cmarkclr}{magenta}

\newcommand{\buac}[1]{\textcolor{#1}{\raisebox{0.22ex}{\protect\contour{#1}{$\uparrow$}}}}  % bold math symbol
\newcommand{\bdac}[1]{\textcolor{#1}{\raisebox{0.22ex}{\protect\contour{#1}{$\downarrow$}}}}  % bold math symbol

\definecolor{mylinkcolor}{rgb}{0.005, 0.3, 0.7}
\definecolor{suppcolor}{rgb}{0.005, 0.3, 0.5}

\newcommand{\jaesik}[1]{\textcolor{cyan}{Jaesik: #1}\PackageWarning{Jaesik:}{#1!}}

\newcommand{\OursAcronym}{Pick-or-Mix}
\newcommand{\OursAcronymShort}{PiX}
\newcommand{\OursModule}{PiX}

\usepackage{balance}

\usepackage{comment}
\definecolor{cvprblue}{rgb}{0.21,0.49,0.74}
\usepackage[pagebackref,breaklinks,colorlinks,citecolor=cvprblue]{hyperref}

\def\paperID{11246} % *** Enter the Paper ID here
\def\confName{CVPR}
\def\confYear{2024}

\title{\OursAcronym{}: Dynamic Channel Sampling for ConvNets \vspace{-0.75ex}}

\author{Ashish Kumar$^\dagger$, Daneul Kim$^\ddagger$, Jaesik Park$^\ddagger$, Laxmidhar Behera$^\dagger$\\
$^\dagger$~Indian Institute of Technology Kanpur, India \\
$^\ddagger$~Seoul National University, Republic of Korea \\
\texttt{\{\footnotesize ashishkumar822@gmail.com, carpedkm@snu.ac.kr, jaesik.park@snu.ac.kr, lbehera@iitk.ac.in\}}
}

\begin{document}

\maketitle

\begin{abstract}
Channel pruning approaches for convolutional neural networks (ConvNets) deactivate the channels, statically or dynamically, and require special implementation.
In addition, channel squeezing in representative ConvNets is carried out via $1 \times 1$ convolutions which dominates a large portion of computations and network parameters. 
Given these challenges, we propose an effective multi-purpose module for dynamic channel sampling, namely \OursAcronym{} (\OursAcronymShort{}), which does not require special implementation.
\OursModule{} divides a set of channels into subsets and then picks from them, where the picking decision is dynamically made per each pixel based on the input activations. 
We plug \OursModule{} into prominent ConvNet architectures and verify its multi-purpose utilities. 
After replacing $1\times1$ channel squeezing layers in ResNet with \OursModule{}, the network becomes $25\%$ faster without losing accuracy. 
We show that \OursModule{} allows ConvNets to learn better data representation than widely adopted approaches to enhance networks' representation power (e.g., SE, CBAM, AFF, SKNet, and DWP).
We also show that \OursModule{} achieves state-of-the-art performance on network downscaling and dynamic channel pruning applications. 
\\ 
\noindent
\textcolor{suppcolor}{\textbf{Code:}} \textcolor{gray}{\footnotesize{\url{https://github.com/ashishkumar822/PiX}}} \\
\vspace{-4.0ex}

\end{abstract}

\section{Introduction}
\label{sec:intro}
Convolutional neural networks (ConvNets) \cite{vgg, resnet} have been successfully applied to many machine vision tasks \cite{fasterrcnn, perfception}. 
With the introduction of larger models, a general trend is to make them faster via channel pruning. 
Prior works in channel pruning \cite{softfilterpruning, channelgating, fbs, ghostnet} focus on making network lighter to accelerate the inference speed. 
However, some approaches require specialized convolution implementations and pre-trained models~\cite{fbs}, or they are constrained by the baseline accuracy \cite{ghostnet}.
Moreover, whether static or dynamic, these channel pruning methods remove or deactivate the network channels, thus hindering the network from handling difficult inputs \cite{fbs, tang2021manifold}.

It is a fundamental property of ConvNets that for a given spatial location or pixel in the ConvNets' feature map, any one channel may have stronger activation, thus of considerable importance, while for another pixel, the same channel might be less important. 
Therefore, it is crucial to allow the network to \emph{prioritize channels differently per each pixel} instead of dropping a whole channel applied by pruning approaches. 
This inspires us to pick \emph{neuron-specific output from the channels} instead of shutting down an entire channel.

In addition, we observe that standard ConvNet designs still have room for improvement, i.e., $1\times 1$ convolution layers (or called channel squeezing layers) dominate in both number and computations without contributing to the receptive field due to their pixel-wise operation nature. 
For instance, ResNet-$50$~\cite{resnet} consists of $16$ such layers out of $50$, accounting for $\sim 25\%$ ($1.05$B$/4.12$B) of overall FLOPs. 

\begin{figure}
\centering

\begin{tikzpicture}

\node (a)[draw=none, xshift=0ex, yshift=0ex]{
\includegraphics[scale=0.55]{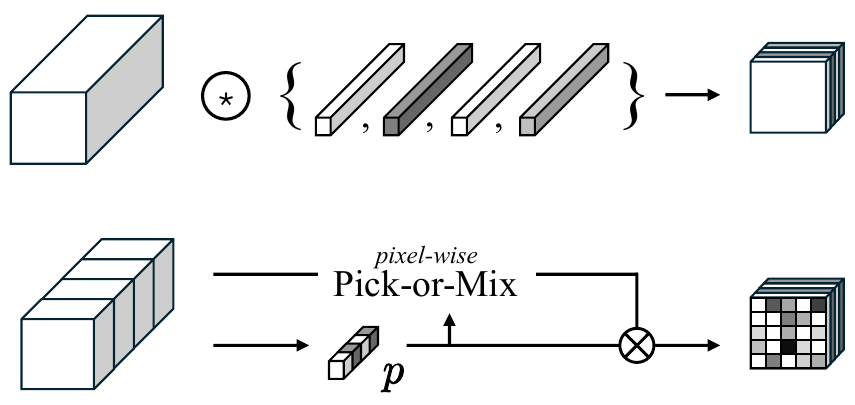}
};
\end{tikzpicture}
% \end{figure}
%

%
\vspace{-1.25ex}
\captionof{figure}{Conceptual overview of \OursModule{} in the context of channel reduction for ConvNets. Top: Traditional dense $1\times 1$ convolution. Although not all channels are important, dense convolutions process all the channels equally. Bottom: \OursModule{} avoids dense convolution and samples the channels dynamically from the input by producing sampling probabilities with far fewer FLOPs. \OursModule{} is multipurpose without requiring specialized implementations.}
\vspace{-2.00ex}
\label{fig:mde}
\end{figure}

In this context, we introduce a novel module, namely \OursAcronym{} (\OursModule{}) that addresses the computational dominance of channel-squeezing layers by \emph{dynamically sampling channels}, 
\OursModule{} transforms a feature map $X \in \mathbb{R}^{C \times H \times W} $ into another one $Y \in \mathbb{R}^{\lceil \nicefrac{C}{\zeta}\rceil \times H \times W} $ (Figure~\ref{fig:mde}). 
Essentially, our method picks or mixes $\lceil C/\zeta \rceil$ channels from the input $C$ channels with a sampling factor $\zeta$. 
It divides a set of channels into subsets and then outputs one channel from each subset via our \OursAcronym{} strategy. \OursModule{} samples the channel based on the \emph{pixel-level runtime decisions} made by the preceding layers; thus, decisions of \OursModule{} are dynamic and input-dependent. 
In addition, \OursAcronym{} does not involve extensive pixel-wise convolution, making the network more efficient. The simple design allows us to plug \OursModule{} into representative ConvNets. 
We plug \OursModule{} into representative ConvNets for the purpose of faster channel squeezing, network downscaling, and dynamic channel pruning.

Our experiments show that \OursModule{} can reduce the computational cost of the vanilla channel squeezing layer (\ie, $1 \times 1$ convolution layer) while maintaining or achieving even better performance, e.g., ResNet becomes $\sim 25\%$ faster without bells and whistles (Sec~\ref{sec:channel_squeezing}, Table~\ref{tab:dcs_as_cs_resnet}). 
\OursModule{} can customize ConvNets in a controlled manner while being faster and more accurate than the baseline counterpart with similar parameters (Sec.~\ref{sec:network_downscaling}, Table~\ref{tab:dcs_as_nds}), e.g., \OursModule{} outperforms recent RepVGG \cite{repvgg} without a complicated training phase while having simple network design. 
We also observe similar accuracy but at reduced parameters (Table~\ref{tab:dcs_vs_existing}). 
\OursModule{} performs better by $\sim3\%$ relative to various recent dynamic channel pruning approaches \cite{fbs,chen2020storage,tang2021manifold,park2023dynamic} on ResNet$18$ with $\sim 2\times$ FLOPs saving.
(Sec.~\ref{sec:dynamic_channel_pruning}, Table~\ref{tab:dcs_as_dcp_res18}). 

We compare the accuracy and FLOPs of \OursModule{} with other state-of-the-art approaches.
We also conduct transfer learning on \OursModule{}-enhanced network on CIFAR-$10$, CIFAR-$100$ for classification, and CityScapes for semantic segmentation. We observe better performance relative to the baselines.

\section{Related Work}
\label{sec:related}
\noindent\textbf{Convolutional Neural Networks.}
The earlier ConvNets \citep{vgg, resnet} are accuracy-oriented but still dominant in the industry \citep{deepquick, repvgg}, thanks to their high representation power, architectural simplicity, and customizability.
EfficientNet~\citep{efficientnet} emerged with network architecture search, but due to its nature of AutoML, it is deep and branched compared to traditional ConvNets~\citep{vgg, resnet}.
Even after half a decade, ResNet continues to improve \citep{attentionalfeaturefusion, sknet}, indicating its architectural significance, while VGG-like architecture continues as it is design-friendly with low-powered computing devices due to its shallow, easily scalable, and low latency design \citep{towards}.
\par
This is also visible from ResNet design space exploration \citep{regnet} that provides a competitive alternative to the advanced ConvNets \citep{efficientnet} while being simpler. 
SENet~\citep{senet}, CBAM~\citep{cbam}, and ResNest~\citep{resnest}, Attentional Feature Fusion \citep{attentionalfeaturefusion} further depict the importance of older architectures by developing novel units to improve the accuracy of ResNet by adding parameters and marginal computational overhead. 
More recently, RepVGG~\citep{repvgg} improves the inference of years old VGG~\citep{vgg} model. 
In this paper, we tackle the overhead of $1 \times 1$ layers in standard ConvNets and expand its application to state-of-the-art transformers.

\vspace{2mm}
\noindent\textbf{Accelerated Inference.}
ConvNet acceleration begins with static pruning \citep{filterpruning} or network compression \citep{automl}. 
These methods~\cite{filterpruning, automl} are model agnostic, but they require the additional overhead of pre-training and fine-tuning, thus increasing the training time~\cite{fbs}. 

Furthermore, by using more efficient convolutions such as depthwise separable convolution~\citep{depthwise}, MobileNets \citep{mobilenetv1, mobilenetv2, shufflenetv1} address this issue at the network architecture level. 
In contrast, \OursAcronymShort{}, without any significant architectural modifications, enables faster inference by providing an alternative to channel squeezing $1 \times 1$ convolutions. 
%

% % % %%%%
\section{\OursAcronym{}~(\OursAcronymShort{})}
\label{sec:dcs}
Modern ConvNets~\citep{resnet, repvgg, resnest} are essentially a stack of convolution layers, but the design of channel squeezing $1 \times 1 $ convolution still has room for improvement.
The main challenge is exploiting the cross-channel information appropriately and developing a suitable mixing strategy to ensure accurate model learning. 

In this section, we introduce \OursAcronym{} (\OursAcronymShort{}) in detail.

\vspace{2mm}
\noindent\textbf{Overview} Consider a tensor $X=\{X^{[1]}, X^{[2]}, ..., X^{[C]}\}$, where $X^{[i]} \in \mathbb{R}^{H \times W}$ denotes $i^{th}$ channel of $X$. We aim to produce $Y=\{Y^{[1]}, Y^{[2]}, ..., Y^{[\lceil C/\zeta \rceil]}\}$, such that $\text{O}(\mathcal{F}_{pix}) \ll \text{O}(\mathcal{F}_{s})$, where $\mathcal{F}_{pix}$ is the \OursAcronymShort{} enhanced network and $\mathcal{F}_{s}$ is the original network. Here, $\zeta \in \mathbb{R}$ is the channel sampling factor which controls the dimensionality of the output $Y$.
The proposed dynamic channel sampling approach (\OursModule{}) progressively infers intermediate $1$D descriptors $z  \in \mathbb{R}^{C}$, $p \in \mathbb{R}^{\lceil \nicefrac{C}{\zeta}\rceil}$ from input feature map $X \in \mathbb{R}^{C \times H \times W}$ for channel sampling by using learnable parameter $\phi = \{\theta, \beta\}$. It then applies per-pixel dynamic channel sampling operator $\pi$ for fusing a subset of channels and produces an output feature map $Y \in \mathbb{R}^{\lceil \nicefrac{C}{\zeta}\rceil \times H \times W}$ of reduced dimensionality that is controllable by the sampling factor $\zeta \in \mathbb{R}_{\geq 1}$.

\begin{figure*}[t]
    \centering
    \includegraphics[width=0.995\linewidth]{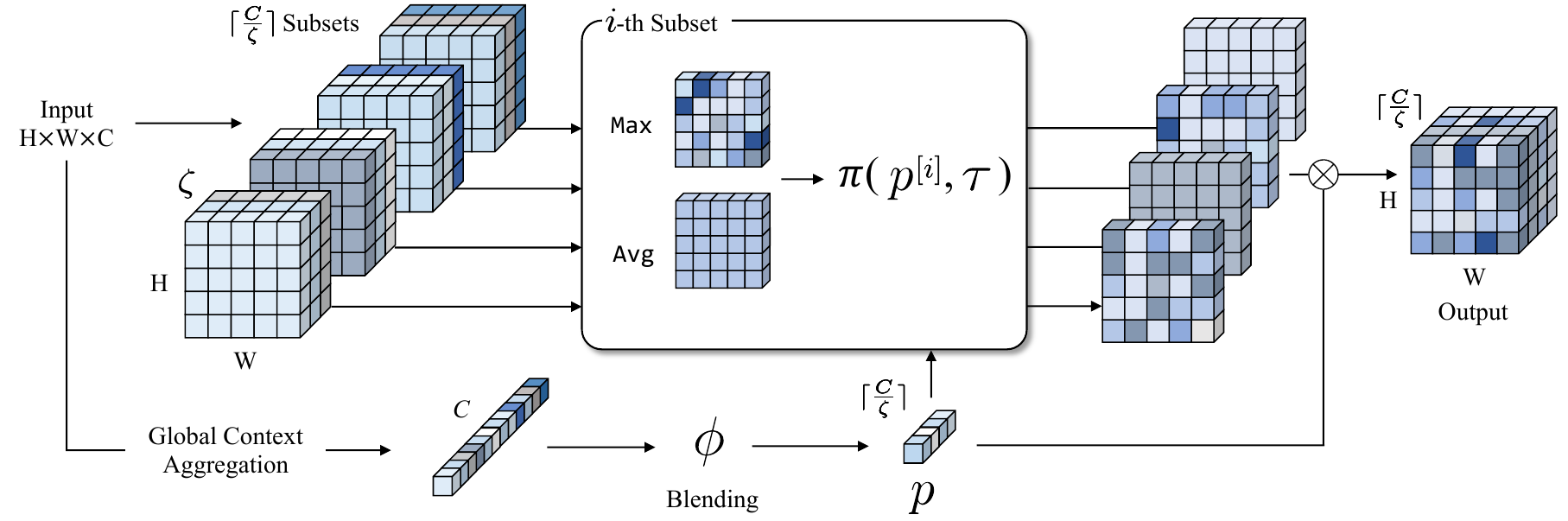}
    % \vspace{-1em}
    \caption{The proposed \OursModule{} module with its Pick-or-Mix dynamic channel sampling strategy. Each subset of input channels is picked (via max operator) or mixed (via average operator) to constitute the squeezed channels of the output. Interestingly, \OursModule{} can fuse channels differently for each pixel (please refer to Sec.~\ref{sec:channelfusion}).  }
     % \vspace{-1.5ex}
    \label{fig:dcsarch}
    \vspace{-2ex}
\end{figure*}

The \OursModule{} module is illustrated in Figure~\ref{fig:dcsarch} and can be sectioned into three stages: (1) global context aggregation, which provides a channel-wise global spatial context in the form of $z$ (Sec.~\ref{sec:gca})
%, similar to \citep{senet}
(2) cross-channel information blending that transforms $ z$ into $p$, referred to as \OursAcronymShort{} sampling probability (Sec.~\ref{sec:fusion_likelihood}), and (3) channel sampling stage that utilizes $p$ and $X$ to produce $Y$. (Sec.~\ref{sec:channelfusion}) 
%
%%
% \vspace{-4cm}
% %
%

\subsection{Global Context Aggregation}
\label{sec:gca}

We define a transformation of global context aggregation as $gca: \mathbb{R}^{C \times H \times W} \rightarrow \mathbb{R}^{C}$ which gathers global context from the input $X$ for each channel:
\begin{equation}
% \resizebox{0.91\hsize}{!}{
gca( X) = \frac{1}{\text{\scriptsize $H$$\times$$W$}} \Big[\text{cc} (X^{[0]} ),\text{cc}(X^{[1]}),..,\text{cc}(X^{[C-1]}) \Big]
% }
\label{eq:gca}
\end{equation}
where, $\text{cc}: \mathbb{R}^{H \times W} \rightarrow \mathbb{R}$ reduces $i^{th}$ channel $ X^{[i]}$ of $ X$ to a scalar. 
We use $l_1$-norm for cc due to its computational efficiency and vectorized parallelization onto GPUs. 
$l_1$-norm of a channel is also known as global pooling, which is commonly employed~\citep{resnet, senet} to aggregate global spatial information.

\subsection{Sampling Probability}
\label{sec:fusion_likelihood}
Now the output of the previous step $ z = gca(X)$ (Eq.~\ref{eq:gca}) is passed through \textit{sampling probability predictor} $\phi$, serving two purposes. 
% First, since each element of $z$ consists of statistics of only a single channel of $X$, the descriptor $z$ lacks cross-channel information.
First, since each element of $z$ consists of spatial information of only a single channel of $X$, the descriptor $z$ lacks cross-channel information.
$\phi$ mitigates this issue by blending the cross-channel information in the elements of $z$. Second, the fusion factor $\zeta$, i.e., $C$ to $\lceil C/\zeta \rceil$, reduces the input number of channels. 
We define $\phi( z) =  z  \theta +  \beta$,
where, $\theta \in \mathbb{R}^{\lceil \nicefrac{C}{\zeta}\rceil \times C}$ and $ \beta \in \mathbb{R}^{\lceil\nicefrac{C}{\zeta}\rceil}$ are the weights and the biases, initialized with \textit{Xavier}~\citep{xavier} and zero respectively.  

After $\phi(z)$, we obtain \emph{channel subset-wise sampling probability} $p \in \mathbb{R}^{\lceil C/\zeta\rceil}_{\geq 0}$ with sigmoid function.

\subsection{Dynamic Channel Sampling}
\label{sec:channelfusion}
We introduce our computationally efficient dynamic channel sampling approach conditioned on $p$ (Sec.~\ref{sec:fusion_likelihood}). 
We can express the dynamic channel sampling with a functor $\mathcal{F} : \mathbb{R}^{C \times H \times W} \rightarrow \mathbb{R}^{\lceil\nicefrac{C}{\zeta}\rceil \times H \times W}$ such that $ Y = \mathcal{F}( X ; p)$.

\vspace{2mm}
\noindent\textbf{Channel Space Partition.}
We partition $X$ into $\lceil \nicefrac{C}{\zeta} \rceil $ subsets. 
Each subset ($\Gamma^{[i]}$, where $i\in\{0,\cdots,\zeta-1\}$) receives a maximum of $\zeta$ channels with the last one lesser than that in case $\nicefrac{C}{\zeta}$ is non-integer.

\vspace{2mm}
\noindent\textbf{Pixel-wise Channel Fusion.}

We devise a channel fusion strategy, namely \textit{\OursAcronym{}} for each partitioned channel subset $\Gamma^{[i]}$. Specifically, for an arbitrary channel subset $\Gamma^{[i]}$, we then apply the channel fusion strategy to obtain a single channel feature map that constitutes one of the output channels. 
$v$ is fused via the following equations:
\begin{equation}
\pi( \Gamma^{[i]}) = 
\begin{cases}
p^{[i]} \times \texttt{Max}(\Gamma^{[i]}),~~~~ & p^{[i]} \leq \tau \\
p^{[i]} \times \texttt{Avg}(\Gamma^{[i]}),~~~~ & p^{[i]} > \tau,
\end{cases}
\label{eq:form1}
\end{equation}
where $\pi$ is \emph{Pick} (selecting the maximum) or \emph{Mix} (averaging responses) channel function function, and $p^{[i]}$ is the pre-calculated sampling probability for a $i$-th subset (Sec.~\ref{sec:fusion_likelihood}). $\tau$ is hyperparameter, set to $0.5$ based on our ablations. 
In Eq.~\ref{eq:form1}, the selection of a fusion operator is performed dynamically via the sampling probability $p$ produced via the input, thus making \OursModule{} input adaptive.

To generalize this idea over the whole input feature map $X$, the functor $\mathcal{F}$ for this strategy can be given as: 
\begin{equation}
\mathcal{F}( X;  p) = \Big[\pi(\Gamma^{[0]}),~\pi(\Gamma^{[1]}),~...,~\pi(\Gamma^{[\lceil \nicefrac{C}{\zeta} \rceil-1]})\Big],
\end{equation}
as depicted in Figure~\ref{fig:dcsarch}.

It is important to note that \emph{channel sampling applies differently for \underline{each spatial location} in \OursModule{}}. 
For example, when $\zeta> 1$ and $p^{[i]} \leq \tau$, with the help of $\texttt{Max}$, the selected channel index in a subset varies for each spatial location (or simply pixel) depending on channel values of that pixel.
Moreover, each $\Gamma^{[i]}$ subset applies a different operator, i.e., some subgroup applies $\texttt{Max}(\cdot)$, and the other applies $\texttt{Avg}(\cdot)$. This subset-wise operation selection introduces $2^{\lceil \nicefrac{C}{\zeta}\rceil}$ combinations, giving numerous ways to fuse the input channels.

Since fusion is done \emph{on a pixel basis}, one pixel may prioritize any channel over another, demonstrating the capability of \OursModule{}.
 
This degree of freedom to \emph{fuse channels dynamically in a spatially varying manner} introduces a high level of non-linearity into the network, which helps to achieve \OursModule{} a competitive accuracy on various tasks with a simplified network structure. When $\zeta=1$, since $\texttt{Max}(v)=\texttt{Avg}(v)$, \OursModule{} will act as global channel-wise attention as in SENet~\citep{senet}.

From the perspective of computation cost, note that $\pi$ just refers to pre-computed $p^{[i]}$ for selecting lightweight operation  ($\texttt{Max}$ or $\texttt{Avg}$), and $\pi$ does not involve expensive pixel-wise $1\times 1$ convolution. Therefore, \OursModule{} can effectively save computation costs.

Our motivation to selectively utilize $\texttt{Max}$ and $\texttt{Avg}$ lies in the fundamentals of ConvNets~\citep{alexnet} where max and avg. pooling are essentially summarization operations. The dynamic decision based on $p^{[i]}$ enables the ConvNets to learn rich representations and allows sub-sampling of the features.

We also support our motivation empirically by employing the $\texttt{Min}$ operator instead of \texttt{Max} or \texttt{Avg}. We observe a performance degradation by roughly $2\%$ (see the supplement).
\subsection{Computational Complexity}
In \OursModule{}, the computation reduction primarily occurs due to collapsing the input tensor $X \in \mathbb{R}^{C \times H \times W}$ into $z\in \mathbb{R}^{C}$. In a naive channel squeezing operation, a $1 \times 1$ convolution is applied densely over $X \in \mathbb{R}^{C \times H \times W}$, having $C \times H \times W$ FLOPs. In contrast, in \OursModule{}, $X$ is first collapsed into $z\in \mathbb{R}^{C}$, and then the sampling probability predictor is applied over $z$, \emph{\underline{resulting in only $C \times (\nicefrac{C}{\zeta})$ FLOPs}}. This is how \OursModule{} saves computations drastically.

Note that the \emph{\underline{only learnable parameter in \OursModule{} is $\theta$ and $\beta$}} as described in Sec.~\ref{sec:fusion_likelihood}.

\subsection{\OursModule{} Embodiment as a Multi-Purpose Module}
\label{sec:multipurpose}
The ability of \OursModule{} to perform channel sampling naturally translates to the underlying operations of different tasks, such as channel squeezing (Sec.~\ref{sec:channel_squeezing}), network scaling (Sec.~\ref{sec:network_downscaling}), and dynamic channel pruning (Sec.~\ref{sec:dynamic_channel_pruning}). 

We describe below in detail how \OursModule{} achieves these objectives despite keeping its structure the same. We also discuss the benefit of using \OursModule{} for these tasks.
Note that it is the functionality of \OursModule{} that it can act as a network downscaler by controlling the channels. However, it is not a direct method of model compression. 
\subsubsection{Channel Squeezing}
\label{sec:channel_squeezing}

Prior works have conducted channel squeezing operations mostly with $1 \times 1$ layers in ResNet-like designs \citep{resnet}.
\OursModule{} maintains a similar level of accuracy to such approaches by utilizing channel sampling probability (Sec.~\ref{sec:fusion_likelihood}) in conjunction with the pixel-wise dynamic channel sampling (Sec.~\ref{sec:channelfusion}).
More importantly, \OursModule{} is \emph{free from expensive dense $1 \times 1$ convolution}. Instead, by operating on a vector $z$, \OursModule{} effectively saves FLOPs and squeezes the channel faster. 

To demonstrate our claims, we replace channel squeezing $1\times 1$ layers in the representative ResNet~\citep{resnet} family (ResNet-$50$, -$101$, and -$152$) with \OursModule{} and evaluate the accuracy, FLOPs, and training and inference time. \OursModule{} speeds up the training and inference, which are empirically verified in Table~\ref{tab:dcs_as_cs_resnet} and Table~\ref{tab:latency} (see the supplement for the details).

Alternatively, channel squeezing can be done via depth-wise pooling in a non-parametric way \citep{dwp}. However, it eliminates all the squeeze convolution layers, resulting in an accuracy drop, as shown in E$4$ in Table~\ref{tab:dcs_vs_existing}.
\subsubsection{Network Downscaling}
\label{sec:network_downscaling}

We can control ConvNets' parameters and computational complexity by adjusting the number of input or output channels. When conducting parameter reduction, it is called network downscaling.
\OursModule{} can achieve this goal via its channel reduction capability. In our approach, the input feature map for each layer is squeezed by the \OursModule{} module with sampling factor $\zeta>1$ and then sent to the next layers.

\OursModule{} module can be inserted into the existing layers, allowing it to downscale ConvNets by changing $\zeta$. 
We use ResNet-$18$, ResNet-$50$, VGG-$16$, and MobileNet for the effectiveness of this application. 
Notably, \OursModule{}-downscaled network variant consistently outperforms the downscaled baseline. \OursModule{}-downscaled networks have the same parameters but lower FLOPs and higher accuracy (Table~\ref{tab:dcs_as_nds}).
\subsubsection{Dynamic Channel Pruning}
\label{sec:dynamic_channel_pruning}
% \vspace{-0.5ex}
%
%

\begin{table*}[t]
\centering

%\vspace{-2ex}

\caption{\OursModule{} as a channel squeezer.
We replace $1\times1$ channel squeezing layers in ResNet with \OursModule{}. We denote the channel squeezing factor of the vanilla network and our modification in the $\zeta$ column. }
\label{tab:dcs_as_cs_resnet}
\vspace{-1.5ex}
\arrayrulecolor{white!60!black}
%\setlength{\arrayrulewidth}{0.1ex}

%\scriptsize
%\footnotesize
% \scriptsize
%\tiny

\resizebox{0.995\linewidth}{!}{
\setlength{\tabcolsep}{6pt}

\begin{tabular}{c l c c l l c c c c c}

\toprule
 & \multicolumn{1}{c}{{Approach}} & \multicolumn{1}{c}{{$\zeta$}} & \multicolumn{1}{c}{{\#Params}} & \multicolumn{1}{c}{{FLOPs~\bdac{accclr}}}  & \multicolumn{1}{c}{{Top-$1$\%~\buac{accclr}}} & \multicolumn{1}{c}{\makecell{Train Time \\ Per-Iteration~\bdac{accclr}}} &  \multicolumn{1}{c}{\makecell{Train Time \\ $120$-Epochs~\bdac{accclr}}} & \multicolumn{1}{c}{\makecell{Train Time \\ $200$-Epochs~\bdac{accclr}}} 
\\

\midrule
 
%\cellcolor{rwclr}

\multicolumn{1}{c|}{\multirow{5}{*}{}} & \bdota{}~\multirow{1}{*}{ResNet-$50$~\cite{resnet}}  & $4$ & $25.5$M    & $4.12$B & $76.30$ & $575$ms & $4.0$ Days  & $6.7$ Days\\
\rowcolor{rwclr}
\multicolumn{1}{c|}{\cellcolor{white}}  &  \bdotb{}~\multirow{1}{*}{ResNet-$50$ + \OursModule{} } & $4$ & $25.5$M  & \textbf{$\mathbf{3.18}$B~(\bdac{accclr}$22.8\%$)} & \textbf{$\mathbf{76.77}$~(\buac{accclr}$0.47\%$)} & \textbf{$\mathbf{359}$ms} & \textbf{$\mathbf{2.5}$ Days} & \textbf{$\mathbf {4.1}$ Days} \\
\rowcolor{rwclr}
\multicolumn{1}{c|}{\cellcolor{white}}  &  \bdotb{}~\multirow{1}{*}{ResNet-$50$ + \OursModule{}($\texttt{Avg}$)}  & $4$ & $25.5$M  & \textbf{$\mathbf{3.18}$B~(\bdac{accclr}$22.8\%$)} & \textbf{$\mathbf{76.58}$~(\buac{accclr}$0.28\%$)} & \textbf{$\mathbf{359}$ms} & \textbf{$\mathbf{2.5}$ Days} & \textbf{$\mathbf {4.1}$ Days}\\
\rowcolor{rwclr}
\multicolumn{1}{c|}{\multirow{-4}{*}{E$0$}\cellcolor{white}}  &  \bdotb{}~\multirow{1}{*}{ResNet-$50$ + \OursModule{}($\texttt{Max}$)}  & $4$ & $25.5$M  & \textbf{$\mathbf{3.18}$B~(\bdac{accclr}$22.8\%$)} & \textbf{$\mathbf{76.57}$~(\buac{accclr}$0.27\%$)} & \textbf{$\mathbf{359}$ms} & \textbf{$\mathbf{2.5}$ Days} & \textbf{$\mathbf {4.1}$ Days}\\

\midrule

\multicolumn{1}{c|}{\multirow{2}{*}{}} & \protect\bdota{}~ResNet-$101$  & $4$  &  $44.5$M & $7.85$B  & $77.21$ & $575$ms & $4.0$ Days & $6.7$ Days \\

\rowcolor{rwclr}
\multicolumn{1}{c|}{\multirow{-2}{*}{E$1$}\cellcolor{white}} &  \protect\bdotb{}~ResNet-$101$ + \OursModule{}   & $4$ & $44.5$M & \textbf{$\mathbf{6.05}$B~(\bdac{accclr}$22.9\%$)}  & \textbf{$\mathbf{77.96}$~(\buac{accclr}$0.45\%$)} & \textbf{$\mathbf{431}$ms} & \textbf{$\mathbf {3.0}$ Days} & \textbf{$\mathbf {5.0}$ Days}\\
\midrule
\multicolumn{1}{c|}{\multirow{2}{*}{}} &  \protect\bdota{}~ResNet-$152$  & $4$  &  $60.1$M & $11.58$B  & $77.78$   & $863$ms & $6.0$ Days & $10.0$ Days \\
\rowcolor{rwclr}
\multicolumn{1}{c|}{\multirow{-2}{*}{E$2$}\cellcolor{white}} &  \protect\bdotb{}~ResNet-$152$ + \OursModule{}   & $4$ & $60.1$M & \textbf{$\mathbf{8.91}$B~(\bdac{accclr}$23.0\%$)}  & \textbf{$\mathbf{78.12}$~(\buac{accclr}$0.44\%$)} & \textbf{$\mathbf{575}$ms} & \textbf{$\mathbf{4.0}$ Days} & \textbf{$\mathbf {6.7}$ Days} \\
\midrule

\multicolumn{1}{c|}{\multirow{2}{*}{}} &  \protect\bdota{}~\multirow{1}{*}{ResNet-$50$}  & $8$ & $12.3$M     & $1.85$B & $73.66$ & $260$ms & $1.8$ Days  & $3.0$ Days\\
\rowcolor{rwclr}
\multicolumn{1}{c|}{\multirow{-2}{*}{E$3$}\cellcolor{white}}  & \protect\bdotb{}~\multirow{1}{*}{ResNet-$50$ + \OursModule{}}  & $8$ & $12.3$M  & \textbf{$\mathbf {1.39}$B~(\bdac{accclr}$24.8\%$)} & \textbf{$\mathbf{74.47}$~(\buac{accclr}$0.81\%$)} & \textbf{$\mathbf{180}$ms} & \textbf{$\mathbf{1.25}$ Days} & \textbf{$\mathbf {2.0}$ Days} \\

\midrule
\multicolumn{1}{c|}{\multirow{2}{*}{}} &   \protect\bdota{}~ResNet-$50$ + SE \cite{senet}  & $4$ & $28.0$M  & $4.13$B & $76.85$ & $575$ms & $4.0$ Days  & $6.7$ Days \\ 

\rowcolor{rwclr}
\multicolumn{1}{c|}{\multirow{-2}{*}{E$4$}\cellcolor{white}} &  \protect\bdotb{}~ResNet-$50$ + SE + \OursModule{} & $4$ & $28.0$M  & \textbf{$\mathbf{3.19}$B~(\bdac{accclr}$22.8\%$)} & $\mathbf{76.95}$~(\buac{accclr}$0.10\%$)   &  \textbf{$\mathbf{359}$ms} & \textbf{$\mathbf{2.5}$ Days} & \textbf{$\mathbf {4.1}$ Days} \\

\bottomrule

\end{tabular}
}
\vspace{-1ex}
\end{table*}
% \end{wraptable}
%

%

When we plug \OursModule{} into a model, it uses $\zeta$ to determine the number of output channels. 
Thus, once $\zeta$ is set, the number of channels obtained from \OursModule{} is deterministic or \emph{static}. 
However, as \OursModule{} selects channels on the fly, meaning that which channels will be sent to the next layer is not predetermined, it leads to a \emph{dynamic reduction behavior}.

For this reason, we call \OursModule{} as static-dynamic channel pruner. This contrasts with the dynamic channel pruning approach, which keeps all the channels in the network intact but decides which ones to compute to save computations. 
This mandates the need for \emph{specialized convolution implementation} to take advantage. 
On the other hand, the static-dynamic behavior of \OursModule{} is free of such necessity, which is of practical significance. 
The static behavior reduces the network's memory footprint and bandwidth while outperforming dynamic channel pruning approaches. 

Please refer to the supplement for the procedure to embody \OursModule{} as a dynamic channel pruner. Table~\ref{tab:dcs_as_dcp_res18} shows a comparison with dynamic pruning approaches. We use ResNet-$18$ and VGG-$16$ for evaluation.

\subsection{Relation With Existing Approaches}

\begin{table}[t]
% \begin{wraptable}[6]{r}[0ex]{0.52\linewidth}
\centering

% \vspace{-3ex}

\caption{A functional comparison of \OursModule{}.}
\label{tab:dcsfunc}
\vspace{-1.5ex}
\arrayrulecolor{white!60!black}
%\setlength{\arrayrulewidth}{0.1ex}

%\scriptsize
%\footnotesize
\scriptsize
%\tiny

\resizebox{0.995\columnwidth}{!}{
\setlength{\tabcolsep}{2pt}
\begin{tabular}{lc c c c c c}

\toprule

\multicolumn{1}{c}{Method} & \tiny \makecell{No\\ Finetuing} & \tiny \makecell{No Custom \\Convolutions} & \tiny \makecell{As a Channel \\ Squeezer} & \tiny \makecell{As a Network\\ Downscalar} & \tiny \makecell{As a Dynamic\\ Pruner} \\ \midrule

\bdota{}~SE \cite{senet}    & \cmarkc{cmarkclr}  & \cmarkc{cmarkclr}  & \xmarkc{xmarkclr}  & \xmarkc{xmarkclr}  & \xmarkc{xmarkclr} \\
\bdota{}~CBAM \cite{cbam}  & \cmarkc{cmarkclr}  & \cmarkc{cmarkclr}  & \xmarkc{xmarkclr}  & \xmarkc{xmarkclr}  & \xmarkc{xmarkclr}  \\ 
\bdota{}~FBS \cite{fbs}   & \xmarkc{xmarkclr}  & \xmarkc{xmarkclr}  & \xmarkc{xmarkclr}  & \xmarkc{xmarkclr}  & \cmarkc{cmarkclr}  \\
\rowcolor{rwclr} 
\bdotb{}~\OursModule{}  & \cmarkc{cmarkclr}  & \cmarkc{cmarkclr}  & \cmarkc{cmarkclr}  & \cmarkc{cmarkclr}  & \cmarkc{cmarkclr} \\

\bottomrule

\end{tabular}
}
\vspace{-2ex}
\end{table}
% \end{wraptable}
%

%%

\noindent\textbf{Using Global Context.} 
We discuss representative approaches that are closest to the proposed \OursModule{}. The idea of using global context was introduced by SENet~\citep{senet} aiming to improve network accuracy, which squeezes and expands a global context vector by using two convolution layers to predict channel saliency. 
CBAM~\citep{cbam} extends SENet, performing both max and avg. pooling during global context extraction then passes them through a shared MLP.
FBS~\citep{fbs} uses global attention to predict channel saliency. FBS picks Top-K channels using the predicted channel saliency, and the suppressed channels are inhibited in the computations of the subsequent layer. 
PiX inherits the idea of using global context to generate sampling probability $p$. (Sec.~\ref{sec:fusion_likelihood})

\vspace{2mm}

\noindent\textbf{Channel Pruning.} 
\OursModule{} differs from existing channel pruning~\citep{fbs} approaches in both structure and functionality. \OursModule{} is not natively a channel pruner; it is the ability of \OursModule{} to sample channels on the fly, which can be utilized as a channel pruner. Therefore, \OursModule{} \emph{does not require an architectural change} to behave as a channel pruner. On the other hand, FBS~\citep{fbs}, for instance, is a channel pruner, and the design is not intended for other purposes, e.g., as a channel squeezer. For reference, we report the accuracy drop when FBS is modified to work as a channel squeezer in Sec.~\ref{sec:dcp_as_cs}.

A functional comparison of \OursModule{} with prior work is shown in Table~\ref{tab:dcsfunc}. We recommend referring to the supplement for visual differences between \OursModule{} and SENet, CBAM, and FBS.
In the supplement, we also provide details on \emph{the memory and FLOPs requirements} of \OursModule{}, SE, CBAM, and FBS. Note that \OursModule{} has the lowest FLOPs and memory consumption.

\vspace{2mm}
\noindent\textbf{Group Convolution.} Apart from the above modules, in terms of operation, the channel space partition should not be confused with group convolution (GC)~\cite{shufflenetv1, shufflenetv2}. In GC, the input channels are divided into groups, and convolution is performed over each group, whereas we perform our \OursModule{} dynamic channel sampling operation onto each pixel. Moreover, the kernel size in GC is a hyperparameter, which does not exist in \OursModule{}. Also, GC requires the input number of channels to be exactly divisible by the number of groups, which is not the case with \OursModule{}. Please see the supplement for the visual differences between GC and \OursModule{}.

% % %%%
% % %%%% 
% % %%%%%%
% % %%%%%% 
% % %%%%
\section{Experiments}
\label{sec:exp}

%%%
\begin{table*}[!t]

\centering

% \vspace{-9ex}

\caption{\OursModule{} as a network downscaler. Increasing $\zeta$ in the networks where our \OursModule{} is applied decreases the number of parameters, working as a network downscaler. For a fair comparison with the baseline networks, we match the size of the ResNet, VGG, and MobileNet family to our downscaled networks. Baseline networks + \OursModule{} consistently shows better accuracy and reduced FLOPs with similar network parameters.} 
\label{tab:dcs_as_nds}
\vspace{-1.5ex}

\arrayrulecolor{white!60!black}
%\setlength{\arrayrulewidth}{0.1ex}

%\scriptsize
%\footnotesize
% \scriptsize

\newcommand{\mymidrule}{\specialrule{0.21em}{0.22em}{0.22em}}

\FPeval{hspa}{5}
\FPeval{hspb}{0-0.5}
\FPeval{vspa}{2}

\resizebox{0.995\linewidth}{!}{
\setlength{\tabcolsep}{8pt}

\begin{tabular}{l r r r | l r r r}

\toprule

\multicolumn{1}{c}{{\multirow{1}{*}{Approach}}} & \multicolumn{1}{c}{{\multirow{1}{*}{\#Param}}} &  \multicolumn{1}{c}{{FLOPs}~\bdac{accclr}} &  \multicolumn{1}{c}{{Top-$1$\%}~\buac{accclr}} & \multicolumn{1}{c}{{\multirow{1}{*}{Approach}}} & \multicolumn{1}{c}{{\multirow{1}{*}{\#Param}}} &  \multicolumn{1}{c}{{FLOPs}~\bdac{accclr}} &  \multicolumn{1}{c}{{Top-$1$\%}~\buac{accclr}} \\
\midrule

\protect\bdota{}~ResNet-$18 \times 1.050$ & $12.80$M & $1.99$B & $71.71$  & \protect\bdota{}~VGG-$16 \times 1.05$ & $16.72$M & $4.20$B & $73.25$  \\ 
\rowcolor{rwclr}
\protect\bdotb{}~ResNet-$18$~+~\OursModule{} $@\zeta=1$ & $12.80$M & \textbf{$\mathbf{1.84}$B} & $\mathbf{73.15}$ & \protect\bdotb{}~VGG-$16$~+~\OursModule{} $@\zeta=1$ & $16.78$M & \textbf{$\mathbf{3.85}$B} & $\mathbf{74.53}$ \\ 

\midrule

\protect\bdota{}~ResNet-$18 \times 0.756$ & $6.77$M & $1.12$B & $69.37$ & \protect\bdota{}~VGG-$16 \times 0.63$ & $8.67$M & $2.26$B & $70.53$   \\ 
\rowcolor{rwclr}
\protect\bdotb{}~ResNet-$18$~+~\OursModule{} $@\zeta=2$ & $6.77$M & \textbf{$\mathbf{0.99}$B} & $\mathbf{70.60}$ & \protect\bdotb{}~VGG-$16$~+~\OursModule{} $@\zeta=2$ & $8.65$M & \textbf{$\mathbf{1.94}$B} & $\mathbf{72.47}$ \\ 

\midrule

\protect\bdota{}~ResNet-$18 \times 0.631$ & $4.78$M & $0.82$B & $67.55$ &\protect\bdota{}~VGG-$16 \times 0.75$ & $5.97$M & $1.59$B & $69.12$ \\ 
\rowcolor{rwclr}
\protect\bdotb{}~ResNet-$18$~+~\OursModule{} $@\zeta=3$ & $4.77$M & \textbf{$\mathbf{0.72}$B} & $\mathbf{68.70}$ & \protect\bdotb{}~VGG-$16$~+~\OursModule{} $@\zeta=3$ & $5.96$M & \textbf{$\mathbf{1.32}$B} & $\mathbf{70.78}$  \\

\midrule

\protect\bdota{}~ResNet-$18 \times 0.555$ & $3.74$M & $0.67$B & $66.10$ & \protect\bdota{} VGG-$16 \times 0.54$ & $4.59$M & $1.25$B & $67.56$ \\ 
\rowcolor{rwclr}
\protect\bdotb{}~ResNet-$18$~+~\OursModule{} $@\zeta=4$ & $3.74$M & \textbf{$\mathbf{0.57}$B} & $\mathbf{67.15}$ & \protect\bdotb{}~VGG-$16$~+~\OursModule{} $@\zeta=4$ & $4.59$M & \textbf{$\mathbf{0.98}$B} & $\mathbf{69.32}$ \\ 

\midrule
\midrule
\protect\bdota{}~ResNet-$50 \times 1.051$ & $28.09$M & $4.51$B & $76.57$ & \protect\bdota{}~MobileNet-v$1$ $\times 1.334$ & $7.04$M & $0.97$B & $74.49$  \\ 
\rowcolor{rwclr}
\protect\bdotb{}~ResNet-$50$~+~\OursModule{} $@\zeta=1$ & $28.08$M & \textbf{$\mathbf{4.13}$B} & $\mathbf{77.65}$ & \protect\bdotb{}~MobileNet-v$1$~+~\OursModule{} $@\zeta=1$ & $7.03$M & \textbf{$\mathbf{0.58}$B} & $\mathbf{74.53}$  \\ 

\midrule

\protect\bdota{}~ResNet-$50 \times 0.732$ & $14.09$M & $2.33$B & $75.62$ & 
\protect\bdota{}~MobileNet-v$1$ $\times 1.0$ & $4.20$M & $0.58$B & $70.60$ \\ 
\rowcolor{rwclr}
\protect\bdotb{}~ResNet-$50$~+~\OursModule{} $@\zeta=2$ & $14.08$M & \textbf{$\mathbf{2.12}$B} & $\mathbf{76.65}$ & \protect\bdotb{}~MobileNet-v$1$~+~\OursModule{} $@\zeta=2$ & $4.06$M & $\mathbf{0.33}$B & $\mathbf{72.27}$  \\ 

\midrule

\protect\bdota{}~ResNet-$50 \times 0.657$ & $11.52$M & $1.95$B & $75.11$  \\ 
\rowcolor{rwclr}
\protect\bdotb{}~ResNet-$50$~+~\OursModule{} $@\zeta=3$ & $11.51$M & \textbf{$\mathbf{1.76}$B} & $\mathbf{75.70}$  \\ 

\cmidrule{1-4}

%\bottomrule

\end{tabular}
}
\vspace{-1ex}
\end{table*}
%\end{wraptable}
%
%

%

\begin{table*}[!t]
\centering

%\vspace{1ex}
\caption{Speed analysis of \OursModule{} as a channel squeezer. \OursModule{} introduces speed gain on various entry-level or low-powered GPUs. We use $@224 \times 224$ px., $@$FP$32$, and the reported numbers are the mean of $25$ runs. `FPS': Frames Per Second. }
\label{tab:latency}
\vspace{-1.5ex}
\arrayrulecolor{white!60!black}
%\setlength{\arrayrulewidth}{0.1ex}

%\scriptsize
%\footnotesize
% \scriptsize
%\tiny

\resizebox{0.995\linewidth}{!}{
\setlength{\tabcolsep}{3pt}
\begin{tabular}{c c c c c c c c c}

\toprule

\multicolumn{1}{c}{{\makecell{NVIDIA GPUs}}} & Cores & Computing power & \multicolumn{1}{c}{{\bdota{}~ResNet-$50$}} & \multicolumn{1}{c}{{\cellcolor{cellclr}
\bdotb{}~\makecell{ResNet-$50$ +\OursModule{}}}} & \multicolumn{1}{c}{{\bdota{}~ResNet-$101$}} & \multicolumn{1}{c}{{\cellcolor{cellclr}
\bdotb{}~\makecell{ResNet-$101$ +\OursModule{}}}} & \multicolumn{1}{c}{{\bdota{}~ResNet-$152$}} & \multicolumn{1}{c}{{\cellcolor{cellclr}
\bdotb{}~\makecell{ResNet-$152$ +\OursModule{}}}}
\\
\midrule

A$40$ & $10752$      & $37.00$ TFLOPs & $142$ FPS & \cellcolor{cellclr}$\mathbf{166}$ FPS ($17\%$~\buac{accclr}) & $90$ FPS& \cellcolor{cellclr}$\mathbf{100}$ FPS ($11\%$~\buac{accclr}) & $66$ FPS & \cellcolor{cellclr}$\mathbf{71}$ FPS ($8\%$~\buac{accclr}) \\
 
RTX-$2080$Ti & $4352$ & $13.45$ TFLOPs & $125$ FPS & \cellcolor{cellclr}$\mathbf{166}$ FPS ($32\%$~\buac{accclr}) & $71$ FPS & \cellcolor{cellclr}$\mathbf{83}$ FPS ($17\%$~\buac{accclr}) & $58$ FPS & \cellcolor{cellclr}$\mathbf{66}$ FPS ($14\%$~\buac{accclr}) \\

GTX-$1080$Ti & $3584$ & $11.45$ TFLOPs & $111$ FPS & \cellcolor{cellclr}$\mathbf{142}$ FPS ($28\%$~\buac{accclr}) & $76$ FPS & \cellcolor{cellclr}$\mathbf{83}$ FPS ($10\%$~\buac{accclr}) & $58$ FPS & \cellcolor{cellclr}$\mathbf{66}$ FPS ($14\%$~\buac{accclr}) \\

Jetson NX & $384$    & $1.00$ TFLOPs  & $20$ FPS & \cellcolor{cellclr}$\mathbf{25}$ FPS ($25\%$~\buac{accclr}) & $13$ FPS & \cellcolor{cellclr}$\mathbf{16}$ FPS ($23\%$~\buac{accclr}) & $10$ FPS & \cellcolor{cellclr}$\mathbf{12}$ FPS ($20\%$~\buac{accclr}) \\

%NVIDIA Jetson Nano & $128$  & $0.23$ TFLOPs  & $140$ms / $7$FPS & \cellcolor{cellclr}$\mathbf{130}$ms / $\mathbf{8}$FPS  & $230$ms / $4$FPS & \cellcolor{cellclr}$\mathbf{200}$ms / $\mathbf{5}$FPS & $320$ms / $3$FPS & \cellcolor{cellclr}$\mathbf{280}$ms / $\mathbf{4}$FPS \\

\bottomrule

\end{tabular}
}
 \vspace{-1ex}
\end{table*}

We evaluate \OursModule{} by plugging it into various prominent ConvNets~\cite{resnet, vgg, mobilenetv1} and Transformers~\cite{efficientvit}, and we compare against recent approaches~\cite{senet, cbam, repvgg, attentionalfeaturefusion, sknet}. 
% Since \OursModule{} is an architecture-level modification into ConvNets, 
We follow the tradition of training the models on ImageNet~\citep{imagenet} with $1.28$M training and $50$K validation images over 1,000 categories for image classification task.
For transfer learning, we use CIFAR-$10$ and CIFAR-$100$ datasets for image classification and CityScapes~\citep{cityscapes} for the downstream task of semantic segmentation. 
We use \citep{flopcounter} for FLOP calculations, which aligns with our theoretical calculations. 

Please see the supplement for training details, code snippets, ablations, and our theoretical FLOP calculations.

\subsection{\OursModule{} as Channel Squeezer} 
Channel Squeezing (Sec.~\ref{sec:channel_squeezing}) aims to reduce FLOPs while maintaining accuracy and parameters (Table~\ref{tab:dcs_as_cs_resnet}).
%

% \noindent\textbf{E$\mathbf{0}$-E$\mathbf{2}$.} 
\vspace{2mm}
\noindent\textbf{{E0 - E2}: \OursModule{} reduces FLOPs by $23\%$ in ResNet family while having better accuracy.}
\OursModule{} achieves computationally efficient squeezing, as visible by the $\sim 23\%$ reduction in FLOPs in all of the \OursModule{} variants. 
Interestingly, ResNet-$101$ + \OursModule{} surpasses the baseline ResNet-$152$ with a significant FLOP difference of $47\%$. 
% This experiment verifies our claim that reusing the parameters of squeeze layer \OursModule{} maintains the non-linearity of the network and learns useful data representations 
We argue that our conjecture on reusing the parameters of \OursModule{} works to maintain the non-linearity of the network is verified.
Also, the empirical result shows that \OursModule{} learns useful data representations (Sec.~\ref{sec:channel_squeezing}). Despite the reduction in FLOPs, \OursModule{} exhibited slight accuracy improvements.
%
%
% \vspace{-2.0ex}
%

% \noindent\textbf{E$\mathbf{3}$.} 
%

\vspace{2mm}
\noindent\textbf{\textbf{E3}: \OursModule{} with a higher squeezing factor.}
We analyze \OursModule{} for a higher squeezing factor, i.e., $\zeta= 8$, and observe that \OursModule{} performs better than the baseline while having almost $25\%$ fewer FLOPs. 
% \dk{\OursModule{} performs better than the baseline while having almost $25\%$ fewer FLOPs, showing that with \OursModule{}, channel sampling is effectively and efficiently done. learns to sample channels efficiently. }
Interestingly, the accuracy gap between ResNet$@\zeta=4$ and $\zeta=8$ is $2.64\%$, while this gap reduces to $2.30\%$ for \OursModule{} at a notable $56\%$ reduction in the FLOPs. 

These empirical results demonstrate the robustness of \OursModule{} towards parameter reduction and its ability to learn to sample channels efficiently. 
% over large scale ImageNet dataset by learning richer data representations. 
%
%
%

% \noindent\textbf{E$\mathbf{4}$.} 
\vspace{2mm}
\noindent\textbf{E4: \OursModule{} enabled squeeze-excitation (SE) networks~\cite{senet} are more accurate.}
\label{sec:exp_dcs_as_cs_with_se}
It is noticeable that \OursModule{} performs better than SE, especially in FLOPs, indicating that \OursModule{} improves the computational performance of SE-like modules. It is because \OursModule{} reduces the computations of the channel squeezing layer from the network equipped with SE-like modules. Hence, the network can take advantage of global attention weighting from SE-like modules and computationally efficient channel squeezing operation via \OursModule{}.

\vspace{2mm}
\noindent\textbf{E0-E4: \OursModule{} reduces training time on ResNet.}
Table~\ref{tab:dcs_as_cs_resnet} also
shows throughput analysis on $8 \times$ NVIDIA $1080$Ti system.
Noticeably, \OursModule{} has the lowest per-iteration time, which reduces the overall training duration. Since \OursModule{} reduces the computations of the channel squeezing $1$$\times $$1$ layers, this indicates that $1$$\times $$1$ squeeze layers are a computational bottleneck in ResNet.

\subsection{Inference Latency}
Since FLOPs are not an accurate measure of the actual speed \cite{repvgg}, we conduct a latency analysis on four different types of GPUs (Table~\ref{tab:latency}). The first three are entry-level desktop GPUs, while the last one is a low-powered ($10$W) embedded computing device that is far less powerful.
The table shows that \OursModule{} brings a maximum of $32\%$ speedup, which demonstrates the practicality of \OursModule{} for real-time applications.
\subsection{\OursModule{} as Network Downscaler}
Along with channel squeezing, \OursModule{} also offers simplified network downscaling (Sec.~\ref{sec:network_downscaling}).
By increasing $\zeta$, we achieve a similar effect to that of network downscaling, outperforming the downscaled networks by other approaches.
We used width scaling (increasing the number of channels in each conv layer) for the baseline.

The empirical result in Table~\ref{tab:dcs_as_nds} shows that our proposed \OursModule{} is seamlessly applicable for network downscaling regardless of network architectures (ResNet-18, ResNet-50, VGG-16, and even on MobileNet-v1), showing superior performance than all the baselines.
It shows the diverse scope and applicability of \OursModule{} in low-powered devices for customizing a network for a dedicated purpose.
%
%

%%
%

%
%

%
%
%

%%%
\begin{table}[!t]
% \begin{wraptable}[8]{r}[0ex]{0.55\linewidth}
\centering

%\vspace{-3ex}
\caption{\OursModule{} + ViT. We replace the vanilla channel squeezing layer with \OursModule{} in the feed-forward network (FFN) of recent EfficientViT \cite{efficientvit}. We observe that the utility of \OursModule{} also transfers to the Transformer models, as evidenced by the reduced runtime. \textit{Note:} EfficientViT uses a squeezing factor of two in its FFN.}

%\vspace{1ex}
\label{tab:efficientvit}
\vspace{-1.5ex}

\arrayrulecolor{white!60!black}
%\setlength{\arrayrulewidth}{0.1ex}

%\scriptsize
%\footnotesize
% \scriptsize

\resizebox{0.995\columnwidth}{!}{
\setlength{\tabcolsep}{2pt}

\begin{tabular}{l c c c c c}

\toprule
 \multicolumn{1}{c}{{Approach}} & \multicolumn{1}{c}{{$\zeta$}} & \multicolumn{1}{c}{\#Param} & \multicolumn{1}{c}{FLOPs}  & \multicolumn{1}{c}{Top-$1$\%~} & Training Hours 
\\

\midrule
 
%\cellcolor{rwclr}

\bdota{}~\multirow{1}{*}{EfficientViT-M$5$~\cite{efficientvit}}  &  & $12$M & $522$M & $76.8$ & $36$ \\
\rowcolor{rwclr}
\bdotb{}~\multirow{1}{*}{EfficientViT-M$5$ + \OursModule{} } & $2$ & $12$M  & $522$M & \textbf{$76.9$} & $\textbf{24}$\\
\midrule
 
%\cellcolor{rwclr}

\bdota{}~\multirow{1}{*}{EfficientViT-M$5$~\cite{efficientvit}} $\times0.5$  &  & $3.2$M & $136$M & $67.8$ & $32$\\
\rowcolor{rwclr}
\bdotb{}~\multirow{1}{*}{EfficientViT-M$5$ + \OursModule{}} $\times0.5$ & $2$ & $3.2$M  & $136$M & \textbf{$67.8$} & $\textbf{24}$\\

\bottomrule

\end{tabular}
}
\vspace{-2ex}
\end{table}

\subsection{\OursModule{} into Vision Transformers (ViT)}
Although our approach is designed for ConvNets, we go even further and apply \OursModule{} into ViT models to investigate the feasibility. We apply \OursModule{} to the feed-forward network (FFN) of the ViTs, which is essentially a stack of channel expansion $1 \times1$ layer followed by a channel squeezing $1 \times1$ layers. We experiment with the latest EfficientViTs \cite{efficientvit}. We choose the EfficientViT-M$5$ variant. 

Since FFN layers form only a small portion of Transformers, the parameter and FLOPs roughly remain the same, as shown in Table~\ref{tab:efficientvit}. However, the wall time of the \OursModule{} variant is smaller, reducing the training time from 36 hours to 24 hours and reducing the downscaled model's training time from 32 hours to 24 hours.
Despite similar FLOPs, the functioning of \OursModule{} requires less memory access, which reduces the memory access cost (MAC) and hence latency \cite{repvgg}.

We believe that with further improvement in the context of ViTs, the classification performance of \OursModule{} can be improved, which we leave as future work.

\subsection{\OursModule{} as Dynamic Channel Pruner}
The ability of \OursModule{} to pick channels dynamically is similar to dynamic pruning (Sec.~\ref{sec:dynamic_channel_pruning}). The difference is that \OursModule{} selects the channels dynamically while existing approaches turn off a few channels. We compare \OursModule{} with dynamic pruning approaches.
%
%

%%%
\begin{table}[t]
% \begin{wraptable}[8]{r}[0ex]{0.55\linewidth}
\centering

%\vspace{-3ex}
\caption{\OursModule{} as a dynamic channel pruner. 
We compare our approach with representative dynamic or static channel pruning methods using ResNet-18 and VGG-16. Vanilla ConvNet + \OursModule{} shows compatible accuracy and FLOPs saving gain.}

%\vspace{1ex}
\label{tab:dcs_as_dcp_res18}
\vspace{-1.5ex}

\arrayrulecolor{white!60!black}
%\setlength{\arrayrulewidth}{0.1ex}

%\scriptsize
%\footnotesize
% \scriptsize

\resizebox{0.995\columnwidth}{!}{
\setlength{\tabcolsep}{1pt}

\begin{tabular}{l c c c c}

\toprule

\multicolumn{1}{c}{{\multirow{2}{*}{$@$~ResNet-$18$}}} & \multicolumn{1}{c}{{\multirow{2}{*}{Dynamic}}}&  \multicolumn{2}{c}{{Top-$1$\%}~\buac{accclr}} &  \multicolumn{1}{c}{{\multirow{2}{*}{FLOPs Saving\buac{accclr}}}} 
\\
\cmidrule{3-4}

& & \multicolumn{1}{c}{{Baseline}} &  \multicolumn{1}{c}{{Downscaled}} & \\

\midrule

\protect\bdota{}~Soft Filter Pruning \cite{softfilterpruning} & & $70.28$  & $67.10$ & $1.72\times$  \\
%ResNet-$18$+NS \cite{softfilterpruning} &  & $68.98$  & $67.21$ & $1.39\times$  \\
\protect\bdota{}~Discrimination-aware~\cite{discriminationpruning} &  & $69.64$  & $67.35$ & $1.89\times$  \\
\protect\bdota{}~Collaborative Layers~\cite{lowcostcolablayer} & \cmark  & $69.98$  & $67.33$ & $1.53\times$  \\
\protect\bdota{}~Channel Gating~\cite{channelgating} & \cmark & $69.02$  & $67.40$ & $1.61\times$  \\
\protect\bdota{}~Boosting and Suppression~\cite{fbs} & \cmark   & $70.71$  & $68.17$ & $1.98\times$  \\ 
\protect\bdota{}~Storage Efficient Pruning \cite{chen2020storage} & \cmark   & $69.76$  & $68.73$ & $1.94\times$  \\
\protect\bdota{}~Manifold Reg. Pruning \cite{tang2021manifold} & \cmark   & $69.76$  & $68.88$ & $2.06\times$  \\ 
\protect\bdota{}~Dynamic Struct. Pruning \cite{park2023dynamic} & \cmark   & $69.76$  & $68.38$ & $2.56\times$  \\ \midrule

\rowcolor{rwclr}
\protect\bdotb{}~\OursModule{} & \cmark  & $\mathbf{73.15}$  & $\mathbf{70.60}$ & $1.85\times$ \\
\bottomrule

\toprule

\multicolumn{1}{c}{$@$~VGG-$16$} & \multicolumn{1}{c}{Dynamic}  & \multicolumn{2}{c}{$\Delta$ Top-$5$~\buac{accclr}} &   \multicolumn{1}{c}{FLOPs Saving\buac{accclr}} 
\\

\midrule

\protect\bdota{}~Filter Pruning  \cite{filterpruning} &  & \multicolumn{2}{c}{$-8.6$}  & $4 \times$\\
\protect\bdota{}~Runtime Neural Pruning \cite{runtimepruning}  & \cmark & \multicolumn{2}{c}{$-2.32$}  & $3 \times$\\
\protect\bdota{}~AutoML Compression \cite{automl} &    & \multicolumn{2}{c}{$-1.4$}  & $5 \times$\\ 
\protect\bdota{}~ThiNet-Conv \cite{thinetconv} &    & \multicolumn{2}{c}{$-0.37$}  & $3 \times$ \\ 
\protect\bdota{}~Boosting and Suppression  \cite{fbs} & \cmark  & \multicolumn{2}{c}{$\mathbf{-0.04}$}  & $3 \times$ \\ \midrule
\rowcolor{rwclr}
\protect\bdotb{}~\OursModule{} & \cmark  & \multicolumn{2}{c}{$\mathbf{-0.04}$}  & $3 \times$ \\
\bottomrule

\end{tabular}
}

 \vspace{-1ex}
\end{table}
% \end{wraptable}
%
% %
%
%%

\vspace{2mm}
\noindent\textbf{\OursModule{} \emph{vs.} dynamic pruning approaches.} 

Referring to Table~\ref{tab:dcs_as_dcp_res18}, the \OursModule{} baseline (i.e., ResNet-$18$ + \OursModule{} $@\zeta=1$, Top-1 Acc. 73.15\%) and the downscaled (ResNet-$18$ + \OursModule{} $@\zeta=3$, Top-1 Acc. 70.60\% in Table~\ref{tab:dcs_as_nds}), shows compelling performance than the state-of-the-art dynamic pruning approaches~\cite{softfilterpruning,discriminationpruning,lowcostcolablayer,channelgating,fbs,chen2020storage,tang2021manifold,park2023dynamic}.

Note that \OursModule{} does not require fine-tuning to obtain better performance, unlike other approaches, such as \citep{fbs}, leading to a simpler pipeline of \OursModule{}.

Following \citep{filterpruning,runtimepruning, fbs}, we report $\Delta$Top-$5$ error with the benefit of FLOP reduction using VGG-$16$ as a baseline. 
Table~\ref{tab:dcs_as_dcp_res18} shows that \OursModule{} offers a competitive performance than other approaches~\cite{filterpruning,runtimepruning,automl,thinetconv,fbs}.

\vspace{2mm}
\noindent\textbf{Existing dynamic channel pruning approach is not multipurpose.}
\label{sec:dcp_as_cs}
To highlight the key advantage of \OursModule{} that it does not need to change its structure to serve different purposes,
we customize FBS \citep{fbs} for channel squeezing, although FBS is not intended to perform. FBS was chosen because of its strong resemblance with disabling channels via global attention.
FBS picks top-k channels in its original operation and has the same input-output dimensions, i.e., $\in \mathbb{R}^{C \times H \times W} $. 
However. for this experiment, we configure FBS to output $\in \mathbb{R}^{\lceil \nicefrac{C}{k}\rceil \times H \times W} $, where $k=\zeta$.

We then replace all the channel squeezing layers with this modified FBS module and train the model. We observe that FBS faces convergence issues.
We identify the underlying cause is due to the drop-out of intermediate channels from the input $X$ when selecting top-k channels.
Also, the channels appearing in the output ($Y$) that lost position identity
or channel index causes convergence issues. 
When $Y$ is operated upon via subsequent convolutions, the approach is not intended to learn the relation between the channels, as the position or index of a given channel in $X$ keeps changing in $Y$. 
This indicates that FBS-like pruning methods can not complement \OursModule{}, but vice-versa is possible, as demonstrated earlier, highlighting the utility of \OursModule{}.

\subsection{\OursModule{} in the Wild}
We compare \OursModule{} with prior works~\cite{senet,cbam,attentionalfeaturefusion,sknet} in improving ResNet accuracy and feature fusion via the attention mechanism~\cite{senet, cbam, attentionalfeaturefusion}. We present the result in Table~\ref{tab:dcs_vs_existing}.
%
%
%%
%

% \indent\textbf{E$\mathbf{0}$-E$\mathbf{2}$}
%
\vspace{2mm}
\noindent\textbf{E0-E2: \OursModule{} \emph{vs.} SE~\cite{senet} and CBAM~\cite{cbam}.}
% We compare \OursModule{} with the methods which aim at improving accuracy by inserting new computing units into a baseline model, such as \citep{senet, cbam}. 
We compare \OursModule{} with the methods that aim to improve performance with the newly proposed layer.
% \dk{We compare \OursModule{} with the methods which improve}
We observe that \OursModule{} performs better than SE and CBAM, even on MobileNet \citep{mobilenetv1}, while the proposed~\OursModule{} has a simpler structure and multi-purpose utility.
%
%

%%%
\begin{table}[t]
% \begin{wraptable}[15]{r}[0ex]{0.57\linewidth}
\centering

% \vspace{-3ex}

\caption{\OursModule{} \textit{vs}. existing approaches for enhancing the accuracy of the vanilla ConvNets. `$^\star$' denotes that \OursModule{} is applied only before the second layer of a ResNet-$18$ block (see the supplement). }
\label{tab:dcs_vs_existing}
\vspace{-1.5ex}

\arrayrulecolor{white!60!black}
%\setlength{\arrayrulewidth}{0.1ex}

%\scriptsize
%\footnotesize
\scriptsize

\resizebox{0.995\columnwidth}{!}{
\setlength{\tabcolsep}{2pt}

\begin{tabular}{c l c c c}

\toprule

& \multicolumn{1}{c}{{\multirow{1}{*}{Approach}}} & \multicolumn{1}{c}{{\multirow{1}{*}{\#Params}~\bdac{accclr}}}&  \multicolumn{1}{c}{{FLOPs}~\bdac{accclr}} &  \multicolumn{1}{c}{{Top-$1$\%}~\buac{accclr}} \\

\midrule

\multicolumn{1}{c|}{\multirow{4}{*}{}} & \protect\bdota{}~ResNet-$18$ \cite{resnet} & $11.60$M & $1.81$B & $70.40$  \\ 

\multicolumn{1}{c|}{\multirow{4}{*}{}} & \protect\bdota{}~ResNet-$18$ + SE \cite{senet} & $11.78$M & $1.81$B & $70.59$  \\ 
\multicolumn{1}{c|}{} & \protect\bdota{}~ResNet-$18$ + CBAM \cite{cbam} & $11.78$M & $1.81$B & $70.73$ \\ 
\rowcolor{rwclr}
\multicolumn{1}{c|}{\cellcolor{white}} & \protect\bdotb{}~ResNet-$18$ + \OursModule{}$^\star$ & $11.88$M & $1.81$B & $\mathbf{71.65}$ \\ 
\rowcolor{rwclr}
\multicolumn{1}{c|}{\multirow{-4}{*}{E$0$\cellcolor{white}}} & \protect\bdotb{}~ResNet-$18$ + \OursModule{} & $12.80$M & $1.84$B & $\mathbf{73.15}$ \\ 

\midrule

\multicolumn{1}{c|}{\multirow{3}{*}{}}  & \protect\bdota{}~ResNet-$50$  & $25.50$M & $4.12$B & $76.30$  \\ 
\multicolumn{1}{c|}{\multirow{3}{*}{}}  & \protect\bdota{}~ResNet-$50$ + SE \cite{senet} & $28.09$M & $4.13$B & $76.85$  \\ 
\multicolumn{1}{c|}{} & \protect\bdota{}~ResNet-$50$ + CBAM \cite{cbam} & $28.09$M & $4.13$B & $77.34$   \\ 
\rowcolor{rwclr}
\multicolumn{1}{c|}{\multirow{-3}{*}{E$1$\cellcolor{white}}} & \protect\bdotb{}~ResNet-$50$ + \OursModule{} & $28.08$M & $4.13$B & $\mathbf{77.65}$ \\ 

\midrule
\multicolumn{1}{c|}{\multirow{3}{*}{}}  & \protect\bdota{}~MobileNet \cite{mobilenetv1} & $4.23$M & $0.56$B & $68.61$  \\ 
\multicolumn{1}{c|}{\multirow{3}{*}{}}  & \protect\bdota{}~MobileNet + SE \cite{senet} & $5.07$M & $0.57$B & $70.03$  \\ 
 \multicolumn{1}{c|}{} & \protect\bdota{}~MobileNet + CBAM \cite{cbam} & $5.07$M & $0.57$B & $70.99$   \\ 
 \rowcolor{rwclr}
\multicolumn{1}{c|}{\multirow{-3}{*}{E$2$}\cellcolor{white}}  & \protect\bdotb{}~MobileNet + \OursModule{} & \textbf{$\mathbf{4.06}$M} & \textbf{$\mathbf{0.33}$B} & $\mathbf{72.27}$   \\ 
% %

%
\midrule
\multicolumn{1}{c|}{\multirow{3}{*}{}}  & \protect\bdota{}~ResNet-$50$ + AFF \cite{attentionalfeaturefusion} $@160$ Epochs  & $30.30$M & $4.30$B & $79.10$  \\ 
\multicolumn{1}{c|}{} & \protect\bdota{}~ResNet-$50$ + SKNet \cite{sknet} $@160$ Epochs & $27.70$M & $4.47$B & $79.21$  \\ 
\rowcolor{rwclr}
\multicolumn{1}{c|}{\multirow{-3}{*}{E$3$}\cellcolor{white}} &\protect\bdotb{} ResNet-$50$ + \OursModule{} $@160$ Epochs & $28.08$M & $\mathbf{4.13}$B & $\mathbf{79.40}$   \\ 

\midrule

\multicolumn{1}{c|}{\multirow{2}{*}{}}  & \protect\bdota{}~RepVGG-A$0$ \cite{repvgg}  & $9.10$M & $1.51$B & $72.41$  \\ 

\rowcolor{rwclr}
\multicolumn{1}{c|}{\multirow{-2}{*}{E$4$}\cellcolor{white}} & \protect\bdotb{}~VGG-$16$ \cite{vgg} + \OursModule{} & $\mathbf{8.65}$M & $1.94$B & $\mathbf{72.47}$  \\ 

\midrule

\multicolumn{1}{c|}{\multirow{2}{*}{}}  &  \protect\bdota{}~ResNet-$50$ + DWP \cite{dwp} & $19.60$M & $2.82$B & $75.35$  \\ 

\rowcolor{rwclr}
\multicolumn{1}{c|}{\multirow{-2}{*}{E$5$}\cellcolor{white}} & \protect\bdotb{}~ResNet-$50$ + \OursModule{} $@\zeta=2$ & $\mathbf{14.08}$M & $\mathbf{2.12}$B & $\mathbf{76.65}$ \\ 

\bottomrule

\end{tabular}
}
\vspace{-3ex}
%
% \end{wraptable}
\end{table}
%
%
%
%%

% \indent\textbf{E$\mathbf{3}$}
%
\vspace{2mm}
\noindent\textbf{E3: \OursModule{} \emph{vs.} AFF~\cite{attentionalfeaturefusion} and SKNet~\cite{sknet}.}
Attentional Feature Fusion (AFF) fuses two feature maps adaptively, and SKNet improves accuracy by adaptively weighting the output of two convolutions with different kernel sizes. These models are trained for longer epochs. Therefore, we also train \OursModule{} at the same setting \citep{attentionalfeaturefusion}. 
% We observe that \OursModule{} outperforms these two methods while being architecturally. 
We observe that \OursModule{} outperforms these two methods while being architecturally simple.
%
%
%
% \vspace{-2.0ex}

%
\vspace{2mm}
\noindent\textbf{E$\mathbf{4}$: \OursModule{} + VGG \emph{vs.} RepVGG~\citep{repvgg}.}
RepVGG is a recent approach that speeds up VGG \citep{vgg} via structural reparameterization (Sec.~\ref{sec:related}) during inference time only. We see that VGG-$16$ + \OursModule{} offers a competitive performance to RepVGG while being simpler at both train and test time.
%
%
%
% \vspace{-2.0ex}
%
%

\vspace{2mm}
\noindent\textbf{E$\mathbf{5}$: \OursModule{} \emph{vs.} DWP~\citep{dwp}.}
%
% We could only find depthwise pooling (DWP) \citep{dwp} as a comparable approach for channel squeezing.
Depth-wise pooling (DWP) is a comparable approach for channel squeezing.
Hence, we trained ResNet-$50$ endowed with DWP. 

As mentioned in Sec.~\ref{sec:channel_squeezing}, eliminating sampling probability predictor $\phi$ from the network removes all the squeezing layers, leading to parameter and accuracy loss.
DWP is an example of this case, which eliminates all the $1\times1$ squeezing layers, facing a loss of accuracy ($1.30\%$), compared to \OursModule{} used for channel squeezing.

Due to the parameter differences in ResNet-$50$ + \OursModule{} and ResNet-$50$ + DWP, we compare the latter with a downscaled variant of ResNet-$50$ + \OursModule{}. 
As a result, \OursModule{} surpasses DWP, verifying our hypothesis that in channel squeezing mode, \OursModule{} preserves the non-linearity that allows for maintaining accuracy.

\subsection{Transfer Learning}
%

%%%
\begin{table}[t]
% \begin{wraptable}[7]{r}[0ex]{0.59\linewidth}
\centering

% \vspace{-3ex}

\caption{\OursModule{} \textit{vs}. ResNet. Transfer learning evaluation for classification (E0) and semantic segmentation (E1) tasks.}
\label{tab:transfer}
\vspace{-1.5ex}

\arrayrulecolor{white!60!black}
%\setlength{\arrayrulewidth}{0.1ex}

%\scriptsize
%\footnotesize
\scriptsize

\resizebox{0.995\columnwidth}{!}{
\setlength{\tabcolsep}{2pt}

\begin{tabular}{c l c c c c}

\toprule

& \multicolumn{1}{c}{Approach} & \#Params & FLOPs~\bdac{accclr} &  \multicolumn{1}{c}{CIFAR-$10$~\buac{accclr}} &  \multicolumn{1}{c}{CIFAR-$100$~\buac{accclr}} \\
\midrule

\multicolumn{1}{c|}{\multirow{2}{*}{}} & \protect\bdota{}~ResNet-$50$~\cite{resnet}  & $25.5$M & $4.12$B & $95.57$  & $81.60$ \\
\rowcolor{rwclr}
\multicolumn{1}{c|}{\multirow{-2}{*}{E$0$}\cellcolor{white}}  & \protect\bdotb{}~\multirow{1}{*}{ResNet-$50$ + \OursModule{}} & $25.5$M & \textbf{$\mathbf{3.18}$B} & $\mathbf{95.67}$  & $\mathbf{82.22}$ \\ 
\midrule
& \multicolumn{1}{c}{Approach} & \#Params & FLOPs~\bdac{accclr} & \multicolumn{2}{c}{CityScapes~\buac{accclr}}\\
\midrule
\multicolumn{1}{c|}{\multirow{2}{*}{}}  & \protect\bdota{}~\multirow{1}{*}{ResNet-$101$} + \cite{pspnet} & $44.5$M & $7.85$B  & \multicolumn{2}{c}{$78.4$} \\
\rowcolor{rwclr}
\multicolumn{1}{c|}{\multirow{-2}{*}{E$1$}\cellcolor{white}} & \protect\bdotb{}~\multirow{1}{*}{ResNet-$101$ + \cite{pspnet} + \OursModule{}}  & $44.5$M & \textbf{$\mathbf{6.05}$B}  &  \multicolumn{2}{c}{$\mathbf{79.1}$}\\

% &  

\bottomrule

\end{tabular}
}
\vspace{-1ex}
\end{table}
% \end{wraptable}
%
%
%%
%
%
% \vspace{2mm}
\noindent\textbf{E$\mathbf{0}$: \OursModule{} transfers better on image classification task.}
To analyze the generalization of \OursModule{} across datasets and tasks, we perform transfer learning from ImageNet to CIFAR-$10$ and CIFAR-$100$. 
Each of the datasets consists of $50$K training and $10$K test images. 
For training, we finetune the models pretrained over ImageNet. 
The training strategy for both datasets remains identical to that of ImageNet except for $200$ epochs. 
From Table~\ref{tab:transfer}, it can be seen that \OursModule{} performs better at lower FLOP requirements.
%
%
% \vspace{-1.75ex}
%

\vspace{2mm}
\noindent\textbf{E$\mathbf{1}$: \OursModule{} transfers better on semantic segmentation task.}
We evaluate \OursModule{} for a challenging task of semantic segmentation. We use a prominent approach \citep{pspnet} and replace the backbone with ResNet-$101$+\OursModule{}. 
Consequently, \OursModule{} outperforms the baseline both in terms of FLOPs and accuracy by $0.7\%$ units mIoU.
% , indicating that \OursModule{} transfers well across tasks and datasets.
%
%
%

% % %%%%
% % %%%
\section{Conclusion}
\label{sec:conc}

In this work, we introduce \OursAcronym{} (\OursModule{}) for dynamic channel sampling. It works by exploiting global spatial context by blending cross-channel information and then picking or mixing channels on \textit{per-pixel basis}. The picked channels can be different for each pixel depending upon the operator selection. This capability allows \OursModule{} to maintain accuracy even by cutting down FLOPs.
\OursModule{} can work as a computationally efficient channel squeezer, can downscale a given model, or function as a dynamic channel pruner. We show that \OursModule{} is easy to plug into the existing ConvNets or even ViT, without altering its structure, and we show that \OursModule{} outperforms state-of-the-art approaches. 

\vspace{2mm}
\noindent\textbf{Limitations.} Currently, our approach is designed for discrete squeezing factors $\zeta$. Future extensions of the proposed approach include developing a more generalized fusion approach that can sample channels at non-integer $\zeta$.

% \section{Acknowledgment}
\vspace{2mm}
\noindent \textbf{Acknowledgment.}
This study was supported by the I-Hub Foundation for Cobotics (IHFC), Technology Innovation Hub of Indian Institute of Technology, Delhi (IIT Delhi) under the project grant IITM/IHFC/IITDELHI/LB/$370$. 
Danuel Kim and Jaesik Park were supported by IITP grant funded by the Korea government (MSIT) (No.$2021$-$0$-$01343$, AI Graduate School Program: Seoul National University, $5\%$) and NRF grant No.$2023$R$1$A$1$C$200781211$ ($95\%$).

% {
% \balance
\small
\bibliographystyle{ieeenat_fullname}
%\balance
\bibliography{bibfile}

\begin{thebibliography}{44}
\providecommand{\natexlab}[1]{#1}
\providecommand{\url}[1]{\texttt{#1}}
\expandafter\ifx\csname urlstyle\endcsname\relax
  \providecommand{\doi}[1]{doi: #1}\else
  \providecommand{\doi}{doi: \begingroup \urlstyle{rm}\Url}\fi

\bibitem[Chen et~al.(2020)Chen, Chen, and Pan]{chen2020storage}
Jianda Chen, Shangyu Chen, and Sinno~Jialin Pan.
\newblock Storage efficient and dynamic flexible runtime channel pruning via
  deep reinforcement learning.
\newblock \emph{Advances in neural information processing systems},
  33:\penalty0 14747--14758, 2020.

\bibitem[Cordts et~al.(2016)Cordts, Omran, Ramos, Rehfeld, Enzweiler, Benenson,
  Franke, Roth, and Schiele]{cityscapes}
Marius Cordts, Mohamed Omran, Sebastian Ramos, Timo Rehfeld, Markus Enzweiler,
  Rodrigo Benenson, Uwe Franke, Stefan Roth, and Bernt Schiele.
\newblock The cityscapes dataset for semantic urban scene understanding.
\newblock In \emph{Proc. of the IEEE Conference on Computer Vision and Pattern
  Recognition (CVPR)}, 2016.

\bibitem[Dai et~al.(2021)Dai, Gieseke, Oehmcke, Wu, and
  Barnard]{attentionalfeaturefusion}
Yimian Dai, Fabian Gieseke, Stefan Oehmcke, Yiquan Wu, and Kobus Barnard.
\newblock Attentional feature fusion.
\newblock In \emph{Proceedings of the IEEE/CVF Winter Conference on
  Applications of Computer Vision}, pages 3560--3569, 2021.

\bibitem[Deng et~al.(2009)Deng, Dong, Socher, Li, Li, and Fei-Fei]{imagenet}
Jia Deng, Wei Dong, Richard Socher, Li-Jia Li, Kai Li, and Li Fei-Fei.
\newblock Imagenet: A large-scale hierarchical image database.
\newblock In \emph{2009 IEEE conference on computer vision and pattern
  recognition}, pages 248--255. Ieee, 2009.

\bibitem[Ding et~al.(2021)Ding, Zhang, Ma, Han, Ding, and Sun]{repvgg}
Xiaohan Ding, Xiangyu Zhang, Ningning Ma, Jungong Han, Guiguang Ding, and Jian
  Sun.
\newblock Repvgg: Making vgg-style convnets great again.
\newblock In \emph{Proceedings of the IEEE/CVF Conference on Computer Vision
  and Pattern Recognition}, pages 13733--13742, 2021.

\bibitem[Dong et~al.(2017)Dong, Huang, Yang, and Yan]{lowcostcolablayer}
Xuanyi Dong, Junshi Huang, Yi Yang, and Shuicheng Yan.
\newblock More is less: A more complicated network with less inference
  complexity.
\newblock In \emph{Proceedings of the IEEE conference on computer vision and
  pattern recognition}, pages 5840--5848, 2017.

\bibitem[flops~counting tool()]{flopcounter}
flops~counting tool.
\newblock \emph{https://github.com/sovrasov/flops-counter.pytorch}.

\bibitem[Gao et~al.(2018)Gao, Zhao, Dudziak, Mullins, and Xu]{fbs}
Xitong Gao, Yiren Zhao, {\L}ukasz Dudziak, Robert Mullins, and Cheng-zhong Xu.
\newblock Dynamic channel pruning: Feature boosting and suppression.
\newblock \emph{arXiv preprint arXiv:1810.05331}, 2018.

\bibitem[Glorot and Bengio(2010)]{xavier}
Xavier Glorot and Yoshua Bengio.
\newblock Understanding the difficulty of training deep feedforward neural
  networks.
\newblock In \emph{Proceedings of the thirteenth international conference on
  artificial intelligence and statistics}, pages 249--256. JMLR Workshop and
  Conference Proceedings, 2010.

\bibitem[Han et~al.(2020)Han, Wang, Tian, Guo, Xu, and Xu]{ghostnet}
Kai Han, Yunhe Wang, Qi Tian, Jianyuan Guo, Chunjing Xu, and Chang Xu.
\newblock Ghostnet: More features from cheap operations.
\newblock In \emph{Proceedings of the IEEE/CVF conference on computer vision
  and pattern recognition}, pages 1580--1589, 2020.

\bibitem[He et~al.(2016)He, Zhang, Ren, and Sun]{resnet}
Kaiming He, Xiangyu Zhang, Shaoqing Ren, and Jian Sun.
\newblock Deep residual learning for image recognition.
\newblock In \emph{Proceedings of the IEEE conference on computer vision and
  pattern recognition}, pages 770--778, 2016.

\bibitem[He et~al.(2018{\natexlab{a}})He, Kang, Dong, Fu, and
  Yang]{softfilterpruning}
Yang He, Guoliang Kang, Xuanyi Dong, Yanwei Fu, and Yi Yang.
\newblock Soft filter pruning for accelerating deep convolutional neural
  networks.
\newblock \emph{arXiv preprint arXiv:1808.06866}, 2018{\natexlab{a}}.

\bibitem[He et~al.(2018{\natexlab{b}})He, Lin, Liu, Wang, Li, and Han]{automl}
Yihui He, Ji Lin, Zhijian Liu, Hanrui Wang, Li-Jia Li, and Song Han.
\newblock Amc: Automl for model compression and acceleration on mobile devices.
\newblock In \emph{Proceedings of the European conference on computer vision
  (ECCV)}, pages 784--800, 2018{\natexlab{b}}.

\bibitem[Howard et~al.(2017)Howard, Zhu, Chen, Kalenichenko, Wang, Weyand,
  Andreetto, and Adam]{mobilenetv1}
Andrew~G Howard, Menglong Zhu, Bo Chen, Dmitry Kalenichenko, Weijun Wang,
  Tobias Weyand, Marco Andreetto, and Hartwig Adam.
\newblock Mobilenets: Efficient convolutional neural networks for mobile vision
  applications.
\newblock \emph{arXiv preprint arXiv:1704.04861}, 2017.

\bibitem[Hu et~al.(2018)Hu, Shen, and Sun]{senet}
Jie Hu, Li Shen, and Gang Sun.
\newblock Squeeze-and-excitation networks.
\newblock In \emph{Proceedings of the IEEE conference on computer vision and
  pattern recognition}, pages 7132--7141, 2018.

\bibitem[Hua et~al.(2019)Hua, Zhou, De~Sa, Zhang, and Suh]{channelgating}
Weizhe Hua, Yuan Zhou, Christopher~M De~Sa, Zhiru Zhang, and G~Edward Suh.
\newblock Channel gating neural networks.
\newblock \emph{Advances in Neural Information Processing Systems}, 32, 2019.

\bibitem[Hussain and Hesheng(2019)]{dwp}
Abid Hussain and Wang Hesheng.
\newblock Depth-wise pooling: A parameter-less solution for channel reduction
  of feature-map in convolutional neural network.
\newblock In \emph{2019 IEEE International Conference on Real-time Computing
  and Robotics (RCAR)}, pages 299--304. IEEE, 2019.

\bibitem[Ioffe and Szegedy(2015)]{bn}
Sergey Ioffe and Christian Szegedy.
\newblock Batch normalization: Accelerating deep network training by reducing
  internal covariate shift.
\newblock In \emph{International conference on machine learning}, pages
  448--456. PMLR, 2015.

\bibitem[Jeong et~al.(2022)Jeong, Shin, Lee, Choy, Anandkumar, Cho, and
  Park]{perfception}
Yoonwoo Jeong, Seungjoo Shin, Junha Lee, Chris Choy, Anima Anandkumar, Minsu
  Cho, and Jaesik Park.
\newblock Perfception: Perception using radiance fields.
\newblock In \emph{Thirty-sixth Conference on Neural Information Processing
  Systems Datasets and Benchmarks Track}, 2022.

\bibitem[Krizhevsky et~al.(2012)Krizhevsky, Sutskever, and Hinton]{alexnet}
Alex Krizhevsky, Ilya Sutskever, and Geoffrey~E Hinton.
\newblock Imagenet classification with deep convolutional neural networks.
\newblock In \emph{Advances in neural information processing systems}, pages
  1097--1105, 2012.

\bibitem[Kumar and Behera(2019)]{deepquick}
Ashish Kumar and Laxmidhar Behera.
\newblock Semi supervised deep quick instance detection and segmentation.
\newblock In \emph{2019 International Conference on Robotics and Automation
  (ICRA)}, pages 8325--8331. IEEE, 2019.

\bibitem[Kumar et~al.(2020)Kumar, Vohra, Prakash, and Behera]{towards}
Ashish Kumar, Mohit Vohra, Ravi Prakash, and Laxmidhar Behera.
\newblock Towards deep learning assisted autonomous uavs for manipulation tasks
  in gps-denied environments.
\newblock In \emph{2020 IEEE/RSJ International Conference on Intelligent Robots
  and Systems (IROS)}, pages 1613--1620. IEEE, 2020.

\bibitem[Li et~al.(2016)Li, Kadav, Durdanovic, Samet, and Graf]{filterpruning}
Hao Li, Asim Kadav, Igor Durdanovic, Hanan Samet, and Hans~Peter Graf.
\newblock Pruning filters for efficient convnets.
\newblock 2016.

\bibitem[Li et~al.(2019)Li, Wang, Hu, and Yang]{sknet}
Xiang Li, Wenhai Wang, Xiaolin Hu, and Jian Yang.
\newblock Selective kernel networks.
\newblock In \emph{Proceedings of the IEEE/CVF conference on computer vision
  and pattern recognition}, pages 510--519, 2019.

\bibitem[Lin et~al.(2017)Lin, Rao, Lu, and Zhou]{runtimepruning}
Ji Lin, Yongming Rao, Jiwen Lu, and Jie Zhou.
\newblock Runtime neural pruning.
\newblock \emph{Advances in neural information processing systems}, 30, 2017.

\bibitem[Liu et~al.(2023)Liu, Peng, Zheng, Yang, Hu, and Yuan]{efficientvit}
Xinyu Liu, Houwen Peng, Ningxin Zheng, Yuqing Yang, Han Hu, and Yixuan Yuan.
\newblock Efficientvit: Memory efficient vision transformer with cascaded group
  attention.
\newblock In \emph{Proceedings of the IEEE/CVF Conference on Computer Vision
  and Pattern Recognition}, pages 14420--14430, 2023.

\bibitem[Loshchilov and Hutter(2016)]{cosineanneal}
Ilya Loshchilov and Frank Hutter.
\newblock Sgdr: Stochastic gradient descent with warm restarts.
\newblock \emph{arXiv preprint arXiv:1608.03983}, 2016.

\bibitem[Luo et~al.(2017)Luo, Wu, and Lin]{thinetconv}
Jian-Hao Luo, Jianxin Wu, and Weiyao Lin.
\newblock Thinet: A filter level pruning method for deep neural network
  compression.
\newblock In \emph{Proceedings of the IEEE international conference on computer
  vision}, pages 5058--5066, 2017.

\bibitem[Ma et~al.(2018)Ma, Zhang, Zheng, and Sun]{shufflenetv2}
Ningning Ma, Xiangyu Zhang, Hai-Tao Zheng, and Jian Sun.
\newblock Shufflenet v2: Practical guidelines for efficient cnn architecture
  design.
\newblock In \emph{Proceedings of the European conference on computer vision
  (ECCV)}, pages 116--131, 2018.

\bibitem[Park et~al.(2023)Park, Kim, Kim, Choi, and Lee]{park2023dynamic}
Jun-Hyung Park, Yeachan Kim, Junho Kim, Joon-Young Choi, and SangKeun Lee.
\newblock Dynamic structure pruning for compressing cnns.
\newblock \emph{arXiv preprint arXiv:2303.09736}, 2023.

\bibitem[Paszke et~al.(2019)Paszke, Gross, Massa, Lerer, Bradbury, Chanan,
  Killeen, Lin, Gimelshein, Antiga, et~al.]{pytorch}
Adam Paszke, Sam Gross, Francisco Massa, Adam Lerer, James Bradbury, Gregory
  Chanan, Trevor Killeen, Zeming Lin, Natalia Gimelshein, Luca Antiga, et~al.
\newblock Pytorch: An imperative style, high-performance deep learning library.
\newblock \emph{Advances in neural information processing systems}, 32, 2019.

\bibitem[Radosavovic et~al.(2020)Radosavovic, Kosaraju, Girshick, He, and
  Doll{\'a}r]{regnet}
Ilija Radosavovic, Raj~Prateek Kosaraju, Ross Girshick, Kaiming He, and Piotr
  Doll{\'a}r.
\newblock Designing network design spaces.
\newblock In \emph{Proceedings of the IEEE/CVF conference on computer vision
  and pattern recognition}, pages 10428--10436, 2020.

\bibitem[Ren et~al.(2015)Ren, He, Girshick, and Sun]{fasterrcnn}
Shaoqing Ren, Kaiming He, Ross Girshick, and Jian Sun.
\newblock Faster r-cnn: Towards real-time object detection with region proposal
  networks.
\newblock \emph{Advances in neural information processing systems},
  28:\penalty0 91--99, 2015.

\bibitem[Sandler et~al.(2018)Sandler, Howard, Zhu, Zhmoginov, and
  Chen]{mobilenetv2}
Mark Sandler, Andrew Howard, Menglong Zhu, Andrey Zhmoginov, and Liang-Chieh
  Chen.
\newblock Mobilenetv2: Inverted residuals and linear bottlenecks.
\newblock In \emph{Proceedings of the IEEE conference on computer vision and
  pattern recognition}, pages 4510--4520, 2018.

\bibitem[Selvaraju et~al.(2017)Selvaraju, Cogswell, Das, Vedantam, Parikh, and
  Batra]{gradcam}
Ramprasaath~R Selvaraju, Michael Cogswell, Abhishek Das, Ramakrishna Vedantam,
  Devi Parikh, and Dhruv Batra.
\newblock Grad-cam: Visual explanations from deep networks via gradient-based
  localization.
\newblock In \emph{Proceedings of the IEEE international conference on computer
  vision}, pages 618--626, 2017.

\bibitem[Sifre and Mallat()]{depthwise}
L Sifre and S Mallat.
\newblock Rigid-motion scattering for image classification. arxiv 2014.
\newblock \emph{arXiv preprint arXiv:1403.1687}.

\bibitem[Simonyan and Zisserman(2014)]{vgg}
Karen Simonyan and Andrew Zisserman.
\newblock Very deep convolutional networks for large-scale image recognition.
\newblock \emph{CoRR}, abs/1409.1556, 2014.

\bibitem[Tan and Le(2019)]{efficientnet}
Mingxing Tan and Quoc Le.
\newblock Efficientnet: Rethinking model scaling for convolutional neural
  networks.
\newblock In \emph{International Conference on Machine Learning}, pages
  6105--6114. PMLR, 2019.

\bibitem[Tang et~al.(2021)Tang, Wang, Xu, Deng, Xu, Tao, and
  Xu]{tang2021manifold}
Yehui Tang, Yunhe Wang, Yixing Xu, Yiping Deng, Chao Xu, Dacheng Tao, and Chang
  Xu.
\newblock Manifold regularized dynamic network pruning.
\newblock In \emph{Proceedings of the IEEE/CVF Conference on Computer Vision
  and Pattern Recognition}, pages 5018--5028, 2021.

\bibitem[Woo et~al.(2018)Woo, Park, Lee, and Kweon]{cbam}
Sanghyun Woo, Jongchan Park, Joon-Young Lee, and In~So Kweon.
\newblock Cbam: Convolutional block attention module.
\newblock In \emph{Proceedings of the European conference on computer vision
  (ECCV)}, pages 3--19, 2018.

\bibitem[Zhang et~al.(2022)Zhang, Wu, Zhang, Zhu, Lin, Zhang, Sun, He, Mueller,
  Manmatha, et~al.]{resnest}
Hang Zhang, Chongruo Wu, Zhongyue Zhang, Yi Zhu, Haibin Lin, Zhi Zhang, Yue
  Sun, Tong He, Jonas Mueller, R Manmatha, et~al.
\newblock Resnest: Split-attention networks.
\newblock In \emph{Proceedings of the IEEE/CVF Conference on Computer Vision
  and Pattern Recognition}, pages 2736--2746, 2022.

\bibitem[Zhang et~al.(2018)Zhang, Zhou, Lin, and Sun]{shufflenetv1}
Xiangyu Zhang, Xinyu Zhou, Mengxiao Lin, and Jian Sun.
\newblock Shufflenet: An extremely efficient convolutional neural network for
  mobile devices.
\newblock In \emph{Proceedings of the IEEE conference on computer vision and
  pattern recognition}, pages 6848--6856, 2018.

\bibitem[Zhao et~al.(2017)Zhao, Shi, Qi, Wang, and Jia]{pspnet}
Hengshuang Zhao, Jianping Shi, Xiaojuan Qi, Xiaogang Wang, and Jiaya Jia.
\newblock Pyramid scene parsing network.
\newblock In \emph{Proceedings of the IEEE conference on computer vision and
  pattern recognition}, pages 2881--2890, 2017.

\bibitem[Zhuang et~al.(2018)Zhuang, Tan, Zhuang, Liu, Guo, Wu, Huang, and
  Zhu]{discriminationpruning}
Zhuangwei Zhuang, Mingkui Tan, Bohan Zhuang, Jing Liu, Yong Guo, Qingyao Wu,
  Junzhou Huang, and Jinhui Zhu.
\newblock Discrimination-aware channel pruning for deep neural networks.
\newblock \emph{Advances in neural information processing systems}, 31, 2018.

\end{thebibliography}
% }

\clearpage

\setcounter{section}{0}
\section{\OursModule{} Instantiation}
Figure~\ref{fig:mipcasns} shows how one can use \OursModule{} in different network architectures and for different tasks.

\section{Difference with Existing Modules}
\begin{figure*}[t]
%\begin{wrapfigure}[14]{r}[0ex]{0.48\linewidth}

\vspace{8ex}

\centering
\begin{tikzpicture}

\FPeval{\xshfta}{0}
\FPeval{\xshftb}{22.45}
\FPeval{\xshftc}{46.35}
\FPeval{\xshftd}{67.15}
\FPeval{\xshfte}{89.15}

% \FPeval{\xshftb}{11.45}
% \FPeval{\xshftc}{24.0}
% \FPeval{\xshftd}{33.4}

\FPeval{\yshfta}{0-0.0}
\FPeval{\yshftb}{0-0.0}
\FPeval{\yshftc}{0-0.0}
\FPeval{\yshftd}{0-0.0}

% \FPeval{\scal}{0.35}
\FPeval{\scal}{0.65}

\colorlet{clr}{green!40!blue}

\colorlet{gpdclr}{blue!50!cyan}
\colorlet{sigmoiddclr}{blue!20!green}
\colorlet{reluclr}{blue!20!yellow}

\colorlet{fusionclr}{black!10!green}

\colorlet{convdclr}{blue!20!magenta}
\colorlet{convclr}{white!80!gray}
\colorlet{conv3clr}{white!60!clr}

\colorlet{dcsclr}{white!30!brown}

\colorlet{reluclr}{white!80!black}
\colorlet{eltclr}{white!50!cyan}

\colorlet{drawclr}{white!99!black}

\colorlet{arrowclr}{white!40!black}
\colorlet{boxclr}{white!80!black}

\colorlet{outerclr}{white!80!black}

\FPeval{\cornerradii}{0.2}

\FPeval{\arrowlw}{0.2}

 \FPeval{\linkcornerradii}{0.5}

\node (na) [xshift = \xshfta ex, yshift=\yshfta ex, scale=\scal]{
\tikz{
\node (outer) [draw=outerclr, rounded corners=0.9ex, xshift=-3.25ex, yshift=-23.0ex, minimum width=21ex, minimum height=62ex]{};
\node (gp) [draw=gpdclr, fill=convclr, rounded corners=\cornerradii ex, xshift=-5ex, yshift=-0ex]{Global Pooling};
\node (conv1) [draw=gpdclr, fill=convclr, rounded corners=\cornerradii ex, xshift=-5ex, yshift=-8ex]{Conv$1\times 1$};
\node (relu) [draw=reluclr, fill=convclr, rounded corners=\cornerradii ex, xshift=-5ex, yshift=-13.0ex]{ReLU};
\node (conv2) [draw=gpdclr, fill=convclr, rounded corners=\cornerradii ex, xshift=-5ex, yshift=-21.5ex]{Conv$1\times 1$};
\node (sigmoid) [draw=sigmoiddclr, fill=convclr, rounded corners=\cornerradii ex, xshift=-5ex, yshift=-26.8ex]{Sigmoid};
\node (bmul) [draw=none, xshift=5ex, yshift=-48.8ex]{\tikz{\node[draw=black, fill=white!70!gray, circle, scale=1.1]{};\node[scale=1.2]{$\times$};}};
\node (ip)[fill=black, circle, yshift=7ex, scale=0.4]{};
\draw [->, line width=\arrowlw ex, rounded corners=\linkcornerradii ex] (ip) -| (gp);
\draw [->, line width=\arrowlw ex, rounded corners=\linkcornerradii ex] (gp) -- (conv1);
\draw [->, line width=\arrowlw ex, rounded corners=\linkcornerradii ex] (conv1) -- (relu);
\draw [->, line width=\arrowlw ex, rounded corners=\linkcornerradii ex] (relu) -- (conv2);
\draw [->, line width=\arrowlw ex, rounded corners=\linkcornerradii ex] (conv2) -- (sigmoid);
\draw [->, line width=\arrowlw ex, rounded corners=\linkcornerradii ex] (sigmoid) |- ($(bmul.west)+(1.5ex, 0ex)$);
\draw [->, line width=\arrowlw ex, rounded corners=\linkcornerradii ex] (ip) -| ($(bmul.north)-(0.0ex, 1.5ex)$);
\draw [->, line width=\arrowlw ex, rounded corners=\linkcornerradii ex] ($(bmul.south)+(0ex,1.5ex)$) -- ($(bmul.south)-(0ex,2ex)$);
\node () [rounded corners, xshift=-3ex, yshift=10.5ex,scale=1.0]{SE \cite{senet}};
}};

\node (nb) [xshift = \xshftb ex, yshift=\yshftb ex, scale=\scal]{
\tikz{
\node (outer) [draw=outerclr, rounded corners=0.9ex, xshift=0.0ex, yshift=-23.0ex, minimum width=40.5ex, minimum height=62ex]{};
\node (gp1) [draw=gpdclr, fill=convclr, rounded corners=\cornerradii ex, xshift=-9ex, yshift=-0ex]{\shortstack{Global Pooling \\ (Max)}};
\node (gp2) [draw=gpdclr, fill=convclr, rounded corners=\cornerradii ex, xshift=9ex, yshift=-0ex]{\shortstack{Global Pooling \\ (Avg)}};
\node (mlp1) [draw=gpdclr, fill=convclr, rounded corners=\cornerradii ex, xshift=-9ex, yshift=-6ex]{MLP};
\node (mlp2) [draw=gpdclr, fill=convclr, rounded corners=\cornerradii ex, xshift=9ex, yshift=-6ex]{MLP};
\node (sum) [draw=none, xshift=0ex, yshift=-9.0ex]{\tikz{\node[draw=black, fill=white!70!gray, circle, scale=1.1]{};\node[scale=1.2]{$+$};}};
\node (sigmoid1) [draw=sigmoiddclr, fill=convclr, rounded corners=\cornerradii ex, xshift=0ex, yshift=-14.0ex]{Sigmoid};
\node (bmul1) [draw=none, xshift=0ex, yshift=-19.0ex]{\tikz{\node[draw=black, fill=white!70!gray, circle, scale=1.1]{};\node[scale=1.2]{$\times$};}};
\node (cgp1) [draw=gpdclr, fill=convclr, rounded corners=\cornerradii ex, xshift=-12ex, yshift=-27.0ex]{\shortstack{Channel-wise \\ Global Pooling \\ (Max)}};
\node (cgp2) [draw=gpdclr, fill=convclr, rounded corners=\cornerradii ex, xshift=12ex, yshift=-27.0ex]{\shortstack{Channel-wise \\ Global Pooling \\ (Avg)}};
\node (concat) [draw=sigmoiddclr, fill=convclr, rounded corners=\cornerradii ex, xshift=0ex, yshift=-34.8ex]{Concat};
\node (convs) [draw=gpdclr, fill=convclr, rounded corners=\cornerradii ex, xshift=0ex, yshift=-39.6ex]{Conv $1 \times 1$};
\node (sigmoid2) [draw=sigmoiddclr, fill=convclr, rounded corners=\cornerradii ex, xshift=0ex, yshift=-44.4ex]{Sigmoid};
\node (bmul2) [draw=none, xshift=0ex, yshift=-49.5ex]{\tikz{\node[draw=black, fill=white!70!gray, circle, scale=1.1]{};\node[scale=1.2]{$\times$};}};
\node (ip)[fill=black, circle, yshift=7ex, scale=0.4]{};
\draw [->, line width=\arrowlw ex, rounded corners=\linkcornerradii ex] (ip) -| (gp1);
\draw [->, line width=\arrowlw ex, rounded corners=\linkcornerradii ex] (ip) -| (gp2);
\draw [->, line width=\arrowlw ex, rounded corners=\linkcornerradii ex] (gp1) -- (mlp1);
\draw [->, line width=\arrowlw ex, rounded corners=\linkcornerradii ex] (gp2) -- (mlp2);
\draw [->, line width=\arrowlw ex, rounded corners=\linkcornerradii ex] (mlp1) |- ($(sum.west)+(1.5ex,0ex)$);
\draw [->, line width=\arrowlw ex, rounded corners=\linkcornerradii ex] (mlp2) |- ($(sum.east)-(1.5ex,0ex)$);
\draw [->, line width=\arrowlw ex, rounded corners=\linkcornerradii ex] ($(sum.south)+(0ex,1.5ex)$) -- (sigmoid1);
\draw [->, line width=\arrowlw ex, rounded corners=\linkcornerradii ex]  (sigmoid1) -- ($(bmul1.north)-(0ex,1.5ex)$);
\draw [->, line width=\arrowlw ex, rounded corners=\linkcornerradii ex] (cgp1) |- (concat);
\draw [->, line width=\arrowlw ex, rounded corners=\linkcornerradii ex] (cgp2) |- (concat);
\draw [->, line width=\arrowlw ex, rounded corners=\linkcornerradii ex] (concat) -- (convs);
\draw [->, line width=\arrowlw ex, rounded corners=\linkcornerradii ex] (convs) -- (sigmoid2);
\draw [->, line width=\arrowlw ex, rounded corners=\linkcornerradii ex] (sigmoid2) -- ($(bmul2.north)-(0.0ex, 1.5ex)$);
\draw [->, line width=\arrowlw ex, rounded corners=\linkcornerradii ex] ($(bmul2.south)+(0.0ex, 1.5ex)$) -- ($(bmul2.south)-(0.0ex, 2ex)$);
\draw [->, line width=\arrowlw ex, rounded corners=\linkcornerradii ex] (ip) -- ($(ip.west)-(18ex, 0ex)$) |- ($(bmul1.west)-(-1.5ex, 0ex)$);
\draw [->, line width=\arrowlw ex, rounded corners=\linkcornerradii ex] ($(bmul1.south)+(0.0ex, 1.4ex)$) -- ($(bmul1.south)+(0.0ex, 0.2ex)$) -| (cgp1);
\draw [->, line width=\arrowlw ex, rounded corners=\linkcornerradii ex] ($(bmul1.south)+(0.0ex, 1.4ex)$) -- ($(bmul1.south)+(0.0ex, 0.2ex)$) -| (cgp2);
\draw [->, line width=\arrowlw ex, rounded corners=\linkcornerradii ex] ($(bmul1.south)+(0.0ex, 1.4ex)$) -- ($(bmul1.south)-(0.0ex, 10.8ex)$) -- ($(bmul1.south)-(-7.0ex, 10.8ex)$) |- ($(bmul2.east)-(1.0ex, 0ex)$) ;
\node () [rounded corners, xshift=0ex, yshift=10.5ex,scale=1.0]{CBAM \cite{cbam}};
}};

\node (nc) [xshift = \xshftc ex, yshift=\yshftc ex, scale=\scal]{
\tikz{
\node (outer) [draw=outerclr, rounded corners=0.9ex, xshift=-0.5ex, yshift=-23.0ex, minimum width=27ex, minimum height=62ex]{};
\node (gp) [draw=gpdclr, fill=convclr, rounded corners=\cornerradii ex, xshift=-5ex, yshift=-0ex]{Global Pooling};
\node (conv1) [draw=gpdclr, fill=convclr, rounded corners=\cornerradii ex, xshift=-5ex, yshift=-9ex]{Conv$1\times 1$};
\node (sigmoid) [draw=sigmoiddclr, fill=convclr, rounded corners=\cornerradii ex, xshift=-5ex, yshift=-18.5ex]{Sigmoid};
\node (topk) [draw=sigmoiddclr, fill=convclr, rounded corners=\cornerradii ex, xshift=8ex, yshift=-18.5ex]{Top-K};
\node (conv2) [draw=convdclr, fill=convclr, rounded corners=\cornerradii ex, xshift=5ex, yshift=-26.5ex]{Conv$3\times3$};
\node (bn) [draw=sigmoiddclr, fill=convclr, rounded corners=\cornerradii ex, xshift=5ex, yshift=-34.0ex]{BN};
\node (bmul) [draw=none, xshift=5ex, yshift=-40.8ex]{\tikz{\node[draw=black, fill=white!70!gray, circle, scale=1.1]{};\node[scale=1.2]{$\times$};}};
\node (relu) [draw=reluclr, fill=convclr, rounded corners=\cornerradii ex, xshift=5ex, yshift=-47.5ex]{ReLU};
\node (ip)[fill=black, circle, yshift=7ex, scale=0.4]{};
\draw [->, line width=\arrowlw ex, rounded corners=\linkcornerradii ex] (ip) -| (gp);
\draw [->, line width=\arrowlw ex, rounded corners=\linkcornerradii ex] (gp) -- (conv1);
\draw [->, line width=\arrowlw ex, rounded corners=\linkcornerradii ex] (conv1) -- (sigmoid);
\draw [->, line width=\arrowlw ex, rounded corners=\linkcornerradii ex] (sigmoid) -- (topk);
\draw [->, line width=\arrowlw ex, rounded corners=\linkcornerradii ex] (ip) -| (topk);
\draw [->, line width=\arrowlw ex, rounded corners=\linkcornerradii ex] (topk) -- ($(conv2.north)+(3ex,0ex)$);
\draw [->, line width=\arrowlw ex, rounded corners=\linkcornerradii ex] (conv2) -- (bn);
\draw [->, line width=\arrowlw ex, rounded corners=\linkcornerradii ex] (bn) -- ($(bmul.north)-(0ex,1.4ex)$);
\draw [->, line width=\arrowlw ex, rounded corners=\linkcornerradii ex] (sigmoid) |- ($(bmul.west)+(1.5ex,0ex)$);
\draw [->, line width=\arrowlw ex, rounded corners=\linkcornerradii ex] ($(bmul.south)+(0ex,1.3ex)$) -- (relu);
\draw [->, line width=\arrowlw ex, rounded corners=\linkcornerradii ex] ($(relu.south)$) -- ($(relu.south)-(0ex,3.5ex)$);
\node () [rounded corners, xshift=-1ex, yshift=10.5ex,scale=1.0]{FBS \cite{fbs}};
}};

\node (nd) [xshift = \xshftd ex, yshift=\yshftd ex, scale=\scal]{
\tikz{
\node (outer) [draw=outerclr, rounded corners=0.9ex, xshift=-2.3ex, yshift=-23.0ex, minimum width=22.5ex, minimum height=62ex]{};
\node (gp) [draw=gpdclr, fill=convclr, rounded corners=\cornerradii ex, xshift=-5ex, yshift=-10ex]{Global Pooling};
\node (conv) [draw=gpdclr, fill=convclr, rounded corners=\cornerradii ex, xshift=-5ex, yshift=-25ex]{Conv$1\times 1$};
\node (sigmoid) [draw=sigmoiddclr, fill=convclr, rounded corners=\cornerradii ex, xshift=-5ex, yshift=-32.5ex]{Sigmoid};
\node (fusion) [draw=fusionclr, fill=white!80!fusionclr,rounded corners=\cornerradii ex,  xshift=-0ex,yshift=-47.5ex]{Channel Fusion};
\node (ip)[fill=black, circle, yshift=7ex, scale=0.4]{};
\draw [->, line width=\arrowlw ex, rounded corners=\linkcornerradii ex] (ip) -| (gp);
\draw [->, line width=\arrowlw ex, rounded corners=\linkcornerradii ex] (gp) -- (conv);
\draw [->, line width=\arrowlw ex, rounded corners=\linkcornerradii ex] (conv) -- (sigmoid);
\draw [->, line width=\arrowlw ex, rounded corners=\linkcornerradii ex] (sigmoid) -- ($(fusion.north)-(5.0ex,0ex)$);
\draw [->, line width=\arrowlw ex, rounded corners=\linkcornerradii ex] (ip) -| ($(fusion.north)+(4ex,0ex)$);
\draw [->, line width=\arrowlw ex, rounded corners=\linkcornerradii ex] ($(fusion.south)$) -- ($(fusion.south)-(0ex,4ex)$);
\node () [rounded corners, xshift=-2ex, yshift=10.5ex,scale=1.0]{\OursModule{}};
}};

\FPeval{\bw}{10}
\FPeval{\bh}{4}

\colorlet{clr1}{blue!40!white}
\colorlet{clr2}{green!40!white}
\colorlet{clr3}{orange!40!white}
\colorlet{clr4}{gray!40!white}

\node (gc) [xshift=\xshfte ex, scale=0.6]
{
\tikz{
\node (ip)[]{\tikz{
\node [draw=black,fill=clr1,minimum width=\bw ex, minimum height=\bh ex, xshift=0ex, yshift=-0* \bh ex]{ch = N};
\node [draw=black,fill=clr2,minimum width=\bw ex, minimum height=\bh ex, xshift=0ex, yshift=-1* \bh ex]{ch = N};
\node [draw=black,fill=clr3,minimum width=\bw ex, minimum height=\bh ex, xshift=0ex, yshift=-2* \bh ex]{ch = N};
\node [draw=black,fill=clr4,minimum width=\bw ex, minimum height=\bh ex, xshift=0ex, yshift=-3* \bh ex]{ch = N};
}};
\node (op)[xshift = 25ex]{\tikz{
\node [draw=black,fill=clr1,minimum width=\bw ex, minimum height=\bh ex, xshift=0ex, yshift=-0* \bh ex]{ch = P};
\node [draw=black,fill=clr2,minimum width=\bw ex, minimum height=\bh ex, xshift=0ex, yshift=-1* \bh ex]{ch = P};
\node [draw=black,fill=clr3,minimum width=\bw ex, minimum height=\bh ex, xshift=0ex, yshift=-2* \bh ex]{ch = P};
\node [draw=black,fill=clr4,minimum width=\bw ex, minimum height=\bh ex, xshift=0ex, yshift=-3* \bh ex]{ch = P};
}};
\draw [->] (ip) -- (op) node [xshift = -13ex, yshift=4ex]{\shortstack{Group \\ Convolution}};
\node (ipt)[xshift = 0ex, yshift=11.5ex]{\shortstack{Input \\ $\in \mathbb{R}^{(g \times N) \times H \times W}$}};
\node (opt)[xshift = 25ex, yshift=11.5ex]{\shortstack{Output \\ $\in \mathbb{R}^{(g \times P) \times H \times W}$}};
}
};

%\node (a) [xshift=\xshfta ex+ 1ex, yshift=-8ex, scale=0.9]{(a)};
%\node (b) [xshift=\xshftb ex+ 0ex, yshift=-8ex, scale=0.9]{(b)};
%\node (c) [xshift=\xshftc ex+ 1ex, yshift=-8ex, scale=0.9]{(c)};

%  
\end{tikzpicture}
\subfloat{\label{fig:mip}}
\subfloat{\label{fig:se}}
\subfloat{\label{fig:cbam}}
\subfloat{\label{fig:fbs}}

\vspace{-1.0ex}
\caption{\OursModule{} vs existing modules: SE~\cite{senet}, CBAM~\cite{cbam}, FBS~\cite{fbs}, and Group convolution~\cite{shufflenetv1, shufflenetv2}.}
%
%\vspace{-2.5ex}
%
\label{fig:dcsvsothers}
\end{figure*}
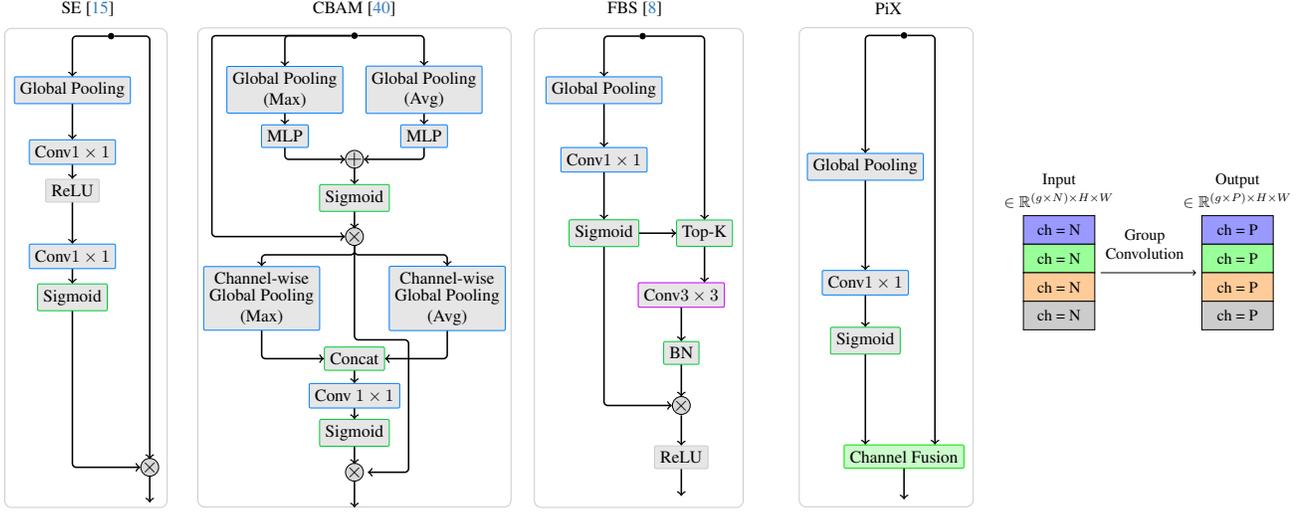

Figure~\ref{fig:dcsvsothers} shows visual differences with the existing modules which aims for accuracy improvement and dynamic pruning approaches.
\section{Computational Complexity}
\label{sec:compute}
We show how \OursModule{} achieves computationally efficient channel sampling. However, for better understanding, we first discuss the FLOPs of different kinds of layers.
\subsection{Convolution}
Consider a convolution layer having $N$ kernels and an input feature map $X \in \mathbb{R}^{C \times H \times W}$. The size of each kernel can be given by $C \times k \times k$. FLOPs for convolution operation is determined using Fusion-Multi-Addition (FMA) instructions. Therefore, the computational demands of a convolution layer can be given as follows:
\begin{equation}
\text{\#FLOPs} = H \times W \times N \times C \times k \times K
\end{equation}
 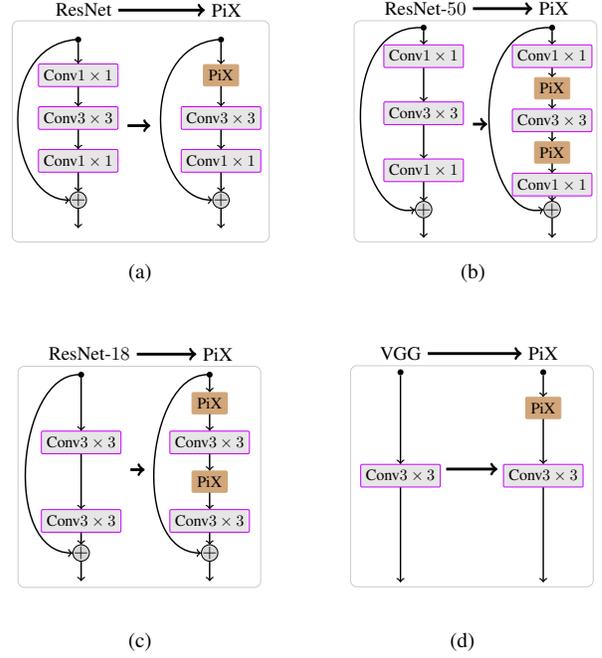
\begin{figure}[h]
%\begin{wrapfigure}[12]{r}[0ex]{0.60\linewidth}

% \vspace{5ex}

\subfloat{\label{fig:dcs_as_cs_resnet50}}
\subfloat{\label{fig:dcs_as_nds_dcp_resnet50}}
\subfloat{\label{fig:dcs_as_nds_dcp_resnet18}}
\subfloat{\label{fig:dcs_as_nds_dcp_vgg16}}

\centering
\begin{tikzpicture}

% \FPeval{\xshfta}{0}
% \FPeval{\xshftb}{12.9}
% \FPeval{\xshftc}{25.7}
% \FPeval{\xshftd}{38.90}

% \FPeval{\xshfta}{0}
% \FPeval{\xshftb}{21.9}
% \FPeval{\xshftc}{43.7}
% \FPeval{\xshftd}{65.90}

\FPeval{\xshfta}{0}
\FPeval{\xshftb}{28.0}
\FPeval{\xshftc}{0}
\FPeval{\xshftd}{28.9}

\FPeval{\yshfta}{0-0.0}
\FPeval{\yshftb}{0-0.0}
\FPeval{\yshftc}{0-29.0}
\FPeval{\yshftd}{0-29.0}

\FPeval{\scal}{0.6}

\colorlet{clr}{green!40!blue}

\colorlet{convdclr}{blue!20!magenta}
\colorlet{convclr}{white!80!gray}
\colorlet{conv3clr}{white!60!clr}

\colorlet{dcsclr}{white!30!brown}

\colorlet{reluclr}{white!80!black}
\colorlet{eltclr}{white!50!cyan}

\colorlet{drawclr}{white!99!black}

\colorlet{arrowclr}{white!40!black}
\colorlet{boxclr}{white!80!black}

\colorlet{outerclr}{white!80!black}

\FPeval{\cornerradii}{0.2}

\FPeval{\arrowlw}{0.2}

 \node (na) [xshift = \xshfta ex, yshift=\yshfta ex, scale=\scal]{
\tikz{
\node (outer) [draw=outerclr, rounded corners=0.9ex, xshift=12ex, yshift=0.2ex, minimum width=36ex, minimum height=31ex]{};
\node (resblock) []{
\tikz{
\node (convsqz) [draw=convdclr, fill=convclr, rounded corners=\cornerradii ex]{Conv$1\times1$};
\node (convce) [draw=convdclr, fill=convclr, rounded corners=\cornerradii ex, yshift=-6ex]{Conv$3\times3$};
\node (convexp) [draw=convdclr, fill=convclr, rounded corners=\cornerradii ex, yshift=-12ex]{Conv$1\times1$};
\node (sum) [draw=none, yshift=-17.5ex]{\tikz{\node[draw=black, fill=white!70!gray, circle, scale=1.1]{};\node[scale=1.2]{$+$};}};
\node (ip)[fill=black, circle, yshift=5ex, scale=0.4]{};
\draw [->, line width=\arrowlw ex] (ip) -- (convsqz);
\draw [->, line width=\arrowlw ex] (convsqz) -- (convce);
\draw [->, line width=\arrowlw ex] (convce) -- (convexp);
\draw [->, line width=\arrowlw ex] (convexp) -- ($(sum.north)-(0ex,1.4ex)$);
\draw [->, line width=\arrowlw ex] ($(sum.south)+(0ex,1.4ex)$) -- ($(sum.south)-(0ex,1.4ex)$);
\draw [->, line width=\arrowlw ex] ($(ip.west)+(0.1ex,0ex)$) .. controls (-12ex, 4ex) and (-10ex, -17.5ex) .. ($(sum.west)+(1.5ex,0ex)$)
}};
\node (dcsblock) [xshift=20ex]{
\tikz{
\node (convsqz) [draw=dcsclr, fill=dcsclr, rounded corners=\cornerradii ex]{\OursModule{}};
\node (convce) [draw=convdclr, fill=convclr, rounded corners=\cornerradii ex, yshift=-6ex]{Conv$3\times3$};
\node (convexp) [draw=convdclr, fill=convclr, rounded corners=\cornerradii ex, yshift=-12ex]{Conv$1\times1$};
\node (sum) [draw=none, yshift=-17.5ex]{\tikz{\node[draw=black, fill=white!70!gray, circle, scale=1.1]{};\node[scale=1.2]{$+$};}};
\node (ip)[fill=black, circle, yshift=5ex, scale=0.4]{};
\draw [->, line width=\arrowlw ex] (ip) -- (convsqz);
\draw [->, line width=\arrowlw ex] (convsqz) -- (convce);
\draw [->, line width=\arrowlw ex] (convce) -- (convexp);
\draw [->, line width=\arrowlw ex] (convexp) -- ($(sum.north)-(0ex,1.4ex)$);
\draw [->, line width=\arrowlw ex] ($(sum.south)+(0ex,1.4ex)$) -- ($(sum.south)-(0ex,1.4ex)$);
\draw [->, line width=\arrowlw ex] ($(ip.west)+(0.1ex,0ex)$) .. controls (-12ex, 4ex) and (-10ex, -17.5ex) .. ($(sum.west)+(1.5ex,0ex)$)
}};
\draw [->, line width=0.4ex] ($(resblock.east)+(0.5ex, 1ex)$) -- ($(dcsblock.west)+(3ex, 1ex)$);
\node (res)[xshift=4ex, yshift=17.2ex,scale=1.2]{ResNet};
\node (dcs)[xshift=24ex, yshift=17.2ex,scale=1.2]{\OursModule{}};
\draw [->, line width=0.4ex] (res) -- (dcs);
}};

\node (nb) [xshift = \xshftb ex, yshift=\yshftb ex, scale=\scal]{
\tikz{
\node (outer) [draw=outerclr, rounded corners=0.9ex, xshift=11ex, yshift=0.0ex, minimum width=34ex, minimum height=31ex]{};
\node (resblock) []{
\tikz{
\node (convsqz) [draw=convdclr, fill=convclr, rounded corners=\cornerradii ex]{Conv$1\times1$};
\node (convce) [draw=convdclr, fill=convclr, rounded corners=\cornerradii ex, yshift=-8ex]{Conv$3\times3$};
\node (convexp) [draw=convdclr, fill=convclr, rounded corners=\cornerradii ex, yshift=-16ex]{Conv$1\times1$};
\node (sum) [draw=none, yshift=-21.5ex]{\tikz{\node[draw=black, fill=white!70!gray, circle, scale=1.1]{};\node[scale=1.2]{$+$};}};
\node (ip)[fill=black, circle, yshift=4ex, scale=0.4]{};
\draw [->, line width=\arrowlw ex] (ip) -- (convsqz);
\draw [->, line width=\arrowlw ex] (convsqz) -- (convce);
\draw [->, line width=\arrowlw ex] (convce) -- (convexp);
\draw [->, line width=\arrowlw ex] (convexp) -- ($(sum.north)-(0ex,1.4ex)$);
\draw [->, line width=\arrowlw ex] ($(sum.south)+(0ex,1.4ex)$) -- ($(sum.south)-(0ex,1.4ex)$);
\draw [->, line width=\arrowlw ex] ($(ip.west)+(0.1ex,0ex)$) .. controls (-13ex, 4.4ex) and (-10ex, -23.5ex) .. ($(sum.west)+(1.5ex,0ex)$)
}};
\node (dcsblock) [xshift=18ex]{
\tikz{
\node (convsqz) [draw=convdclr, fill=convclr, rounded corners=\cornerradii ex]{Conv$1\times1$};
\node (dcs1) [draw=dcsclr, fill=dcsclr, rounded corners=\cornerradii ex, yshift= -4.5ex]{\OursModule{}};
\node (convce) [draw=convdclr, fill=convclr, rounded corners=\cornerradii ex, yshift=-9ex]{Conv$3\times3$};
\node (dcs2) [draw=dcsclr, fill=dcsclr, rounded corners=\cornerradii ex, yshift= -13.5ex]{\OursModule{}};
\node (convexp) [draw=convdclr, fill=convclr, rounded corners=\cornerradii ex, yshift=-18ex]{Conv$1\times1$};
\node (sum) [draw=none, yshift=-21.5ex]{\tikz{\node[draw=black, fill=white!70!gray, circle, scale=1.1]{};\node[scale=1.2]{$+$};}};
\node (ip)[fill=black, circle, yshift=4ex, scale=0.4]{};
\draw [->, line width=\arrowlw ex] (ip) -- (convsqz);
\draw [->, line width=\arrowlw ex] (convsqz) -- (dcs1);
\draw [->, line width=\arrowlw ex] (dcs1) -- (convce);
\draw [->, line width=\arrowlw ex] (convce) -- (dcs2);
\draw [->, line width=\arrowlw ex] (dcs2) -- (convexp);
\draw [->, line width=\arrowlw ex] (convexp) -- ($(sum.north)-(0ex,1.4ex)$);
\draw [->, line width=\arrowlw ex] ($(sum.south)+(0ex,1.4ex)$) -- ($(sum.south)-(0ex,1.4ex)$);
\draw [->, line width=\arrowlw ex] ($(ip.west)+(0.1ex,0ex)$) .. controls (-13ex, 4.4ex) and (-10ex, -23.5ex) .. ($(sum.west)+(1.5ex,0ex)$)
}};
\draw [->, line width=0.4ex] ($(resblock.east)+(0.5ex, 1ex)$) -- ($(dcsblock.west)+(4.8ex, 1ex)$);
\node (res)[xshift=4ex, yshift=17.2ex,scale=1.2]{ResNet-$50$};
\node (dcs)[xshift=22ex, yshift=17.2ex,scale=1.2]{\OursModule{}};
\draw [->, line width=0.4ex] (res) -- (dcs);
}};

\node (nc) [xshift = \xshftc ex, yshift=\yshftc ex, scale=\scal]{
\tikz{
\node (outer) [draw=outerclr, rounded corners=0.9ex, xshift=11ex, yshift=0.0ex, minimum width=34ex, minimum height=31	ex]{};
\node (resblock) []{
\tikz{
\node (convce) [draw=convdclr, fill=convclr, rounded corners=\cornerradii ex, yshift=-8ex]{Conv$3\times3$};
\node (convexp) [draw=convdclr, fill=convclr, rounded corners=\cornerradii ex, yshift=-19ex]{Conv$3\times3$};
\node (sum) [draw=none, yshift=-23.5ex]{\tikz{\node[draw=black, fill=white!70!gray, circle, scale=1.1]{};\node[scale=1.2]{$+$};}};
\node (ip)[fill=black, circle, yshift=1.5ex, scale=0.4]{};
\draw [->, line width=\arrowlw ex] (ip) -- (convce);
\draw [->, line width=\arrowlw ex] (convce) -- (convexp);
\draw [->, line width=\arrowlw ex] (convexp) -- ($(sum.north)-(0ex,1.4ex)$);
\draw [->, line width=\arrowlw ex] ($(sum.south)+(0ex,1.4ex)$) -- ($(sum.south)-(0ex,1.4ex)$);
\draw [->, line width=\arrowlw ex] ($(ip.west)+(0.1ex,0ex)$) .. controls ($(ip.west)+(-8.5ex,0ex)$) and ($(sum.west)+(-8.5ex,0ex)$) .. ($(sum.west)+(1.5ex,0ex)$)
}};
\node (dcsblock) [xshift=18ex]{
\tikz{
\node (dcs1) [draw=dcsclr, fill=dcsclr, rounded corners=\cornerradii ex, yshift= -2.5ex]{\OursModule{}};
\node (convce) [draw=convdclr, fill=convclr, rounded corners=\cornerradii ex, yshift=-8ex]{Conv$3\times3$};
\node (dcs2) [draw=dcsclr, fill=dcsclr, rounded corners=\cornerradii ex, yshift= -13.5ex]{\OursModule{}};
\node (convexp) [draw=convdclr, fill=convclr, rounded corners=\cornerradii ex, yshift=-19ex]{Conv$3\times3$};
\node (sum) [draw=none, yshift=-23.5ex]{\tikz{\node[draw=black, fill=white!70!gray, circle, scale=1.1]{};\node[scale=1.2]{$+$};}};
\node (ip)[fill=black, circle, yshift=1.5ex, scale=0.4]{};
\draw [->, line width=\arrowlw ex] (ip) -- (dcs1);
\draw [->, line width=\arrowlw ex] (dcs1) -- (convce);
\draw [->, line width=\arrowlw ex] (convce) -- (dcs2);
\draw [->, line width=\arrowlw ex] (dcs2) -- (convexp);
\draw [->, line width=\arrowlw ex] (convexp) -- ($(sum.north)-(0ex,1.4ex)$);
\draw [->, line width=\arrowlw ex] ($(sum.south)+(0ex,1.4ex)$) -- ($(sum.south)-(0ex,1.4ex)$);
\draw [->, line width=\arrowlw ex] ($(ip.west)+(0.1ex,0ex)$) .. controls ($(ip.west)+(-8.5ex,0ex)$) and ($(sum.west)+(-8.5ex,0ex)$) .. ($(sum.west)+(1.5ex,0ex)$)
}};
\draw [->, line width=0.4ex] ($(resblock.east)+(0.5ex, 1ex)$) -- ($(dcsblock.west)+(3ex, 1ex)$);
\node (res)[xshift=4ex, yshift=17.2ex,scale=1.2]{ResNet-$18$};
\node (dcs)[xshift=22ex, yshift=17.2ex,scale=1.2]{\OursModule{}};
\draw [->, line width=0.4ex] (res) -- (dcs);
}};

\node (nd) [xshift = \xshftd ex, yshift=\yshftd ex, scale=\scal]{
\tikz{
\node (outer) [draw=outerclr, rounded corners=0.9ex, xshift=10ex, yshift=0.0ex, minimum width=34ex, minimum height=31ex]{};
\node (vggblock) []{
\tikz{
\node (convce) [draw=convdclr, fill=convclr, rounded corners=\cornerradii ex, yshift=-12.5ex]{Conv$3\times3$};
\node (ip)[fill=black, circle, yshift=2ex, scale=0.4]{};
\draw [->, line width=\arrowlw ex] (ip) -- (convce);
\draw [->, line width=\arrowlw ex] ($(convce.south)+(0ex,0ex)$) -- ($(sum.south)-(0ex,1.4ex)$);
}};
\node (dcsblock) [xshift=20ex]{
\tikz{
\node (convsqz) [draw=dcsclr, fill=dcsclr, rounded corners=\cornerradii ex, yshift=-3ex]{\OursModule{}};
\node (convce) [draw=convdclr, fill=convclr, rounded corners=\cornerradii ex, yshift=-12.5ex]{Conv$3\times3$};
\node (ip)[fill=black, circle, yshift=2ex, scale=0.4]{};
\draw [->, line width=\arrowlw ex] (ip) -- (convsqz);
\draw [->, line width=\arrowlw ex] (convsqz) -- (convce);
\draw [->, line width=\arrowlw ex] ($(convce.south)+(0ex,0ex)$) -- ($(sum.south)-(0ex,1.4ex)$);
}};
\draw [->, line width=0.4ex] ($(vggblock.east)+(0.0ex, 1ex)$) -- ($(dcsblock.west)+(0ex, 1ex)$);
\node (vgg)[xshift=0ex, yshift=17.2ex,scale=1.2]{VGG};
\node (dcs)[xshift=20ex, yshift=17.2ex,scale=1.2]{\OursModule{}};
\draw [->, line width=0.4ex] (vgg) -- (dcs);
}};
%
%% 

%

% 
%
%%
%\draw [ gray,  line width=0.4ex] ($(na.north)+(8.7ex, 0.25ex)$) -- ($(na.south)+(8.7ex,-0.25ex)$);
%\draw [ gray,  line width=0.4ex] ($(nb.north)+(9.8ex, 0.25ex)$) -- ($(nb.south)+(9.8ex,-0.25ex)$);

% \node (a) [xshift=\xshfta ex+ 1ex, yshift=-7.3ex, scale=0.8]{(a)};
% \node (b) [xshift=\xshftb ex+ 1ex, yshift=-7.3ex, scale=0.8]{(b)};
% \node (c) [xshift=\xshftc ex+ 1ex, yshift=-7.3ex, scale=0.8]{(c)};
% \node (d) [xshift=\xshftd ex- 0.8ex, yshift=-7.3ex, scale=0.8]{(d)};
%
\node (a) [xshift=\xshfta ex+ 1ex, yshift=-12.9ex, scale=0.8]{(a)};
\node (b) [xshift=\xshftb ex+ 1ex, yshift=-12.9ex, scale=0.8]{(b)};
\node (c) [xshift=\xshftc ex+ 1ex, yshift=-43.9ex, scale=0.8]{(c)};
\node (d) [xshift=\xshftd ex- 0.8ex, yshift=-43.9ex, scale=0.8]{(d)};
\end{tikzpicture}
%
%
%\vspace{-2ex}
%%
\caption{Embedding the proposed \OursModule{} into various standard networks for various purposes. (a) \textbf{Channel Squeezing Mode}: we replace $1\times 1$ channel squeezing layers in ResNet~\citep{resnet} with \OursModule{}, where the remaining $1\times 1$ conv layers in the original ResNet are untouched as it is intended for expanding channel dimensions. (b \& c) \textbf{Network Downscaling Mode}: We insert \OursModule{} modules into ResNet and VGG~\citep{vgg}. We make the output channel dimension smaller than the input channel dimension by adjusting sampling factor $\zeta$ in \OursModule{}. In other words, depending on $\zeta$, The input and output channel dimensions of $1\times 1$ and $3\times 3$ conv layers change accordingly. As a result, as $\zeta$ gets larger, the channel dimension of the original network reduces. (c \& d) \textbf{Dynamic Channel Pruning}: These configurations are used for comparing \OursModule{} with other dynamic channel pruning approaches.}
\vspace{-2.0ex}
\label{fig:mipcasns}
%
%
%
%\end{wrapfigure}
\end{figure}
%
%%%%%%%%%%%%%%%%%%%%%%%%%%%%%%%%%%%%%%%%%%%%%%%%%%%%%%%%%%%%%%%%%%%%%%%%%%%%%%%%%%%%%%%%%%%%%%%%%%%%%%%%
%

%
%\end{wrapfigure}
%
%%%%%%%%%%%%%%%%%%%%%%%%%%%%%%%%%%%%%%%%%%%%%%%%%%%%%%%%%%%%%%%%%%%%%%%%%%%%%%%%%%%%%%%%%%%%%%%%%%%%%%%%
%

\subsection{BatchNorm}
The BatchNorm \citep{bn} operation is performed per spatial location and can be given as $\hat{X} = (X-\mu) \frac{\gamma}{\sigma} + \beta$. It can be implemented in three FLOPs, i.e., first for computing $X-\mu$, second for $\nicefrac{\gamma}{\sigma}$, and last as FMA with $\beta$. In general, $\sigma$ is stored as $\sigma^2$, therefore, it requires to compute square-root of $\sigma^2$ to obtain $\sigma$. Overall, it takes four FLOPs to implement a BatchNorm operation per spatial location. Thus, the total number of FLOPs for a BatchNorm layer can be given as:
\begin{equation}
\text{\#FLOPs} = 4 \times C \times H \times W
\end{equation}
Optionally, during inference, BN can be fused with a Conv operation where convolution is followed by BN, but we remain agnostic to such cases to account for the training phase and other architectures.
\subsection{ReLU}
A ReLU operation is given by $Y = X$ for $X \geq 0$ and $Y = 0$ for $X < 0$. It simply requires a comparison instruction, leading to the total number of FLOPs given by:
\begin{equation}
\text{\#FLOPs} = C \times H \times W
\end{equation}
\subsection{Sigmoid}
A Sigmoid operation is given by $Y = \nicefrac{1}{(1+\exp^{-X})}$. It can be implemented in four FLOPs. Therefore, the total FLOPs for a Sigmoid layer can be given by:
\begin{equation}
\text{\#FLOPs} = 4 \times C \times H \times W
\end{equation}
\subsection{Global pooling}
Apart from the above layers, in the \OursModule{} module, a global pooling operation is also performed. There are several ways to implement a global pooling operation. However, the most common is by using matrix multiplication routines and Fused-Multiply-Add (FMA) instructions. The whole channel of a feature map can be considered as a vector of size $H \times W$ which can be reduced to a scalar by taking its dot product with a vector whose all elements are equal to one. Hence, the total number of FLOPs for the global pooling operation can be given by:
\begin{equation}
\text{\#FLOPs} = C \times H \times W
\end{equation}
\subsection{Channel Sampling}
Channel fusion operates on $(C/\zeta)$ subsets, each of $\zeta$ channels. For the \texttt{Max} operation, ($\zeta-1$) compare instructions, while for \texttt{Avg} operation, ($k-1$) FMA instructions are required per-location i.e. $\Gamma_{hw}$. Thus, the total number of FLOPs for channel sampling can be given by:
\begin{equation}
\text{\#FLOPs} = (\zeta-1) \times (C/\zeta) \times H \times W
\end{equation}
The computational complexity of the \OursModule{} block can be calculated based on the several equations developed above.
%
%
%
%%%
\section{Computations \& Memory Requirements}
\label{sec:dcs_se_cbam_fbs_mem_flops_calc}
By using the above equations, we can easily compute the FLOP overhead of various modules such as SE \citep{senet}, CBAM \citep{cbam}, or FBS \citep{fbs} and demonstrated below:
\subsection{SE \citep{senet} }
\subsubsection*{Compute}
\begin{align}
\#\text{Global\_pool\_FLOPs} &= C \times H \times W \\
\#\text{Conv\_Sqz\_FLOPs} &= (C/16) \times C \\
\#\text{ReLU\_FLOPs} &= (C/16) \\
\#\text{Conv\_Exp\_FLOPs} &= C \times (C/16) \\
\#\text{Sigmoid\_FLOPs} &= 4 * C \\
\#\text{Broadcast\_Multiply\_FLOPs} &= C \times H \times W
\end{align}
%
% $\#\text{Total Flops} = 2CHW + 0.125C^2+4C$.
$\#\text{Total Flops} = 2CHW + 0.125C^2+ (65/16)C$.

\subsubsection*{Memory}
\begin{align}
\#\text{Global\_pool\_Mem} &= C\\
\#\text{Conv\_Sqz\_Mem} &= C/16\\
\#\text{Conv\_Exp\_Mem} &= C \\
\#\text{Broadcast\_Multiply\_Mem} &= C \times H \times W
\end{align}
$\#\text{Total Memory} = CHW + (33/16)C$.

\textit{Note}: ReLU and Sigmoid are ignored in memory due to their In-place operations.
\subsection{CBAM \citep{cbam}}
\subsubsection*{Compute}
\begin{align}
\#\text{Global\_Max\_pool\_FLOPs} &= C \times H \times W \\
\#\text{Global\_Avg\_pool\_FLOPs} &= C \times H \times W \\
\#\text{Conv\_Sqz\_FLOPs} &= (C/16) \times C \\
\#\text{ReLU\_FLOPs} &= (C/16) \\
\#\text{Conv\_Exp\_FLOPs} &= C \times (C/16) \\
\#\text{Sigmoid\_FLOPs} &= 4 * C \\
\#\text{Sum\_FLOPs} &= C \\
\#\text{Broadcast\_Multiply\_FLOPs} &= C \times H \times W\\
\#\text{Channel\_Max\_Pool\_FLOPs} &= (C-1) \times H \times W \\
\#\text{Channels\_Avg\_Pool\_FLOPs} &= (C-1) \times H \times W \\
\#\text{Concat\_FLOPs} &= 2 \times H \times W \\
\#\text{Conv\_FLOPs} &= 1 \times 2 \times H \times W \\
\#\text{Sigmoid\_FLOPs} &= 4 \times 1 \times H \times W \\
\#\text{Broadcast\_Multiply\_FLOPs} &= C \times H \times W
\end{align}
%
% $\#\text{Total Flops} = 6CHW + 0.125C^2+(65/16)C + 6HW$.
$\#\text{Total Flops} = 6CHW + 0.125C^2+(81/16)C + 6HW$.

\subsubsection*{Memory}
\begin{align}
\#\text{Global\_Max\_pool\_Mem} &= C\\
\#\text{Global\_Avg\_pool\_Mem} &= C\\
\#\text{Conv\_Sqz\_Mem} &= C/16\\
\#\text{Conv\_Exp\_Mem} &= C\\
\#\text{Sum\_Mem} &= C \\
\#\text{Broadcast\_Multiply\_Mem} &= C \times H \times W\\
\#\text{Channel\_Max\_Pool\_Mem} &= H \times W \\
\#\text{Channels\_Avg\_Pool\_Mem} &= H \times W \\
\#\text{Concat\_Mem} &= 2 \times H \times W \\
\#\text{Conv\_Mem} &=  H \times W \\
\#\text{Broadcast\_Multiply\_Mem} &= C \times H \times W
\end{align}
$\#\text{Total Memory} = 2CHW+5HW+(65/16)C$.

\subsection{FBS \citep{fbs}}
\subsubsection*{Compute}
\begin{align}
\#\text{Global\_pool\_FLOPs} &= C \times H \times W \\
\#\text{Conv\_Sqz\_FLOPs} &= C \times C \\
\#\text{Sigmoid\_FLOPs} &= 4 \times C \\
\#\text{Top-k\_FLOPs} &= \sum_{i \in [1, k]} (C - i) \\
\#\text{BatchNorm\_FLOPs} &= 4 \times C \times H \times W \\
\#\text{Broadcast\_Multiply\_FLOPs} &= C \times H \times W \\
\#\text{ReLU\_FLOPs} &= C \times H \times W
\end{align}

$\#\text{Total Flops} = 7CHW + C^2+ 4C + \sum_{i \in [1, k]} (C - i)$.

\subsubsection*{Memory}
\begin{align}
\#\text{Global\_pool\_Mem} &= C\\
\#\text{Conv\_Sqz\_Mem} &= C \\
\#\text{Top-k\_Mem} &= C \times H \times W \\
\#\text{Broadcast\_Multiply} &= C \times H \times W
\end{align}
$\#\text{Total Memory} = 2CHW + 2C$.

\textit{Note}: In memory, BatchNorm is ignored due to its In-place operations.

\subsection{\OursModule{}}
\subsubsection*{Compute} 
\begin{align}
\#\text{Global\_pool\_FLOPs} &= C \times H \times W \\
\#\text{Conv\_Sqz\_FLOPs} &= (C/\zeta) \times C \\
\#\text{Sigmoid\_FLOPs} &= 4 * (C/\zeta) \\
\#\text{Chanl\_Fusion\_FLOPs} &= (\zeta-1) \times (C/\zeta) \times H \times W
\end{align}
\noindent
$\#\text{Total Flops} = CHW + \frac{C^2}{\zeta} + 4 (C/\zeta) + ((\zeta-1) / \zeta) CHW $. \\
$\#\text{Total Flops}(@\zeta =1) = CHW + C^2+ 4C$.
\subsubsection*{Memory}
\begin{align}
\#\text{Global\_pool\_Mem} &= C\\
\#\text{Conv\_Sqz\_Mem} &= C/\zeta\\
\#\text{Channel Fusion Mem} &= C \times H \times W
\end{align}

$\#\text{Total Memory} = CHW + ((1+\zeta)/\zeta)C$.

\par
From the above equations, it can be seen that \OursModule{} has the lowest FLOPs and Memory required compared to all the approaches. Values are highlighted in Table~\ref{tab:flops_mem_usage}.
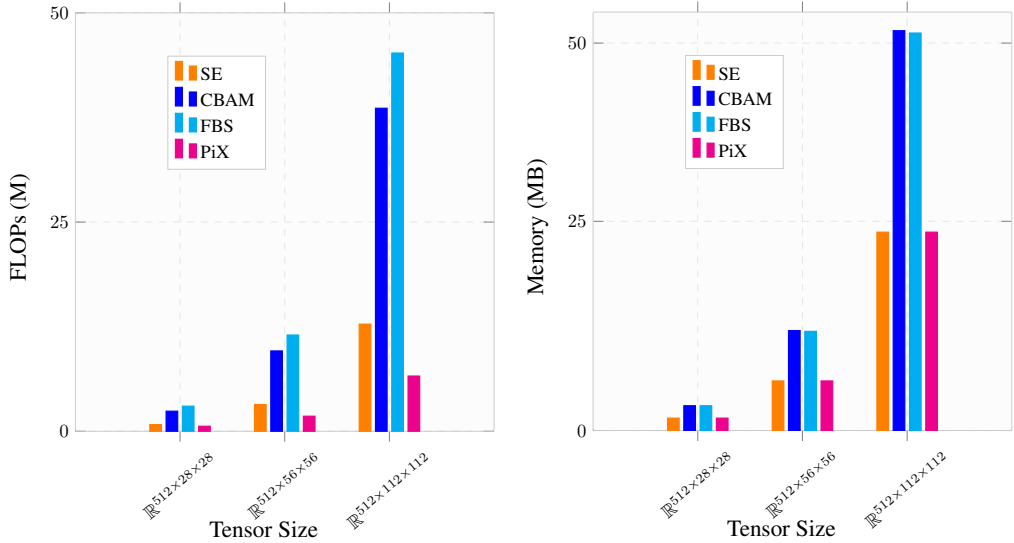
\begin{figure*}[!t]
\centering
\begin{tikzpicture}

\FPeval{\xshfta}{0}
\FPeval{\xshftb}{43.5}
\FPeval{\xshftc}{35.0}

\FPeval{\yshfta}{0-0.0}
\FPeval{\yshftb}{0-0.0}
\FPeval{\yshftc}{0-0.0}

\FPeval{\xshftd}{75}
\FPeval{\yshftd}{0-0.09}

\FPeval{\pltw}{50}
\FPeval{\plth}{50}

\FPeval{\scal}{0.88}

\FPeval{\spltw}{\pltw * \scal}
\FPeval{\splth}{\plth * \scal}

\FPeval{\mrksize}{0.2}

\FPeval{\netxshft}{5}
\FPeval{\netyshft}{0}

\FPeval{\lxshfta}{13.5}
\FPeval{\lxshftb}{13.5}
\FPeval{\lxshftc}{8.0}

\FPeval{\lyshfta}{0+25}
\FPeval{\lyshftb}{0+25}
\FPeval{\lyshftc}{0+0.5}

\FPeval{\limscale}{1.0}
\FPeval{\lscale}{0.8}

\FPeval{\xlabelxshift}{18.0}
\FPeval{\xlabelyshift}{0-4.5}
\FPeval{\ylabelxshift}{2.5}
\FPeval{\ylabelyshift}{19.5}

\FPeval{\labelscale}{1.0}
\FPeval{\ticklabelscale}{0.8}

\FPeval{\axislinecornerradii}{0.15}

\colorlet{gridclr}{white!90!black}
\colorlet{axisclr}{white!80!black}
\colorlet{axisbgclr}{white!99!black}

\colorlet{dlegendclr}{white!80!black}
\colorlet{legendclr}{white!100!black}

\colorlet{comnetclr}{white!0!magenta}
\colorlet{repvggclr}{white!0!blue}
\colorlet{resnetclr}{white!0!orange}
\colorlet{resnextclr}{white!0!cyan}
\colorlet{effnetclr}{white!0!brown}
\colorlet{netclr}{white!0!green}

\node (na) [draw=none,xshift = \xshfta ex, yshift=\yshfta ex, scale=\scal]{
\scalebox{1.0}{
\tikz{
\pgfplotsset{width=\pltw ex, height=\plth ex}
\begin{axis}[
   axis background style={fill=axisbgclr},
    title={},
    xlabel={Tensor Size },
    ylabel={FLOPs (M)},
    xmin=0.0, xmax=1.2,
    ymin=0, ymax=50,
    xtick={0.3, 0.6, 0.9},
    ytick={0, 25, 50},
  xticklabels={$\mathbb{R}^{512 \times 28 \times 28}$, $\mathbb{R}^{512 \times 56 \times 56}$, $\mathbb{R}^{512 \times 112 \times 112}$},
   yticklabels={$0$, $25$, $50$},
     axis line style={axisclr, rounded corners=\axislinecornerradii ex},
%        yticklabels={\footnotesize 0,\footnotesize 20,\footnotesize 40,\footnotesize 60,\footnotesize 80,\footnotesize 100},,
%    yticklabels={\scriptsize 0,\scriptsize 20,\scriptsize 40,\scriptsize 60,\scriptsize 80,\scriptsize 100},
  %  yticklabels={ -1, 0,  1},
        %legend pos=north west,
    legend image post style={scale =\limscale},
    legend style={at={(\lxshfta ex,\lyshfta ex)},anchor=south, legend columns = 1, draw = {dlegendclr}, fill={legendclr}, nodes={scale=\lscale}},
    ymajorgrids=true, 
    xmajorgrids=true,
    grid style={dashed, gridclr},
    major tick length=1ex,
    y label style={at={(\ylabelxshift ex, \ylabelyshift ex)}, scale=\labelscale},
    x label style={at={(\xlabelxshift ex, \xlabelyshift ex)}, scale=\labelscale},
    xticklabel style={scale=\ticklabelscale, rotate=40},
    yticklabel style={scale=\ticklabelscale},
    legend cell align={left},
    ybar,
    bar width = 5pt,
]
\addplot[
    fill=resnetclr,draw = resnetclr,
    very thick,
    line width= 0.5pt,
    mark = none%square*,
    mark size = \mrksize ex,
    ]
    coordinates {
(0.3, 0.8)
(0.6, 3.2)
(0.9, 12.8)
};
\addplot[
    fill=repvggclr,draw = repvggclr,
    very thick,
    line width= 0.5pt,
    mark = none%square*,
    mark size = \mrksize ex,
    ]
    coordinates {
(0.3, 2.4)
(0.6, 9.6)
(0.9, 38.6)
};
\addplot[
    fill=resnextclr,draw = resnextclr,
    very thick,
    line width= 0.5pt,
    mark = none%square*,
    mark size = \mrksize ex,
    ]
    coordinates {
(0.3, 3.0)
(0.6, 11.5)
(0.9, 45.2)
};
 \addplot[
    fill=comnetclr,draw = comnetclr,
    very thick,
    line width= 0.5pt,%dashed,dash pattern=on 19pt off 1pt,
    mark = none%square*,
    mark size = \mrksize ex,
    ]
    coordinates {
(0.3, 0.6)
(0.6, 1.8)
(0.9, 6.6)
};
%
 %  
 % \legend{SE \cite{senet}, CBAM \cite{cbam}, FBS \cite{fbs}, \OursModule{}}
 \legend{SE , CBAM, FBS , \OursModule{}}
\end{axis}
}}};
%
% 
%
%%%%%%%%%%%%%%%%%%%%%%%%%%%%%%%%%%%%%%%%%%%%%%%%%%%%%%%%%%%%%%%%%%%%%%%%%%%%%%%%%%%%%%%%%%%%%%%%%%%%%%%%
%
%
%
%   

\node (nb) [draw=none,xshift = \xshftb ex, yshift=\yshftb ex, scale=\scal]{
\scalebox{1.0}{
\tikz{
\pgfplotsset{width=\pltw ex, height=\plth ex}
\begin{axis}[
   axis background style={fill=axisbgclr},
    title={},
    xlabel={Tensor Size },
    ylabel={Memory (MB)},
    xmin=0.0, xmax=1.2,
    ymin=0, ymax=54,
    xtick={0.3, 0.6, 0.9},
    ytick={0, 27, 50},
  xticklabels={$\mathbb{R}^{512 \times 28 \times 28}$, $\mathbb{R}^{512 \times 56 \times 56}$, $\mathbb{R}^{512 \times 112 \times 112}$},
   yticklabels={$0$, $25$, $50$},
     axis line style={axisclr, rounded corners=\axislinecornerradii ex},
%        yticklabels={\footnotesize 0,\footnotesize 20,\footnotesize 40,\footnotesize 60,\footnotesize 80,\footnotesize 100},,
%    yticklabels={\scriptsize 0,\scriptsize 20,\scriptsize 40,\scriptsize 60,\scriptsize 80,\scriptsize 100},
  %  yticklabels={ -1, 0,  1},
        %legend pos=north west,
    legend image post style={scale =\limscale},
    legend style={at={(\lxshftb ex,\lyshftb ex)},anchor=south, legend columns = 1, draw = {dlegendclr}, fill={legendclr}, nodes={scale=\lscale}},
    ymajorgrids=true, 
    xmajorgrids=true,
    grid style={dashed, gridclr},
    major tick length=1ex,
    y label style={at={(\ylabelxshift ex, \ylabelyshift ex)}, scale=\labelscale},
    x label style={at={(\xlabelxshift ex, \xlabelyshift ex)}, scale=\labelscale},
    xticklabel style={scale=\ticklabelscale, rotate=40},
    yticklabel style={scale=\ticklabelscale},
    legend cell align={left},
    ybar,
    bar width = 5pt,
]
\addplot[
    fill=resnetclr,draw = resnetclr,
    very thick,
    line width= 0.5pt,
    mark = none%square*,
    mark size = \mrksize ex,
    ]
    coordinates {
(0.3, 1.6)
(0.6, 6.4)
(0.9, 25.6)
};
\addplot[
    fill=repvggclr,draw = repvggclr,
    very thick,
    line width= 0.5pt,
    mark = none%square*,
    mark size = \mrksize ex,
    ]
    coordinates {
(0.3, 3.2)
(0.6, 12.9)
(0.9, 51.6)
};
\addplot[
    fill=resnextclr,draw = resnextclr,
    very thick,
    line width= 0.5pt,
    mark = none%square*,
    mark size = \mrksize ex,
    ]
    coordinates {
(0.3, 3.2)
(0.6, 12.8)
(0.9, 51.3)
};
 \addplot[
    fill=comnetclr,draw = comnetclr,
    very thick,
    line width= 0.5pt,%dashed,dash pattern=on 0.5pt off 1pt,
    mark = none%square*,
    mark size = \mrksize ex,
    ]
    coordinates {
(0.3, 1.6)
(0.6, 6.4)
(0.9, 25.6)
};
%
% %  
 % \legend{SE \cite{senet}, CBAM \cite{cbam}, FBS \cite{fbs}, \OursModule{}}
 \legend{SE , CBAM, FBS , \OursModule{}}
\end{axis}
}}};
%
% 

%
%%%%%%%%%%%%%%%%%%%%%%%%%%%%%%%%%%%%%%%%%%%%%%%%%%%%%%%%%%%%%%%%%%%%%%%%%%%%%%%%%%%%%%%%%%%%%%%%%%%%%%%%
%
%

%  
\end{tikzpicture}
\vspace{-1.0ex}
\caption{Flops and Memory performance of \OursModule{} in contrast to SE \cite{senet} CBAM \cite{cbam}, and FBS \cite{fbs} per-instance of a module. In the memory plot, SE and PiX has almost same overhead but PiX lesser than SE in terms of Bytes ($\sim$ 1,000), and same is with CBAM and FBS. For this reason plots are overlapping in the memory plot. The actual values are also highlighted in Table~\ref{tab:flops_mem_usage}.}
%
%\vspace{-2.5ex}
%
\label{fig:flops_mem}
% %

% %
% %
\end{figure*}

\begin{table}[!h]
\centering

%\vspace{1ex}
\caption{This table shows FLOPs and memory usage per instance of different modules corresponding to Figure~\ref{fig:flops_mem}. These values are computed at different heights and widths of the tensor. It can be seen that PiX has the lowest FLOP overhead and also requires less memory, equivalent to SE \cite{senet} but half of CBAM \cite{cbam} and FBS \cite{fbs}.}
\label{tab:flops_mem_usage}
\vspace{-1.5ex}
\arrayrulecolor{white!60!black}
%\setlength{\arrayrulewidth}{0.1ex}

%\scriptsize
%\footnotesize
\scriptsize
%\tiny

\setlength{\tabcolsep}{11pt}

\begin{tabular}{l c c}

\toprule

\multicolumn{3}{c}{ $@\mathbb{R}^{512 \times 112 \times 112}$} \\ \midrule

\multicolumn{1}{c}{Method} & \#FLOPs (M) & \#Memory (MB) \\ \midrule

\bdota{}~SE \cite{senet}    & $12.8$  & $25.694336$   \\
\bdota{}~CBAM \cite{cbam}  & $38.6$ & $51.639424$ \\ 
\bdota{}~FBS \cite{fbs}   & $45.2$ & $51.384320$ \\
\rowcolor{rwclr} 
\bdotb{}~\OursModule{}  & $\mathbf{6.6}$ & $\mathbf{25.694208}$ \\
\midrule
\multicolumn{3}{c}{ $@\mathbb{R}^{512 \times 56 \times 56}$} \\ \midrule
\multicolumn{1}{c}{Method} & \#FLOPs (M) & \#Memory (MB) \\ \midrule

\bdota{}~SE \cite{senet}    & $3.2$  & $6.426752$   \\
\bdota{}~CBAM \cite{cbam}  & $9.6$ & $12.916096$ \\ 
\bdota{}~FBS \cite{fbs}   & $11.5$ & $12.849152$ \\
\rowcolor{rwclr} 
\bdotb{}~\OursModule{}  & $\mathbf{1.8}$ & $\mathbf{6.426624}$ \\ \midrule

\multicolumn{3}{c}{ $@\mathbb{R}^{512 \times 28 \times28}$} \\ \midrule
\multicolumn{1}{c}{Method} & \#FLOPs (M) & \#Memory (MB) \\ \midrule

\bdota{}~SE \cite{senet}    & $.837$  & $1.609856$   \\
\bdota{}~CBAM \cite{cbam}  & $2.4$ & $3.235264$ \\ 
\bdota{}~FBS \cite{fbs}   & $3.0$ & $3.215360$ \\
\rowcolor{rwclr} 
\bdotb{}~\OursModule{}  & $\mathbf{0.6}$ & $\mathbf{1.609728}$ \\

\bottomrule

\end{tabular}
\vspace{-3ex}
\end{table}
\vspace{-1em}
\section{Computation Reduction by \OursModule{} in Channels Squeezing i.e. $\zeta > 1$}
\label{sec:computeexample}
In the baseline method, the squeeze layer operates upon $\mathbf{X} \in \mathbb{R}^{C \times H \times W}$ which requires $\nicefrac{C}{\zeta} \times C \times H \times W$ FLOPs. Whereas in \OursModule{}, the global context aggregation requires $C \times H \times W$ FLOPs, cross-channel information blending requires $\nicefrac{C}{\zeta} \times C$ FLOPs. and channel fusion requires $\nicefrac{C}{\zeta} \times (\zeta-1) \times H \times W$ FLOPs. 
\par
As an example, consider an input tensor $\mathbf{X} \in \mathbb{R}^{12 \times 5 \times 5}$ to a squeeze layer kernels of size $1 \times 1$. With $\mathcal{\zeta} = 4$, the number of subsets becomes $12/ \zeta = 3$. From the equations discussed, the total number of FLOPs for a squeeze layer equals 1,275.
\begin{align}
\#\text{Conv\_FLOPs} &= 5 \times 5 \times 3 \times 12 \times 1 \times 1 = 900 \\
\#\text{BN\_FLOPs} &= 4 \times 3 \times 5 \times 5 = 300 \\
\#\text{ReLU\_FLOPs} &= 3 \times 5 \times 5 = 75
\end{align}
On the other hand, the FLOPs for the \OursModule{} module with $\zeta = 4$ equals only $811$, as described below.
\begin{align}
% \#\text{Pooling\_FLOPs} &= 12 \times 5 \times 5 = 250 \\
\#\text{Pooling\_FLOPs} &= 12 \times 5 \times 5 = 300 \\
\#\text{Conv\_FLOPs} &= 1 \times 1 \times 3 \times 12 \times 1 \times 1 = 36 \\
\#\text{Sigmoid\_FLOPs} &= 4 \times 3 \times 1 \times 1 = 12 \\
\#\text{Sampling\_FLOPs} &= 3 \times 3 \times 5 \times 5 = 225
\end{align}
In the above example, the baseline squeezing method requires 1,275 FLOPs, whereas \OursModule{} requires only $523$ and $748$ FLOPs for \OursAcronymShort{} and w-\OursAcronymShort{} fusion strategy respectively. In a similar manner, we achieve huge gains when \OursModule{} is plugged into the existing networks, which have been discussed in the experiments section of the paper.
%%%
%
%
\section{Effect of \OursAcronym{} on Memory in Channel Squeezing}
\label{sec:memory}
Despite the computational benefits, \OursModule{} does not introduce any memory overhead. The total memory required by the baseline squeeze operation with $\zeta=4$ can be given by: $\text{\#M} = C/4 \times H \times W$. On the other hand, the memory required for \OursModule{} is given by: $\text{\#M} = C + C/4 + C/4 \times H \times W$. We can see that there is a negligible increment in the memory footprint, i.e., from  $0.75 \times C \times H \times W$ to $0.75 \times C \times H \times W + 1.25C$. For FP$32$ precision, the raw memory footprint will be $4 \times M$.
\begin{table}[t]
\centering

%\vspace{1ex}
\caption{Ablation study of ResNet-$50$ + \OursModule$@\zeta=4$. Top-$1$ Accuracy on ImageNet.}
\label{tab:facsabl}
\vspace{-1.5ex}
\arrayrulecolor{white!60!black}
%\setlength{\arrayrulewidth}{0.1ex}

%\scriptsize
%\footnotesize
\scriptsize
%\tiny

\setlength{\tabcolsep}{11pt}

\begin{tabular}{c l c c}

\toprule

& \multirow{1}{*}{{Ablation}} & \multicolumn{1}{c}{{Parameter}} & \multicolumn{1}{c}{{Top-$1$ Accuracy}}
\\ \midrule

\multicolumn{1}{c|}{\multirow{2}{*}{E$0$}} & \multirow{2}{*}{{Fusion Activation}} & Sigmoid & $76.77\%$ \\
\multicolumn{1}{c|}{} & & TanH  & $76.39\%$ \\ \midrule

\multicolumn{1}{c|}{\multirow{2}{*}{E$1$}} & \multirow{2}{*}{{Batch-Norm }} & \xmark & $76.77\%$ \\
\multicolumn{1}{c|}{} & & \cmark  & $76.44\%$ \\ \midrule

\multicolumn{1}{c|}{\multirow{3}{*}{E$2$}} & \multirow{3}{*}{{$\tau$}} & $0.0$ & $76.58\%$ \\
\multicolumn{1}{c|}{} &  & $0.5$ & $76.77\%$ \\
\multicolumn{1}{c|}{} &  & $1.0$ & $76.54\%$ \\ 

\midrule 
\multicolumn{1}{c|}{\multirow{4}{*}{E$3$}} & \multirow{4}{*}{{Operator}} & \texttt{Min} & $74.68\%$ \\
\multicolumn{1}{c|}{} &  & \texttt{Max} & $76.57\%$ \\
\multicolumn{1}{c|}{} &  & \texttt{Avg} & $76.58\%$ \\ 
\multicolumn{1}{c|}{} &  & \texttt{Max+Avg} & $76.77\%$ \\ 

%
% \multicolumn{1}{c|}{\multirow{2}{*}{E$3$}} & \multirow{2}{*}{{Sampling Strategy}} & \OursAcronymShort{} (Eq.~\ref{eq:form1}) & $76.77\%$ \\
% \multicolumn{1}{c|}{} & & w-\OursAcronymShort{} (Eq.~\ref{eq:form2})  & $76.52\%$ \\ 

\bottomrule

\end{tabular}
%
%\vspace{-4ex}
%
\end{table}
\section{Ablation Study}
\label{sec:ablations}

We empirically validate \OursAcronym{} design practices using the most pertinent ablations possible. ResNet-$50$ is adopted as the baseline for this purpose, and channel squeezing mode. To begin with, we first analyze the effect of changing the activation function in the cross-channel information blending stage and then examine the effect of placing a BatchNorm prior to the sigmoidal activation. Further, we verify the behavior of proposed channel fusion strategies and also the effect of varying fusion threshold $\tau$.
\paragraph{E$\mathbf{0}$: Fusion Activation.}
The channel fusion stage utilizes the sampling probability $p$. Given that the value of $p$ lies in the interval $[0,1]$, we wish to examine the behavior of \OursModule{} if this range is achieved via a different activation function. For this purpose, we select \texttt{TanH} function which natively squeezes the input into a range $[-1,1]$. Therefore, we rewrite the mathematical expression to $0.5 \times (1 + \texttt{TanH})$ in order to place the output of TanH into the desired range of $[0,1]$. We replace the sigmoidal activation with the above expression and retrain the network. From Table~\ref{tab:facsabl}, it can be seen that sigmoidal activation outperforms the \texttt{TanH} activation for the case of \OursModule{}.

\paragraph{E$\mathbf{1}$: BatchNorm in Global Context Aggregation.}
Out of curiosity, we also analyze the behavior of \OursModule{} module by placing a BatchNorm \citep{bn} after the sampling probability predictor because the squeeze layer in the baseline method is also followed by a BatchNorm layer. We observe that BatchNorm negatively impacts performance.
\paragraph{E$\mathbf{2}$: Effect of Fusion Threshold ($\tau$).}
The hyperparameter $\tau$ is evaluated against three values $\in \{0.0, 0.5, 1.0\}$. In accordance with Eq. 2 of the main manuscript, $\tau = 0$ corresponds to \texttt{Max} operator, $\tau =1.0$ corresponds to \texttt{Avg} operator regardless of the value of $p$. Whereas $\tau = 0.5$ offers equal opportunity to the \texttt{Max} and \texttt{Avg} fusion operators which are adaptively taken care of by the value of $p$. We present an ablation over the aforementioned three values of $\tau$. 

From Table~\ref{tab:facsabl}, we observe that $\tau = 0.5$ results in best performance, which is the case when the network has the flexibility to choose from both reduction operators adaptively.
Hence, in the experiments, we use $\tau = 0.5$ for threshold-based fusion. 
\paragraph{E$\mathbf{4}$: Effect of Operator Type.}
We also experiment for operator \texttt{Min} other than \texttt{Max} and \texttt{Avg}. We found out that \texttt{Min} performs severely worse. This justifies our choice of operators and is in line with the performance achieved by using the pooling operation when they are used spatially. 

\section{Role of Fusion Probability}
 \begin{figure*}
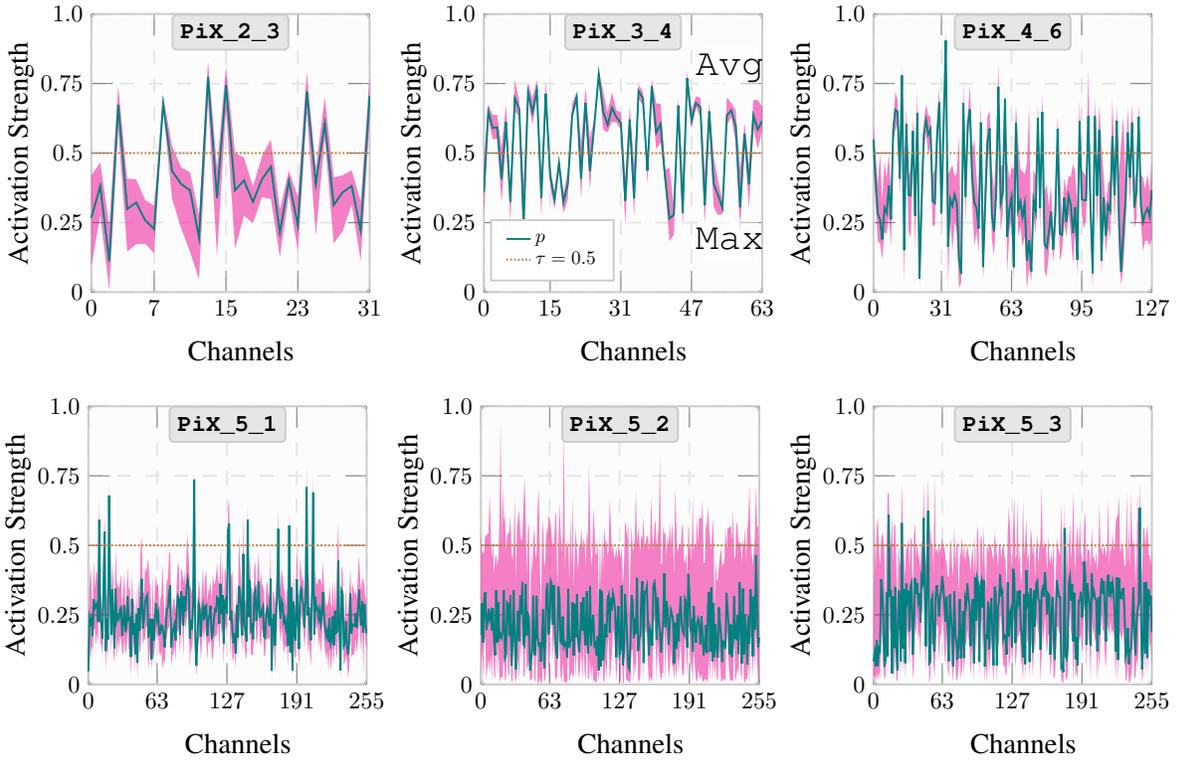

 
%\begin{wrapfigure}[14]{r}[0ex]{0.48\linewidth}
%

\vspace{-4ex}

\centering
\begin{tikzpicture}

% \FPeval{\xshfta}{0}
% \FPeval{\xshftb}{14.5}
% \FPeval{\xshftc}{29.0}
% \FPeval{\xshftd}{0-0}
% \FPeval{\xshfte}{14.5}
% \FPeval{\xshftf}{29.0}

% \FPeval{\yshfta}{0-0.0}
% \FPeval{\yshftb}{0-0.0}
% \FPeval{\yshftc}{0-0.0}
% \FPeval{\yshftd}{0-14}
% \FPeval{\yshfte}{0-14}
% \FPeval{\yshftf}{0-14}

\FPeval{\xshfta}{0}
\FPeval{\xshftb}{33.0}
\FPeval{\xshftc}{66.0}
\FPeval{\xshftd}{0-0}
\FPeval{\xshfte}{33.0}
\FPeval{\xshftf}{66.0}

\FPeval{\yshfta}{0-0.0}
\FPeval{\yshftb}{0-0.0}
\FPeval{\yshftc}{0-0.0}
\FPeval{\yshftd}{0-33}
\FPeval{\yshfte}{0-33}
\FPeval{\yshftf}{0-33}

\FPeval{\pltw}{23}
\FPeval{\plth}{23}

\FPeval{\scal}{1.8}

\FPeval{\spltw}{\pltw * \scal}
\FPeval{\splth}{\plth * \scal}

\FPeval{\mrksize}{0.2}

\FPeval{\netxshft}{5}
\FPeval{\netyshft}{0}

\FPeval{\lxshfta}{3.2}
\FPeval{\lxshftb}{3.2}
\FPeval{\lxshftc}{3.2}

\FPeval{\lyshfta}{0+0.5}
\FPeval{\lyshftb}{0+0.5}
\FPeval{\lyshftc}{0+0.5}

\FPeval{\limscale}{0.3}
\FPeval{\lscale}{0.4}

\FPeval{\xlabelxshift}{7.0}
\FPeval{\xlabelyshift}{1.5}
\FPeval{\ylabelxshift}{4.5}
\FPeval{\ylabelyshift}{6.5}

\FPeval{\labelscale}{0.6}
\FPeval{\ticklabelscale}{0.5}

\FPeval{\axislinecornerradii}{0.3}

\FPeval{\titlescale}{0.5}
\FPeval{\titlexshift}{6.5}
\FPeval{\titleyshift}{12.15}

\colorlet{gpudclr}{white!80!black}
\colorlet{gpuclr}{white!90!black}
\colorlet{gputxtclr}{white!0!black}

\colorlet{gridclr}{white!90!black}
\colorlet{axisclr}{white!80!black}
\colorlet{axisbgclr}{white!99!black}

\colorlet{dlegendclr}{white!80!black}
\colorlet{legendclr}{white!100!black}

\colorlet{comnetclr}{white!0!magenta}
\colorlet{repvggclr}{white!0!blue}
\colorlet{resnetclr}{white!0!orange}
\colorlet{resnextclr}{white!0!cyan}
\colorlet{effnetclr}{white!0!brown}

\colorlet{maxclr}{white!0!cyan}
\colorlet{avgclr}{white!0!magenta}

\FPeval{\avgmaxw}{13}
\FPeval{\avgmaxh}{6.5}

\FPeval{\avgxshift}{6.5}
\FPeval{\maxxshift}{6.5}

\FPeval{\avgyshift}{0+9.75}
\FPeval{\maxyshift}{0+3.25}

\FPeval{\txtmaxxshift}{11.5}
\FPeval{\txtmaxyshift}{10.5}
\FPeval{\txtavgxshift}{11.5}
\FPeval{\txtavgyshift}{2.5}

\FPeval{\avgmaxopacity}{0.3}

\FPeval{\opacty}{0.5}
\FPeval{\linew}{0.4}

\colorlet{clr1}{green!50!blue}
\colorlet{clr2}{white!0!magenta}

\colorlet{clr3}{white!0!brown}
\FPeval{\tauon}{0.4}
\FPeval{\tauoff}{0.4}

\FPeval{\taulinew}{0.4}

\FPeval{\txtavgmaxscale}{0.8}

\colorlet{txtmaxfillclr}{white!100!maxclr}
\colorlet{txtavgfillclr}{white!100!avgclr}
\FPeval{\txtavgmaxopacty}{0.5}

\node (na) [draw=none,xshift = \xshfta ex, yshift=\yshfta ex, scale=\scal]{
\tikz{
\pgfplotsset{width=\pltw ex, height=\plth ex}
\begin{axis}[
   axis background style={fill=axisbgclr},
    title={},
    xlabel={Channels},
    ylabel={Activation Strength},
    xmin=0, xmax=31,
    ymin=0, ymax=1.0,
    xtick={0, 7, 15, 23, 31},
    ytick={0, 0.25, 0.5, 0.75, 1.0},
   xticklabels={$0$, $7$, $15$, $23$, $31$},
   yticklabels={$0$, $0.25$, $0.5$, $0.75$, $1.0$},
%   extra y ticks={75},
%   extra y tick labels={},
%   extra x ticks={35, 60, 85},
%   extra x tick labels={},
     axis line style={axisclr, rounded corners=\axislinecornerradii ex},
%        yticklabels={\footnotesize 0,\footnotesize 20,\footnotesize 40,\footnotesize 60,\footnotesize 80,\footnotesize 100},,
%    yticklabels={\scriptsize 0,\scriptsize 20,\scriptsize 40,\scriptsize 60,\scriptsize 80,\scriptsize 100},
  %  yticklabels={ -1, 0,  1},
        %legend pos=north west,
    legend image post style={scale =\limscale},
    legend style={at={(\lxshfta ex,\lyshfta ex)},anchor=south, legend columns = 1, draw = {dlegendclr}, fill={none}, nodes={scale=\lscale}},
    ymajorgrids=true, 
    xmajorgrids=true,
    grid style={dashed, gridclr},
    major tick length=1ex,
    y label style={at={(\ylabelxshift ex, \ylabelyshift ex)}, scale=\labelscale},
    x label style={at={(\xlabelxshift ex, \xlabelyshift ex)}, scale=\labelscale},
    xticklabel style={scale=\ticklabelscale},
    yticklabel style={scale=\ticklabelscale},
    legend cell align={left},
%    ybar,
%    bar width = 4pt,
]
\input{plots/fusion_likelihood/zeta_2/prob_max_min_2}
\input{plots/fusion_likelihood/zeta_2/prob_mean_2}
\input{plots/fusion_likelihood/zeta_2/prob_tau_2}
\end{axis}
 \node [draw=gpudclr,fill=gpuclr,rounded corners=0.2ex, minimum width=10.0ex, xshift=\titlexshift ex, yshift=\titleyshift ex, scale=\titlescale]{\textcolor{gputxtclr}{\textbf{\texttt{\OursModule{}\_2\_3}}}};
}};
%
% 
%
%%%%%%%%%%%%%%%%%%%%%%%%%%%%%%%%%%%%%%%%%%%%%%%%%%%%%%%%%%%%%%%%%%%%%%%%%%%%%%%%%%%%%%%%%%%%%%%%%%%%%%%%
%
%
%
%   
\node (nb) [draw=none,xshift = \xshftb ex, yshift=\yshftb ex, scale=\scal]{
\tikz{
\pgfplotsset{width=\pltw ex, height=\plth ex}
\begin{axis}[
   axis background style={fill=axisbgclr},
    title={},
    xlabel={Channels},
    ylabel={Activation Strength},
    xmin=0, xmax=63,
    ymin=0, ymax=1.0,
    xtick={0, 15, 31, 47, 63},
    ytick={0, 0.25, 0.5, 0.75, 1.0},
   xticklabels={$0$, $15$, $31$, $47$, $63$},
   yticklabels={$0$, $0.25$, $0.5$, $0.75$, $1.0$},
%   extra y ticks={75},
%   extra y tick labels={},
%   extra x ticks={35, 60, 85},
%   extra x tick labels={},
     axis line style={axisclr, rounded corners=\axislinecornerradii ex},
%        yticklabels={\footnotesize 0,\footnotesize 20,\footnotesize 40,\footnotesize 60,\footnotesize 80,\footnotesize 100},,
%    yticklabels={\scriptsize 0,\scriptsize 20,\scriptsize 40,\scriptsize 60,\scriptsize 80,\scriptsize 100},
  %  yticklabels={ -1, 0,  1},
        %legend pos=north west,
    legend image post style={scale =\limscale},
    legend style={at={(\lxshfta ex,\lyshfta ex)},anchor=south, legend columns = 1, draw = {dlegendclr}, fill={legendclr}, nodes={scale=\lscale}},
    ymajorgrids=true, 
    xmajorgrids=true,
    grid style={dashed, gridclr},
    major tick length=1ex,
    y label style={at={(\ylabelxshift ex, \ylabelyshift ex)}, scale=\labelscale},
    x label style={at={(\xlabelxshift ex, \xlabelyshift ex)}, scale=\labelscale},
    xticklabel style={scale=\ticklabelscale},
    yticklabel style={scale=\ticklabelscale},
    legend cell align={left},
%    ybar,
%    bar width = 4pt,
]
\input{plots/fusion_likelihood/zeta_2/prob_max_min_3}
\input{plots/fusion_likelihood/zeta_2/prob_mean_3}
\input{plots/fusion_likelihood/zeta_2/prob_tau_3}%
 \node (max) [fill=txtmaxfillclr, minimum width=3.2ex, minimum height=2.5ex,xshift=\txtmaxxshift ex, yshift=\txtmaxyshift ex, scale=\txtavgmaxscale,opacity=\txtavgmaxopacty]{};
% %
\node (max) [fill=txtavgfillclr, minimum width=3.2ex, minimum height=2.5ex, xshift=\txtavgxshift ex, yshift=\txtavgyshift ex, scale=\txtavgmaxscale,opacity=\txtavgmaxopacty]{};
\node (max) [fill=none, xshift=\txtmaxxshift ex, yshift=\txtmaxyshift ex, scale=\txtavgmaxscale]{\texttt{Avg}};
% %
\node (max) [fill=none, xshift=\txtavgxshift ex, yshift=\txtavgyshift ex, scale=\txtavgmaxscale]{\texttt{Max}};
\legend{$p$, $\tau=0.5$}
\end{axis}
 \node [draw=gpudclr,fill=gpuclr,rounded corners=0.2ex, minimum width=10.0ex, xshift=\titlexshift ex, yshift=\titleyshift ex, scale=\titlescale]{\textcolor{gputxtclr}{\textbf{\texttt{\OursModule{}\_3\_4}}}};
}};
%
% 
%
%%%%%%%%%%%%%%%%%%%%%%%%%%%%%%%%%%%%%%%%%%%%%%%%%%%%%%%%%%%%%%%%%%%%%%%%%%%%%%%%%%%%%%%%%%%%%%%%%%%%%%%%
%
%
%
%   

\node (nc) [draw=none,xshift = \xshftc ex, yshift=\yshftc ex, scale=\scal]{
\tikz{
\pgfplotsset{width=\pltw ex, height=\plth ex}
\begin{axis}[
   axis background style={fill=axisbgclr},
    title={},
    xlabel={Channels},
    ylabel={Activation Strength},
    xmin=0, xmax=127,
    ymin=0, ymax=1.0,
    xtick={0, 31, 63, 95, 127},
    ytick={0, 0.25, 0.5, 0.75, 1.0},
   xticklabels={$0$, $31$, $63$, $95$, $127$},
   yticklabels={$0$, $0.25$, $0.5$, $0.75$, $1.0$},
%   extra y ticks={75},
%   extra y tick labels={},
%   extra x ticks={35, 60, 85},
%   extra x tick labels={},
     axis line style={axisclr, rounded corners=\axislinecornerradii ex},
%        yticklabels={\footnotesize 0,\footnotesize 20,\footnotesize 40,\footnotesize 60,\footnotesize 80,\footnotesize 100},,
%    yticklabels={\scriptsize 0,\scriptsize 20,\scriptsize 40,\scriptsize 60,\scriptsize 80,\scriptsize 100},
  %  yticklabels={ -1, 0,  1},
        %legend pos=north west,
    legend image post style={scale =\limscale},
    legend style={at={(\lxshfta ex,\lyshfta ex)},anchor=south, legend columns = 1, draw = {dlegendclr}, fill={legendclr}, nodes={scale=\lscale}},
    ymajorgrids=true, 
    xmajorgrids=true,
    grid style={dashed, gridclr},
    major tick length=1ex,
    y label style={at={(\ylabelxshift ex, \ylabelyshift ex)}, scale=\labelscale},
    x label style={at={(\xlabelxshift ex, \xlabelyshift ex)}, scale=\labelscale},
    xticklabel style={scale=\ticklabelscale},
    yticklabel style={scale=\ticklabelscale},
%    ybar,
%    bar width = 4pt,
]
\input{plots/fusion_likelihood/zeta_2/prob_max_min_4}
\input{plots/fusion_likelihood/zeta_2/prob_mean_4}
\input{plots/fusion_likelihood/zeta_2/prob_tau_4}%
\end{axis}
\node [draw=gpudclr,fill=gpuclr,rounded corners=0.2ex, minimum width=10.0ex, xshift=\titlexshift ex, yshift=\titleyshift ex, scale=\titlescale]{\textcolor{gputxtclr}{\textbf{\texttt{\OursModule{}\_4\_6}}}};
}};
%
% 
%
%%%%%%%%%%%%%%%%%%%%%%%%%%%%%%%%%%%%%%%%%%%%%%%%%%%%%%%%%%%%%%%%%%%%%%%%%%%%%%%%%%%%%%%%%%%%%%%%%%%%%%%%
%
%
%
\node (nd) [draw=none, xshift = \xshftd ex, yshift=\yshftd ex, scale=\scal]{
\tikz{
\pgfplotsset{width=\pltw ex, height=\plth ex}
\begin{axis}[
   axis background style={fill=axisbgclr},
    title={},
    xlabel={Channels},
    ylabel={Activation Strength},
    xmin=0, xmax=255,
    ymin=0, ymax=1.0,
    xtick={0, 63, 127, 191, 255},
    ytick={0, 0.25, 0.5, 0.75, 1.0},
   xticklabels={$0$, $63$, $127$, $191$, $255$},
   yticklabels={$0$, $0.25$, $0.5$, $0.75$, $1.0$},
%   extra y ticks={75},
%   extra y tick labels={},
%   extra x ticks={35, 60, 85},
%   extra x tick labels={},
     axis line style={axisclr, rounded corners=\axislinecornerradii ex},
%        yticklabels={\footnotesize 0,\footnotesize 20,\footnotesize 40,\footnotesize 60,\footnotesize 80,\footnotesize 100},,
%    yticklabels={\scriptsize 0,\scriptsize 20,\scriptsize 40,\scriptsize 60,\scriptsize 80,\scriptsize 100},
  %  yticklabels={ -1, 0,  1},
        %legend pos=north west,
    legend image post style={scale =\limscale},
    legend style={at={(\lxshfta ex,\lyshfta ex)},anchor=south, legend columns = 1, draw = {dlegendclr}, fill={legendclr}, nodes={scale=\lscale}},
    ymajorgrids=true, 
    xmajorgrids=true,
    grid style={dashed, gridclr},
    major tick length=1ex,
    y label style={at={(\ylabelxshift ex, \ylabelyshift ex)}, scale=\labelscale},
    x label style={at={(\xlabelxshift ex, \xlabelyshift ex)}, scale=\labelscale},
    xticklabel style={scale=\ticklabelscale},
    yticklabel style={scale=\ticklabelscale},
%    ybar,
%    bar width = 4pt,
]
\input{plots/fusion_likelihood/zeta_2/prob_max_min_5_1}
\input{plots/fusion_likelihood/zeta_2/prob_mean_5_1}
\input{plots/fusion_likelihood/zeta_2/prob_tau_5_1}%
\end{axis}
 \node [draw=gpudclr,fill=gpuclr,rounded corners=0.2ex, minimum width=10.0ex, xshift=\titlexshift ex, yshift=\titleyshift ex, scale=\titlescale]{\textcolor{gputxtclr}{\textbf{\texttt{\OursModule{}\_5\_1}}}};
}};
%
% 
%
%%%%%%%%%%%%%%%%%%%%%%%%%%%%%%%%%%%%%%%%%%%%%%%%%%%%%%%%%%%%%%%%%%%%%%%%%%%%%%%%%%%%%%%%%%%%%%%%%%%%%%%%
%
%
%
%   

\node (ne) [draw=none,xshift = \xshfte ex, yshift=\yshfte ex, scale=\scal]{
\tikz{
\pgfplotsset{width=\pltw ex, height=\plth ex}
\begin{axis}[
   axis background style={fill=axisbgclr},
    title={},
    xlabel={Channels},
    ylabel={Activation Strength},
    xmin=0, xmax=255,
    ymin=0, ymax=1.0,
    xtick={0, 63, 127, 191, 255},
    ytick={0, 0.25, 0.5, 0.75, 1.0},
   xticklabels={$0$, $63$, $127$, $191$, $255$},
   yticklabels={$0$, $0.25$, $0.5$, $0.75$, $1.0$},
%   extra y ticks={75},
%   extra y tick labels={},
%   extra x ticks={35, 60, 85},
%   extra x tick labels={},
     axis line style={axisclr, rounded corners=\axislinecornerradii ex},
%        yticklabels={\footnotesize 0,\footnotesize 20,\footnotesize 40,\footnotesize 60,\footnotesize 80,\footnotesize 100},,
%    yticklabels={\scriptsize 0,\scriptsize 20,\scriptsize 40,\scriptsize 60,\scriptsize 80,\scriptsize 100},
  %  yticklabels={ -1, 0,  1},
        %legend pos=north west,
    legend image post style={scale =\limscale},
    legend style={at={(\lxshfta ex,\lyshfta ex)},anchor=south, legend columns = 1, draw = {dlegendclr}, fill={legendclr}, nodes={scale=\lscale}},
    ymajorgrids=true, 
    xmajorgrids=true,
    grid style={dashed, gridclr},
    major tick length=1ex,
    y label style={at={(\ylabelxshift ex, \ylabelyshift ex)}, scale=\labelscale},
    x label style={at={(\xlabelxshift ex, \xlabelyshift ex)}, scale=\labelscale},
    xticklabel style={scale=\ticklabelscale},
    yticklabel style={scale=\ticklabelscale},
%    ybar,
%    bar width = 4pt,
]
\input{plots/fusion_likelihood/zeta_2/prob_max_min_5_2}
\input{plots/fusion_likelihood/zeta_2/prob_mean_5_2}
\input{plots/fusion_likelihood/zeta_2/prob_tau_5_2}%
\end{axis}
 \node [draw=gpudclr,fill=gpuclr,rounded corners=0.2ex, minimum width=10.0ex, xshift=\titlexshift ex, yshift=\titleyshift ex, scale=\titlescale]{\textcolor{gputxtclr}{\textbf{\texttt{\OursModule{}\_5\_2}}}};
}};
%
% 
%
%%%%%%%%%%%%%%%%%%%%%%%%%%%%%%%%%%%%%%%%%%%%%%%%%%%%%%%%%%%%%%%%%%%%%%%%%%%%%%%%%%%%%%%%%%%%%%%%%%%%%%%%
%
%
%
%   

\node (nf) [draw=none,xshift = \xshftf ex, yshift=\yshftf ex, scale=\scal]{
\tikz{
\pgfplotsset{width=\pltw ex, height=\plth ex}
\begin{axis}[
   axis background style={fill=axisbgclr},
    title={},
    xlabel={Channels},
    ylabel={Activation Strength},
    xmin=0, xmax=255,
    ymin=0, ymax=1.0,
    xtick={0, 63, 127, 191, 255},
    ytick={0, 0.25, 0.5, 0.75, 1.0},
   xticklabels={$0$, $63$, $127$, $191$, $255$},
   yticklabels={$0$, $0.25$, $0.5$, $0.75$, $1.0$},
%   extra y ticks={75},
%   extra y tick labels={},
%   extra x ticks={35, 60, 85},
%   extra x tick labels={},
     axis line style={axisclr, rounded corners=\axislinecornerradii ex},
%        yticklabels={\footnotesize 0,\footnotesize 20,\footnotesize 40,\footnotesize 60,\footnotesize 80,\footnotesize 100},,
%    yticklabels={\scriptsize 0,\scriptsize 20,\scriptsize 40,\scriptsize 60,\scriptsize 80,\scriptsize 100},
  %  yticklabels={ -1, 0,  1},
        %legend pos=north west,
    legend image post style={scale =\limscale},
    legend style={at={(\lxshfta ex,\lyshfta ex)},anchor=south, legend columns = 1, draw = {dlegendclr}, fill={legendclr}, nodes={scale=\lscale}},
    ymajorgrids=true, 
    xmajorgrids=true,
    grid style={dashed, gridclr},
    major tick length=1ex,
    y label style={at={(\ylabelxshift ex, \ylabelyshift ex)}, scale=\labelscale},
    x label style={at={(\xlabelxshift ex, \xlabelyshift ex)}, scale=\labelscale},
    xticklabel style={scale=\ticklabelscale},
    yticklabel style={scale=\ticklabelscale},
%    ybar,
%    bar width = 4pt,
]
\input{plots/fusion_likelihood/zeta_2/prob_max_min_5_3}
\input{plots/fusion_likelihood/zeta_2/prob_mean_5_3}
\input{plots/fusion_likelihood/zeta_2/prob_tau_5_3}%
\end{axis}
 \node [draw=gpudclr,fill=gpuclr,rounded corners=0.2ex, minimum width=10.0ex, xshift=\titlexshift ex, yshift=\titleyshift ex, scale=\titlescale]{\textcolor{gputxtclr}{\textbf{\texttt{\OursModule{}\_5\_3}}}};
}};
%
% 
%
%%%%%%%%%%%%%%%%%%%%%%%%%%%%%%%%%%%%%%%%%%%%%%%%%%%%%%%%%%%%%%%%%%%%%%%%%%%%%%%%%%%%%%%%%%%%%%%%%%%%%%%%
%
%
%
%   

%  
\end{tikzpicture}
%
%
%
% \vspace{-2.5ex}
%%
\caption{Sampling probability at different stages of ResNet-$50$ + \OursModule{}. Stage named as: \texttt{\OursModule{}\_STAGE\_ID\_BLOCK\_ID} \cite{resnet}.}
%
%\vspace{-2.5ex}
%
\label{fig:fusionlikelihood}
%

%\vspace{-0.5ex}
%
%
 \end{figure*}
%\end{wrapfigure}
%
%%%%%%%%%%%%%%%%%%%%%%%%%%%%%%%%%%%%%%%%%%%%%%%%%%%%%%%%%%%%%%%%%%%%%%%%%%%%%%%%%%%%%%%%%%%%%%%%%%%%%%%%
%
% 
%
%%%%%%%%%%%%%%%%%%%%%%%%%%%%%%%%%%%%%%%%%%%%%%%%%%%%%%%%%%%%%%%%%%%%%%%%%%%%%%%%%%%%%%%%%%%%%%%%%%%%%%%%
%
%
%

%
We analyze the sampling probabilities across all classes in the ImageNet validation set for ResNet-$50$ + \OursModule{} $@\zeta=2$ for the last block of each stage (Figure~\ref{fig:fusionlikelihood}). 
\par
It can be seen that the importance of probability is significant since distribution for the fusion operator selection is variable, i.e., while training, the network does not bias towards only one type of fusion operator, indicating that both of the fusion operators are crucial. In the deeper layers (stage-$5$), the variance starts increasing, indicating deeper layers are class-specific and need different activation distributions. This is in line with \citep{senet}. Moreover, we notice that, unlike \citep{senet}, none of the layers in the stage-$5$ show saturation. This is also an indication that \OursModule{} naturally pushes a convolution layer to learn more complex representation. 

\section{GradCAM Visualization}
 \begin{figure*}[!h]
%\begin{wrapfigure}[8]{r}[0ex]{0.6\linewidth}
\centering

\vspace{-3.5ex}

\FPeval{\imw}{7.5}
\FPeval{\imh}{7.5}

\begin{tikzpicture}

\node (resnet) [scale=1.2]{
\tikz{
%\node (outer) [draw=white!20!gray, rounded corners=0.2ex, minimum height =25ex, minimum width=43ex, xshift=15ex, yshift = -8ex]{};
%
\node (im1) [xshift=0ex]{\includegraphics[width=\imw ex, height=\imh ex]{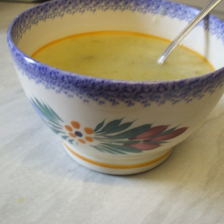}};
\node (im2) [xshift=(\imw ex+0.5ex)]{\includegraphics[width=\imw ex, height=\imh ex]{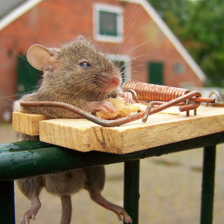}};
\node (im3) [xshift=2*(\imw ex+0.5ex)]{\includegraphics[width=\imw ex, height=\imh ex]{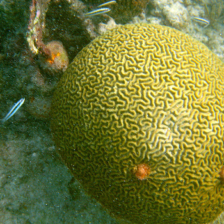}};
\node (im4) [xshift=3*(\imw ex+0.5ex)]{\includegraphics[width=\imw ex, height=\imh ex]{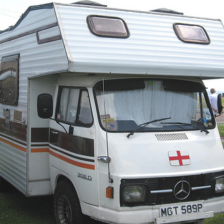}};
\node (im5) [xshift=4*(\imw ex+0.5ex)]{\includegraphics[width=\imw ex, height=\imh ex]{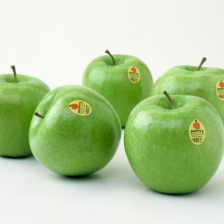}};
\node (im1) [xshift=0ex, yshift=-1*(\imh ex+0.5ex)]{\includegraphics[width=\imw ex, height=\imh ex]{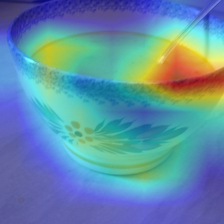}};
\node (im2) [xshift=(\imw ex+0.5ex), yshift=-1*(\imh ex+0.5ex)]{\includegraphics[width=\imw ex, height=\imh ex]{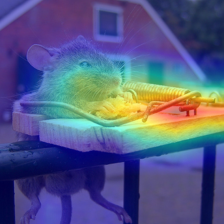}};
\node (im3) [xshift=2*(\imw ex+0.5ex), yshift=-1*(\imh ex+0.5ex)]{\includegraphics[width=\imw ex, height=\imh ex]{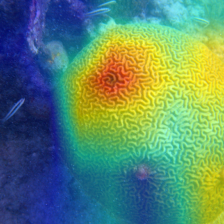}};
\node (im4) [xshift=3*(\imw ex+0.5ex), yshift=-1*(\imh ex+0.5ex)]{\includegraphics[width=\imw ex, height=\imh ex]{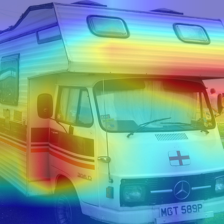}};
\node (im5) [xshift=4*(\imw ex+0.5ex), yshift=-1*(\imh ex+0.5ex)]{\includegraphics[width=\imw ex, height=\imh ex]{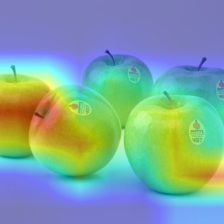}};
\node (im1) [xshift=0ex, yshift=-2*(\imh ex+0.5ex)]{\includegraphics[width=\imw ex, height=\imh ex]{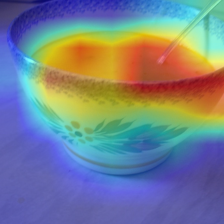}};
\node (im2) [xshift=(\imw ex+0.5ex), yshift=-2*(\imh ex+0.5ex)]{\includegraphics[width=\imw ex, height=\imh ex]{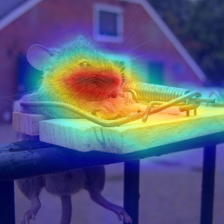}};
\node (im3) [xshift=2*(\imw ex+0.5ex), yshift=-2*(\imh ex+0.5ex)]{\includegraphics[width=\imw ex, height=\imh ex]{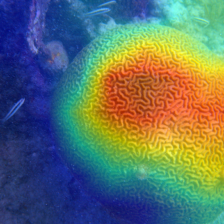}};
\node (im4) [xshift=3*(\imw ex+0.5ex), yshift=-2*(\imh ex+0.5ex)]{\includegraphics[width=\imw ex, height=\imh ex]{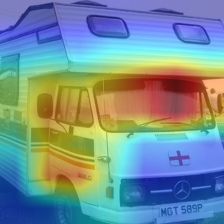}};
\node (im5) [xshift=4*(\imw ex+0.5ex), yshift=-2*(\imh ex+0.5ex)]{\includegraphics[width=\imw ex, height=\imh ex]{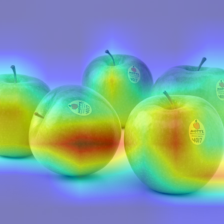}};
\node () [xshift=-5ex, rotate=90]{\scriptsize Image};
\node () [xshift=-5ex, yshift=-8.3ex, rotate=90]{\scriptsize ResNet-$50$};
\node () [xshift=-5ex, yshift=-16ex, rotate=90]{\scriptsize \OursModule{}};
\FPeval{\iyshft}{4.7}
\node (i0) [xshift=0*(\imw ex+0.5ex), yshift=\iyshft ex, scale=1.22]{\scriptsize R-I$0$};
\node (i1) [xshift=1*(\imw ex+0.5ex), yshift=\iyshft ex, scale=1.22]{\scriptsize R-I$1$};
\node (i2) [xshift=2*(\imw ex+0.5ex), yshift=\iyshft ex, scale=1.22]{\scriptsize R-I$2$};
\node (i3) [xshift=3*(\imw ex+0.5ex), yshift=\iyshft ex, scale=1.22]{\scriptsize R-I$3$};
\node (i4) [xshift=4*(\imw ex+0.5ex), yshift=\iyshft ex, scale=1.22]{\scriptsize R-I$4$};
}
};
\hspace{5mm}
\node (vgg) [xshift= 50.7ex, scale=1.2]{
\tikz{
%\node (outer) [draw=white!20!gray, rounded corners=0.2ex, minimum height =25ex, minimum width=43ex, xshift=15ex, yshift = -8ex]{};
%
\node (im6) [xshift=0*(\imw ex+0.5ex)]{\includegraphics[width=\imw ex, height=\imh ex]{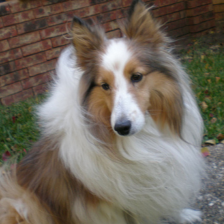}};
\node (im7) [xshift=1*(\imw ex+0.5ex)]{\includegraphics[width=\imw ex, height=\imh ex]{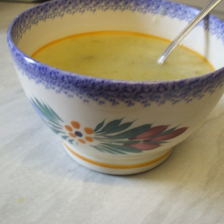}};
\node (im8) [xshift=2*(\imw ex+0.5ex)]{\includegraphics[width=\imw ex, height=\imh ex]{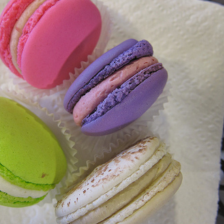}};
\node (im9) [xshift=3*(\imw ex+0.5ex)]{\includegraphics[width=\imw ex, height=\imh ex]{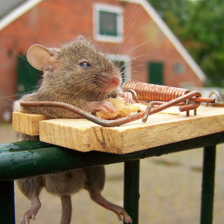}};
\node (im10) [xshift=4*(\imw ex+0.5ex)]{\includegraphics[width=\imw ex, height=\imh ex]{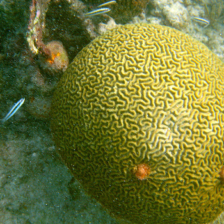}};
\node (im6) [xshift=0*(\imw ex+0.5ex), yshift=-1*(\imh ex+0.5ex)]{\includegraphics[width=\imw ex, height=\imh ex]{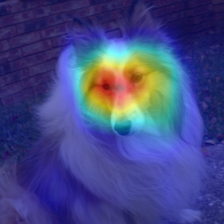}};
\node (im7) [xshift=1*(\imw ex+0.5ex), yshift=-1*(\imh ex+0.5ex)]{\includegraphics[width=\imw ex, height=\imh ex]{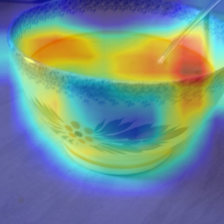}};
\node (im8) [xshift=2*(\imw ex+0.5ex), yshift=-1*(\imh ex+0.5ex)]{\includegraphics[width=\imw ex, height=\imh ex]{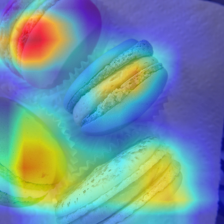}};
\node (im9) [xshift=3*(\imw ex+0.5ex), yshift=-1*(\imh ex+0.5ex)]{\includegraphics[width=\imw ex, height=\imh ex]{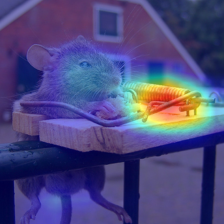}};
\node (im10) [xshift=4*(\imw ex+0.5ex), yshift=-1*(\imh ex+0.5ex)]{\includegraphics[width=\imw ex, height=\imh ex]{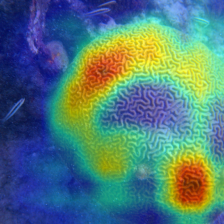}};
\node (im6) [xshift=0*(\imw ex+0.5ex), yshift=-2*(\imh ex+0.5ex)]{\includegraphics[width=\imw ex, height=\imh ex]{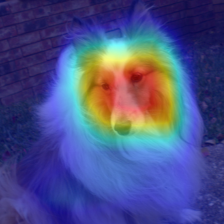}};
\node (im7) [xshift=1*(\imw ex+0.5ex), yshift=-2*(\imh ex+0.5ex)]{\includegraphics[width=\imw ex, height=\imh ex]{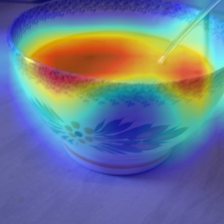}};
\node (im8) [xshift=2*(\imw ex+0.5ex), yshift=-2*(\imh ex+0.5ex)]{\includegraphics[width=\imw ex, height=\imh ex]{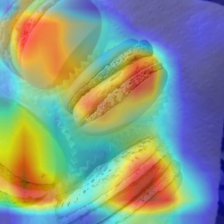}};
\node (im9) [xshift=3*(\imw ex+0.5ex), yshift=-2*(\imh ex+0.5ex)]{\includegraphics[width=\imw ex, height=\imh ex]{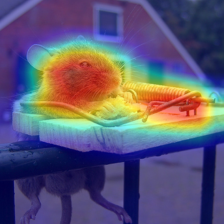}};
\node (im10) [xshift=4*(\imw ex+0.5ex), yshift=-2*(\imh ex+0.5ex)]{\includegraphics[width=\imw ex, height=\imh ex]{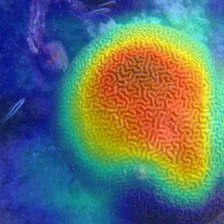}};
\node () [xshift=-5ex, rotate=90]{\scriptsize Image};
\node () [xshift=-5ex, yshift=-8.3ex, rotate=90]{\scriptsize VGG-$16$};
\node () [xshift=-5ex, yshift=-16ex, rotate=90]{\scriptsize \OursModule{}};
\FPeval{\iyshft}{4.7}
\node (i0) [xshift=0*(\imw ex +0.5ex), yshift=\iyshft ex, scale=1.22]{\scriptsize V-I$0$};
\node (i1) [xshift=1*(\imw ex+0.5ex), yshift=\iyshft ex, scale=1.22]{\scriptsize V-I$1$};
\node (i2) [xshift=2*(\imw ex+0.5ex), yshift=\iyshft ex, scale=1.22]{\scriptsize V-I$2$};
\node (i3) [xshift=3*(\imw ex+0.5ex), yshift=\iyshft ex, scale=1.22]{\scriptsize V-I$3$};
\node (i4) [xshift=4*(\imw ex+0.5ex), yshift=\iyshft ex, scale=1.22]{\scriptsize V-I$4$};
}
};

\end{tikzpicture}
\vspace{-1ex}
\caption{GradCAM for ResNet-$50$ + \OursModule{}, VGG-$16$ + \OursModule{}. solid red shows more confidence for a pixel to belong to a class.}
\label{fig:gradcam}
%
%\vspace{-3ex}
%
\end{figure*}
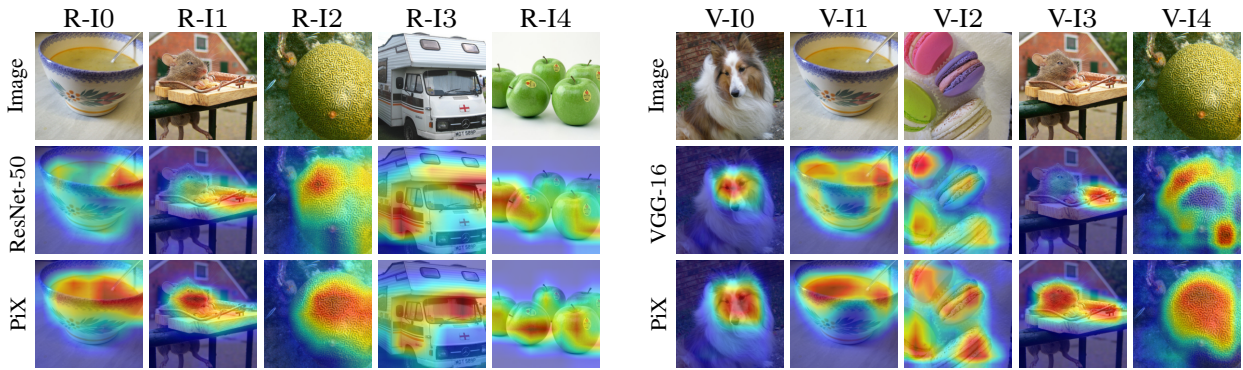
%\end{wrapfigure}

%
The performance of \OursModule{}, especially in the channel squeezing mode, inspires us to analyze how \OursModule{} attends the spatial regions relative to the baseline. 
It explains qualitatively the improved performance of \OursModule{} despite the reduction in FLOPs. We use GradCAM \citep{gradcam} for this purpose. 
\par
Figure~\ref{fig:gradcam} shows the analysis for ResNet and VGG. Noticeably, \OursModule{} shows improvement in the attended regions of a target class relative to the baseline (R-I$2$, V-I$4$). Also, in images with multiple instances, \OursModule{} focuses on each instance strongly (R-I$4$, V-I$2$), indicating that \OursModule{} enhances network's generalization by learning to emphasize class-specific parts.
\section{GPU Deployment for \OursAcronym{}}
\label{sec:gpuimpl}
The implementation of \OursAcronymShort{} is quite straightforward and fully parallelizable. The sampling probability and output feature map computations are parallelizable because they are pointwise operations. 

\OursModule{} can be implemented directly with the fundamental operators of Pytorch \citep{pytorch}. However, since we perform operations over each subset and each location independently, therefore, \OursModule{} requires merely $10-15$ lines of NVIDIA's CUDA kernel code or any other parallelization paradigm.
\section{Codes and Implementation}
\label{sec:codes}
The code and the pre-trained models are open-sourced in PyTorch \citep{pytorch}. See below for Python and CUDA snippets.

\section{Training Specifications.}
\label{sec:training}
The training procedure is kept standard to ensure reproducibility. We use a batch size of $256$, which is split across $8$ GPUs. We use a RandomResized crop \citep{pytorch} of $224$$\times $$224$ pixels, along with a horizontal flip. We use \texttt{SGD} with \texttt{Nesterov} momentum of $0.9$, \texttt{base\_lr}=$0.1$ with CosineAnnealing~\citep{cosineanneal} rate scheduler and a weight decay of $0.0001$. Unless otherwise stated, all models are trained from scratch for $120$ epochs following~\citep{resnet}.

\clearpage

%{
%\onecolumn{
%\inputminted[fontsize=\scriptsize, linenos=true, python3=true]{python}{code/pix_layer.py}
%}
%
%{
%\onecolumn{
%\inputminted[fontsize=\scriptsize, linenos]{cpp}{code/pix_cuda.cu}
%
%}

\clearpage

\end{document}